\documentclass[hidelinks,onefignum,onetabnum]{siamart220329}

\usepackage{bm}
\usepackage{subcaption}
\usepackage{amssymb, mathtools, amsmath}

\usepackage{cancel}
\usepackage{lipsum}
\usepackage{amssymb}
\usepackage{amsfonts}
\usepackage{graphicx}
\usepackage{epstopdf}
\usepackage{algorithmic}
\ifpdf
  \DeclareGraphicsExtensions{.eps,.pdf,.png,.jpg}
\else
  \DeclareGraphicsExtensions{.eps}
\fi

\newsiamremark{remark}{Remark}
\newsiamremark{hypothesis}{Hypothesis}
\crefname{hypothesis}{Hypothesis}{Hypotheses}
\newsiamthm{claim}{Claim}

\headers{B-KRnet and its applications to density estimation and approximation}{Li Zeng, Xiaoliang Wan, and Tao Zhou}

\title{Bounded KRnet and its applications to density estimation and approximation\thanks{Submitted to the editor's DATE.
\funding{
This work is supported by the NSF of China (No. 12288201), the Strategic Priority Research Program of Chinese Academy of Sciences (Grant No. XDA25010404), the National Key R\&D Program of China (2020YFA0712000), the youth innovation promotion association (CAS), and Henan Academy of Sciences.
The second author is supported by NSF grant DMS-1913163.
}}}

\author{Li Zeng\thanks{School of Mathematics and Statistics, Fuzhou University, Fuzhou, China
  (\email{zengli@lsec.cc.ac.cn}).}
\and Xiaoliang Wan\thanks{Department of Mathematics and Center for Computation and Technology, Louisiana State University, Baton Rouge 70803, USA 
  (\email{xlwan@lsu.edu}).}
\and Tao Zhou\thanks{Institute of Computational Mathematics and Scientific/Engineering
Computing, Academy of Mathematics and Systems Science, Chinese Academy
of Sciences, Beijing, China (\email{tzhou@lsec.cc.ac.cn}).}}

\usepackage{amsopn}

\ifpdf
\hypersetup{
  pdftitle={Bounded KRnet and its applications to density estimation and approximation},
  pdfauthor={Li Zeng, Xiaoliang Wan, and Tao Zhou}
}
\fi
\usepackage{ulem}

\usepackage{tikz}

\begin{document}

\maketitle

\begin{abstract}
In this paper, we develop an invertible mapping, called B-KRnet, on a bounded domain and apply it 
to density estimation/approximation for data or the solutions of PDEs such as the Fokker-Planck equation and the Keller-Segel equation. Similar to KRnet, B-KRnet consists of a series of coupling layers with progressively fewer active transformation dimensions, inspired by the triangular structure of the Knothe–Rosenblatt (KR) rearrangement. The main difference between B-KRnet and KRnet is that B-KRnet is defined on a hypercube while KRnet is defined on the whole space, in other words, a new mechanism is introduced in B-KRnet to maintain the exact invertibility. 
Using B-KRnet as a transport map, we obtain an explicit probability density function (PDF) model that corresponds to the pushforward of a base (uniform) distribution on the hypercube. 
It can be directly applied to density estimation when only data are available.
By coupling KRnet and B-KRnet, we define a deep generative model on a high-dimensional domain where some dimensions are bounded and other dimensions are unbounded. A typical case is the solution of the stationary kinetic Fokker-Planck equation, which is a PDF of position and momentum. Based on B-KRnet, we develop an adaptive learning approach to approximate partial differential equations whose solutions are PDFs or can be treated as PDFs. A variety of numerical experiments is presented to demonstrate the effectiveness of B-KRnet.
\end{abstract}



\section{Introduction}\label{sec:introduction}
Density estimation (approximation) of high-dimensional data (probability density functions) plays a key role in many applications in a broad spectrum of fields, such as statistical inference, computer vision, machine learning, as well as physics and engineering.    
There is a variety of conventional density estimation approaches, such as histograms \cite{scott1979optimal}, orthogonal series estimation \cite{schwartz1967estimation}, and kernel density estimation \cite{chen2017tutorial}, 
to name a few. It is very often that the density is defined on a bounded interval, e.g.,  human ages, and estimates that yield a positive probability outside the bounded interval are unacceptable \cite{silverman2018density}. On the other hand, 
some popular methods, like kernel density estimation, can suffer from the so-called boundary bias problem when dealing with bounded data \cite{jones1993simple}, which 
has motivated the development of more advanced variants \cite{chen1999beta, kang2018kernel, bertin2019adaptive}. Nevertheless, classical methods are usually limited to low-dimensional cases. 

In recent years, deep generative models have achieved impressive performance in high-dimensional density estimation and sample generation. In \cite{goodfellow2020generative}, generative adversarial networks (GANs) are proposed to generate samples that closely match the data distribution by training two networks --- a generator and a discriminator --- through a zero-sum game. 
In \cite{kingma2013auto} variational autoencoder (VAE) has been developed to learn a probability distribution over a latent space and use it to generate new samples.
Normalizing flows (NFs) \cite{dinh2014nice, dinh2016density, kingma2016improved, papamakarios2017masked, kingma2018glow, muller2019neural, durkan2019cubic, durkan2019neural, ho2019flow++, tang2020deep} construct an invertible transformation from a target distribution to a base distribution (e.g., Gaussian) and build an explicit PDF model via the change of variables. The Normalizing flow can be viewed as a deep learning framework for learning transport-based representation of PDFs. Transport maps have attracted strong interest in the context of modeling complex PDFs\cite{zech2022sparse1,zech2022sparse2, baptista2023representation, wang2022minimax}.
We pay particular attention to flow-based generative models because they can also be used directly as an approximator for a given PDF, whereas GANs do not offer an explicit likelihood and VAEs require an auxiliary latent variable. KRnet proposed in \cite{tang2020deep} is a generalization of the real NVP \cite{dinh2016density}, employing a structure in which the dimensionality of active transformations decreases progressively. Its design is inspired by the KR rearrangement, which defines a strictly triangular transport map. However, KRnet departs from the classical formulation by replacing each component of the triangular map with a composition of coupling layers. This breaks the strict triangularity but introduces greater flexibility. This technique has gained encouraging performance in density estimation \cite{tang2020deep, wan2021augmented}, solving partial differential equations (PDEs) \cite{tang2023pinns} especially Fokker-Planck equations \cite{tang2022adaptive, feng2021solving, zeng2023adaptive} and Bayesian inverse problems \cite{feng2023dimension, wan2020vae}. Notice that KRnet is defined on the whole space $\mathbb{R}^d$, constructing a bijection from $\mathbb{R}^d$ to $\mathbb{R}^d$. 
If the unknown PDF has a compact domain $\Omega\subset\mathbb{R}^d$, the regular KRnet may not be effective for PDF approximation. 
 To address this issue, we propose a new normalizing flow called bounded KRnet (B-KRnet), specifically designed for bounded domains. B-KRnet inherits the structure with descending active transformation dimensions of KRnet but replaces its inner unbounded transformations with nonlinear bijections constructed from cumulative distribution functions (CDFs).
The CDF coupling layer differs from the previous spline flows \cite{muller2019neural, durkan2019cubic, durkan2019neural} in three main characters: 
\begin{itemize}
	\item 
 By inheriting the structure of KRnet, a coarse mesh can be applied to each dimension. We rescale a rectangular domain to $[-1,1]^d$ and divide the interval $[-1,1]$ into a small number of subintervals.
	\item We use a tangent hyperbolic function to explicitly define the range of the parameters in the CDF coupling layer, ensuring
 that the conditioning of the transformation is well maintained. 
       \item 
       The grid points of the partition of $[-1,1]$ are determined by the outputs of a neural network, enabling implicit mesh adaptivity in the training process.
\end{itemize}

An important application of deep generative modeling in scientific computing is the approximation of PDFs or quantities that can be 
considered as unnormalized PDFs. 
Note that many (high-dimensional) PDEs, describing complex and physical systems, are related to probability densities, e.g., the Boltzmann equation \cite{harris2004introduction}, the Keller-Segel equation \cite{keller1970initiation} and the Poisson-Nernst-Planck equation \cite{golovnev2010steady}, etc. These equations have a wide range of applications in biology and chemistry \cite{eisenberg1998ionic, corrias2004global, horstmann20031970} but cannot be solved analytically, 
prompting an increase in active research on their numerical solutions. Besides traditional methods including both deterministic and stochastic approaches, deep learning methods have recently demonstrated a growing success in solving PDEs. 
In \cite{yu2018deep}, a deep Ritz method is proposed to solve PDEs with a variational formulation. 
In \cite{raissi2019physics}, physics-informed neural networks (PINNs) are developed based on the strong form of PDEs. In \cite{zang2020weak}, the approximate solution is obtained by solving a min-max problem whose loss function is based on the weak formulation of PDEs. In addition, many hybrid methods have also been proposed, e.g., PINNs and generative models are coupled to deal with stochastic differential equations in \cite{yang2020physics, yang2019adversarial,guo2022normalizing}.

In this work, we also combine the B-KRnet model and PINNs to solve PDEs whose solutions are or can be 
considered as PDFs. In particular, we couple B-KRnet with the regular KRnet to obtain more flexible PDF models for complex physical systems. For example, the solution of the kinetic Fokker-Planck equation \cite{dong2021kinetic} is a high-dimensional PDF depending on the location and velocity of a particle, where the location is bounded while the velocity is unbounded. Classical approaches such as spectral methods and finite element methods are not efficient due to the curse of dimensionality. 
In \cite{JCP2020_Lee}, a neural network approach is applied to approximate the kinetic Fokker-Planck equation. However, this approach faces two issues: the constraints on a PDF cannot be exactly satisfied, and the statistics derived from the approximated PDF cannot be efficiently computed.
These two issues can be naturally avoided by using a normalizing flow model such as KRnet, which explicitly defines a PDF 
and generates exact random samples through an invertible mapping. When applying KRnet to approximate PDFs on a bounded domain, 
handling the jump on the boundary can be challenging. A more effective strategy is to couple B-KRnet and KRnet to
create an explicit PDF model on a mixed bounded-unbounded domain. 
When approximating the kinetic Fokker-Planck equation, we use the exact random samples from the approximate PDF to update the training set to improve accuracy. Similar techniques have been successfully applied in \cite{tang2022adaptive, feng2021solving, zeng2023adaptive} for solving unbounded Fokker-Planck equations.  

The rest of this paper is structured as follows. 
\Cref{sec:B-KRnet} provides the definition of B-KRnet in detail. Our methodology for density estimation is presented in 
\cref{sec:density_estimation} along with numerical verification. In \cref{sec:adaptive_method} and \cref{sec:mixed_domain}, we present an adaptive density approximation scheme for solving PDEs involving density followed by some numerical experiments in \cref{sec:numerical_experiments_for_pdes}.
The concluding remarks are given in \cref{sec:conclusions}.

\section{B-KRnet}\label{sec:B-KRnet}
Similar to the regular KRnet, B-KRnet is a flow-based model, which can serve
as a generic PDF model for both density estimation/approximation and sample generation for scientific computing problems. In this section, we define the structure of B-KRnet. Let $\Omega\subset\mathbb{R}^d$ be a compact set. Let$\bm{X}\in \Omega$ be an unknown random vector associated with a given dataset, and let $\bm{Z}\in \Omega$ be a simple random variable with a known PDF $p_{\bm{Z}}(\bm{z})$, e.g. uniform distribution. 
Flow-based models aim to construct an invertible mapping $f:\bm{x}\, \mapsto\, \bm{z}$. Then the PDF of $\bm{X}=f^{-1}(\bm{Z})$ can be modeled by the change of variables,
\begin{equation}
p_{\bm{X}}(\bm{x})=p_{\bm{Z}}(f(\bm{x}))\Big|\det \nabla_{\bm{x}} f(\bm{x})\Big|.
\label{eqn:variable_formula}
\end{equation}
We refer to \(\bm{Z}\) as the base random variable, which is called prior in the setting of deep generative models. The corresponding PDF \(p_{\bm{Z}}(\cdot)\) is referred to as the base distribution in the context.
Given observations of $\bm{X}$, the unknown invertible mapping can be learned via the maximum likelihood estimation. Flow-based models construct a complex invertible mapping by stacking a sequence of simple bijections, i.e., 
\begin{equation}
\bm{z}=f(\bm{x})=f_{[L]}\circ f_{[L-1]}\circ \cdots \circ f_{[1]}(\bm{x}),
\label{com_no_time}
\end{equation}
where $f_{[i]}(\cdot)$ are based on shallow neural networks. The inverse and Jacobian determinant are given as 
\begin{eqnarray}
\bm{x} = f^{-1}(\bm{z})=f_{[1]}^{-1}\circ \cdots \circ f_{[L-1]}^{-1}\circ f^{-1}_{[L]}(\bm{z}),\\
\vert \det \nabla_{\bm{x}} f(\cdot)\vert = \prod_{i=1}^L\vert \det \nabla _{\bm{x}_{[i-1]}}f_{[i]}(\cdot)\vert,
\end{eqnarray}
where $\bm{x}_{[i-1]}$ indicates the immediate variables with $\bm{x}_{[0]}=\bm{x}, \bm{x}_{[L]}=\bm{z}.$ The inverse and Jacobi matrix of $f_{[i]}(\cdot)$ can be computed efficiently. 
\subsection{Architecture}

\subsubsection{Overview of the Overall Structure of B-KRnet}
We inherit the structure with descending transformation dimensions of the regular KRnet \cite{tang2020deep}, which is motivated by the KR rearrangement \cite{santambrogio2015optimal} to improve the accuracy and reduce the model complexity especially when the dimension is high. 
For ease of understanding, we briefly recall the form of the KR rearrangement \cite{zech2022sparse1}.

Let \(\rho\) and \(\pi\) be two absolutely continuous meansures on \(\Omega\), associated with the PDFs 
\(p_{\bm{X}}>0\) and \(p_{\bm{Z}}>0\), respectively. We denote by \(T_{\#}\rho=\pi\) the pushforward of \(\rho\) under \(T\), defined by \(T_{\#}\rho(A)=\rho(T^{\rm{-1}}(A))\), for any measurable set  \(A\subset\Omega\).
KR rearrangement ensures the existence of a triangular mapping \(T:\Omega\to\Omega \) such that \(T_{\#}\rho=\pi\):
\begin{equation}
\label{eqn:KR_rearrangement}
\begin{aligned}
\bm{x}=(x_1,\dots, x_d)^{\rm{T}}, \, \bm{z}=(z_1,\dots, z_d)^{\rm{T}},\\
\bm{z}=T(\bm{x})=\left(\;
\begin{aligned}
&T_1(x_1)\\
&T_2(x_1, x_2)\\
&\quad \vdots\\
&T_{d}(x_1,x_2,\dots, x_d)
\end{aligned}\;\right).
\end{aligned}
\end{equation}

Denote \(\bm{x}_{i:j}=(x_i,x_{i+1},\dots,x_j)^{\mathrm{T}}\) for \(1\leq i\leq j\leq d\).
The triangular mapping \(T\) can be reformulated as a composition of mappings: \begin{equation}
\label{eqn:KR_rearrangement_ref}
       \bm{z}= T(\bm{x})=\left(\;\begin{aligned}
&T_1\\
&x_{2:d}
\end{aligned}\;\right)\circ\left(\;
\begin{aligned}
&x_1\\
&T_2\\
&x_{3:d}
\end{aligned}\;\right)\circ\cdots\circ \left(\;
\begin{aligned}
&x_{1:d-2}\\
&T_{d-1}\\
&x_d\\
\end{aligned}\;\right)\circ\left(\;
\begin{aligned}
&x_{1:d-1}\\
&T_{d}\\
\end{aligned}\;\right)(\bm{x}).
\end{equation}

The KR rearrangement inspires the reduction of active dimensions as the simple bijections in \eqref{com_no_time} progress. In other words, some dimensions remain unchanged after certain transformations. On the other hand, coupling layers have demonstrated strong capabilities both theoretically \cite{teshima2020coupling} and in practice \cite{dinh2014nice,dinh2016density}, which motivates us to incorporate them when constructing bijective mappings. To this end, we relax the strictly triangular structure and instead adopt a more flexible design that combines a composition of block-triangular transformations with a transformation constructed from a series of coupling layers that act on the full input. Each block-triangular transformation is constructed by concatenating a composition of coupling layers with an identity mapping. For simplicity, let's first consider the invertible mapping on \([-1,1]^d\).
Let \(\bm{x}=\left(\left(\bm{x}^{(1)}\right)^{\mathrm{T}},\left(\bm{x}^{(2)}\right)^{\mathrm{T}}, \dots, \left(\bm{x}^{(K)}\right)^{\mathrm{T}}\right)^{\mathrm{T}}\)
be a partition of $\bm{x}$, where $\bm{x}^{(i)}=\left(x_1^{(i)},\dots,  x_{d_i}^{(i)}\right)^{\mathrm{T}}$ with $1\leq K\leq d$, $1\leq d_i\leq d$ and $\sum_{i=1}^Kd_i=d$. Denote \(\bm{x}^{(i:j)}=\left(\left(\bm{x}^{(i)}\right)^{\mathrm{T}},\left(\bm{x}^{(i+1)}\right)^{\mathrm{T}}, \dots, \left(\bm{x}^{(j)}\right)^{\mathrm{T}}\right)^{\mathrm{T}}\).
The B-KRnet is defined as follows,
\begin{equation}
\label{eqn:B_KRnet_stru}
\bm{z}=f_{\text{KR}}(\bm{x})=\left(\;\begin{aligned}
&\tilde{f}_1\\
&\bm{x}^{(2:K)}\\
\end{aligned}\;\right)\circ\left(\;
\begin{aligned}
&\tilde{f}_2\\
&\bm{x}^{(3:K)}\\
\end{aligned}\;\right)\circ\cdots\circ \left(\;
\begin{aligned}
&\tilde{f}_{K-1}\\
&\bm{x}^{(K)}\\
\end{aligned}\;\right)\circ\tilde{f}_K(\bm{x}),
\end{equation}
where \(\tilde{f}_k:[-1,1]^{\sum^k_{i=1}d_i}\to[-1,1]^{\sum^k_{i=1}d_i}\) is a mapping  of \(\bm{x}^{(1:k)}\) and \(\tilde{f}_k\) can be constructed by several simple bijections.
Thus B-KRnet is mainly constructed by two loops: outer loop and inner loop, where the outer loop has $K$ stages, corresponding to the $K$ mappings in equation \eqref{eqn:B_KRnet_stru}, and the inner loop means that each $\tilde{f}_k$ consists of $l_k$ coupling layers as shown later in equation \eqref{eqn:CDF_composition}.
Namely, 
\begin{itemize}
	\item Outer loop. We rewrite equation \eqref{eqn:B_KRnet_stru} for convenience in the following description:
 \begin{equation}\bm{z}=\left(\begin{array}{c}f^{outer}_{[K]}\\
	\bm{x}^{(2:K)}\end{array}\right)\circ\cdots\circ\left(\begin{array}{c}f^{outer}_{[2]}\\
	\bm{x}^{(K)}\end{array}\right)\circ f^{outer}_{[1]}(\bm{x}),
 \end{equation}
 where \(f_{[k]}^{outer}=\tilde{f}_{K-k+1}\). 
 Denote
	\begin{align}
	\bm{x}_{[0]}&=\bm{x},\\
	\bm{x}_{[1]}&=f^{outer}_{[1]}(\bm{x}_{[0]}),\\
	\bm{x}_{[k]}&=\left(\begin{array}{c}
 f^{outer}_{[k]}\left(\bm{x}^{(1:K-k+1)}_{[k-1]}\right)\\
 \bm{x}_{[k-1]}^{(K-k+2:K)}\end{array}\right), \quad k=2,\dots, K,
	\end{align}
 where each $\bm{x}_{[k]}$ shares the same partition as $\bm{x}$ for $k=1,\dots, K$. The $k$-th partition remains unchanged after $K-k+1$ stage, where $k=1,\dots, K$. It can be considered as a squeezing operation that deactivates some dimensions as the outer loop proceeds.
	\item Inner loop. 
	\begin{align}
 \label{eqn:CDF_composition}
	f^{outer}_{[k]}&= L_{\text{CDF},~[k,l_k]}\circ\cdots\circ L_{\text{CDF},~[k,1]},\quad k=1,\dots, K,
	\end{align}
	where $L_{\text{CDF}, ~[k,i]}: [-1,1]^{\sum_{i=1}^{K-k+1}d_i}\to [-1,1]^{\sum_{i=1}^{K-k+1}d_i}$ 
 is a CDF coupling layer defined in \cref{sec:CDF}. Note that as the outer loop progresses, the input dimension of \(f^{outer}_{[k]}\) gradually decreases, encouraging us to reduce the depth $l_k$. Also note that if $d_1=1$, equation \eqref{eqn:CDF_composition} holds for $k=1,\ldots, K-1$  
 while $f_{[K]}^{outer}$ becomes an identity (see \cref{rmk:1st_d}).
 \end{itemize}

\subsubsection{CDF coupling layer}\label{sec:CDF}
We introduce the crucial layer of B-KRnet, which defines a componentwise nonlinear invertible mapping on $[-1,1]^l$. We start with a one-dimensional nonlinear mapping $F(s):[-1,1]\to[0,1]:$
\begin{equation}
\label{eqn:CDF_motivation}
F(s)=\int_{-1}^sp(t)dt,
\end{equation}
where $p(s)$ is a probability density function (PDF) defined on $[-1,1]$ and $F(s)$ is nothing but the CDF for $p(s)$. Let $-1= s_0< s_1< \dots< s_n=1$ be a mesh of the interval $[-1,1]$ with \(n\) elements $[s_{i-1},s_{i}]$, $i=1, \dots, n$ on which we define a piecewise PDF
\begin{equation}
p(s)=\frac{w_{i+1}-w_i}{h_i}(s-s_i)+w_i,\quad \forall s\in[s_i, s_{i+1}],
\end{equation}
with $p(s_i)=w_i\geq0$, $h_i=s_{i+1}-s_i$. Then $F(s)$ can be written as 
\begin{equation}
\label{eqn:CDF_F}
F(s)=\frac{w_{i+1}-w_i}{2h_i}(s-s_i)^2+w_i(s-s_i)+\sum\limits_{k=0}^{i-1}\frac{w_k+w_{k+1}}{2}(s_{k+1}-s_k), \quad \forall s\in[s_i, s_{i+1}].
\end{equation}

The inverse of $F(s)$ can be computed efficiently, which is a root of a quadratic polynomial. 
Letting $q_0=0$, $q_i=\sum\limits_{k=0}^{i-1}\frac{w_k+w_{k+1}}{2}(s_{k+1}-s_k)$, $i=1, \dots, n$, and solving 
\begin{equation}
\frac{w_{i+1}-w_i}{2h_i}(s-s_i)^2+w_i(s-s_i)=q-q_i, \text{ if } q\in[q_i,q_{i+1}],
\end{equation}
we have
\begin{equation}
\label{eqn:CDF_inverse}
F^{\mathrm{-1}}(q)=s=s_i+\frac{-w_i+\sqrt{w_i^2+2(w_{i+1}-w_i)(q-q_i)/h_i}}{{(w_{i+1}-w_i)}/{h_i}}.
\end{equation}
In case that the denominator is too small, we rewrite the equation \eqref{eqn:CDF_inverse} as
\begin{equation}
F^{\mathrm{-1}}(q)=s_i +\frac{2(q-q_i)}{w_i+\sqrt{w_i^2+2(w_{i+1}-w_i)(q-q_i)/h_i}}.
\end{equation}

Moreover, \(\tilde{F}(s)=2F(s)-1\) defines the bijection from \([-1,1]\to[-1,1]\). We are now ready to define a componentwise mapping $\bm{F}(\bm{y}):[-1,1]^m\rightarrow [-1,1]^m$ for any nonnegative \(m\in\mathbb{N}\) based on \(\tilde{F}\) by letting
\begin{equation}
\label{eqn:vector_quadratic}
F_i(y_i)=\tilde{F}(y_i;\bm{\theta}_i),
\end{equation}
where $\bm{\theta}_i$ consists of model parameters for the $i$-th dimension. Adopting the concept of affine coupling mapping, we define the following invertible bijection for
$\bm{y}=(\bm{y}_{1}^{\mathrm{T}}, \bm{y}_{2}^{\mathrm{T}})^{\mathrm{T}}\in [-1,1]^l$ with $\bm{y}_{1}\in\mathbb{R}^{m}$ and $\bm{y}_2\in\mathbb{R}^{l-m}$,
\begin{equation}
\label{eqn:CDF_layer}
\left(\begin{array}{c}
\hat{\bm{y}}_1\\
\hat{\bm{y}}_2
\end{array}\right)=L_{\text{CDF},\bm{\theta}}\left(\begin{array}{c}
\bm{y}_1\\
\bm{y}_2
\end{array}\right), \quad
\left\{\enspace\begin{aligned}
& \hat{\bm{y}}_{1}=\bm{y}_{1},\\
& \hat{\bm{y}}_{2}=\bm{F}(\bm{y}_{2}; \bm{\theta}(\bm{y}_{1})),
\end{aligned}\right.
\end{equation}
where the model parameters
$\bm{\theta}=\left({\bm{s}}_1^{\mathrm{T}}, \dots, {\bm{s}}_{n-1}^{\mathrm{T}}, {\bm{w}}_0^{\mathrm{T}}, \dots, {\bm{w}}_{n}^{\mathrm{T}}\right)^{\rm{T}}\in\mathbb{R}^{2n(l-m)}$ only depends on $\bm{y}_1$ with $\bm{s}_i, \bm{w}_i\in\mathbb{R}^{l-m}.$ 
These parameters can be built by formatting the output of a neural network.
Define the following neural network
\begin{equation}
\label{eqn:theta_NN}
\left(\hat{\bm{s}}_1^{\mathrm{T}}, \dots, \hat{\bm{s}}_{n-1}^{\mathrm{T}}, \hat{\bm{w}}_0^{\mathrm{T}}, \dots,\hat{\bm{w}}_n^{\mathrm{T}}\right)^{\rm{T}}\in\mathbb{R}^{2n(l-m)}={\text{NN}}(\bm{y}_{1}). 
\end{equation}
Let
\begin{equation}
\label{eqn:params_of_CDF2}
\left\{\enspace\begin{aligned}
&\bm{s}_1=-1+\frac{2}{n}(1+\beta_1\tanh(\hat{\bm{s}}_1)),\\
&\bm{s}_{i+1}=\bm{s}_i+\left(\frac{1-\bm{s}_i}{n-i}\right)\left(1+\beta_{i+1}\tanh(\hat{\bm{s}}_{i+1})\right),\\
&\bm{w}_i=(1+\gamma_i\tanh(\hat{\bm{w}}_i))/C,
\end{aligned}\right.
\end{equation}
where $C$ is a normalization constant such that the total probability is equal to one. $\beta_i$ and $\gamma_i$ are chosen such that $-1<\bm{s}_1<\dots<\bm{s}_{n-1}<1$ and $\bm{w}_i>0.$ Let $\beta_1=\frac{65}{66}, \beta_i=0.97$ for \(i\geq 2\) and $\gamma_i=0.99$ in practice. 

The bijection given by equation \eqref{eqn:CDF_layer} is referred to as the CDF coupling layer since the mapping is based on a CDF. On the other hand, the mapping \eqref{eqn:CDF_layer} only updates the second part $\bm{y}_2$, so another CDF coupling layer is needed for a complete
update. The next CDF coupling layer can be defined by switching the roles of the two input parts:
\begin{equation}
\label{eqn:CDF_layer_x1}
\left(\begin{array}{c}
\hat{\hat{\bm{y}}}_1\\
\hat{\hat{\bm{y}}}_2
\end{array}\right)=\tilde{L}_{\text{CDF},\tilde{\bm{\theta}}}\left(\begin{array}{c}\hat{\bm{y}}_1\\
\hat{\bm{y}}_2
\end{array}\right),\quad
\left\{\enspace\begin{aligned}
& \hat{\hat{\bm{y}}}_{1}=\tilde{\bm{F}}(\hat{\bm{y}}_{1}; \bm{\tilde{\theta}}(\hat{\bm{y}}_{2})),\\
& \hat{\hat{\bm{y}}}_{2}=\hat{\bm{y}}_{2},
\end{aligned}\right.
\end{equation}
where \(\tilde{F}:[-1,1]^m\to[-1,1]^m\) and \(\bm{\tilde{\theta}}\in\mathbb{R}^{2nm}\) is defined similarly to the equation \eqref{eqn:theta_NN} and \eqref{eqn:params_of_CDF2}. The above transformation \eqref{eqn:CDF_layer_x1} updates the first part  $\hat{\bm{y}}_1$ while keeping $\hat{\bm{y}}_2$ unchanged.

Therefore, \(L_{\text{CDF},[k,i]}: [-1,1]^{\sum_{j=1}^{K-k+1}d_j}\to[-1,1]^{\sum_{j=1}^{K-k+1}d_j}\) in equation \eqref{eqn:CDF_composition} can be given as below:
\begin{equation}
\label{eqn:L_CDF_layer}
\begin{aligned}
&\bm{y}_1=\bm{x}^{(1:K-k)}_{[k-1]},\; \bm{y}_2=\bm{x}^{(K-k+1)}_{[k-1]},\\
&L_{\text{CDF},[k,i]}\left(\begin{array}{c}
\bm{x}^{(1:K-k)}_{[k-1]}\\
\bm{x}_{[k-1]}^{(K-k+1)}
\end{array}\right)=
\left\{\enspace\begin{aligned}
& {L}_{\text{CDF}, {\bm{\theta}}^{[k, i]}}\left(\begin{array}{c}
\bm{y}_1\\
\bm{y}_{2}
\end{array}\right) , \text{ if } i \text{ is odd};
\\
& \tilde{L}_{\text{CDF},\tilde{\bm{\theta}}^{[k, i]}}\left(\begin{array}{c}
\bm{y}_1\\
\bm{y}_2
\end{array}\right), \text{ if } i \text{ is even}.
\end{aligned}\right.
\end{aligned}
\end{equation}

The definition of the CDF coupling layer and the specific choices given by equation \eqref{eqn:params_of_CDF2} are based on the following considerations: First, the CDF coupling layer can be relatively simple, meaning that we do not have to consider a
very fine mesh on $[-1,1]^d$. It is seen that in the regular KRnet, the affine coupling layer simply defines a linear mapping for the data to be updated. 
Second, in equation \eqref{eqn:params_of_CDF2}, a tangent hyperbolic function is used to control the conditioning of the mapping. This ensures that both the adaptive mesh grids $s_i$ and the probability density function (PDF) on grids $w_i$ do not vary too much, which would otherwise make the problem ill-posed.
In addition, CDF mapping is a natural way to build the transport between PDFs by substituting a uniformly distributed random vector \(\bm{Z}\) into the explicit construction of KR rearrangement \cite{zech2022sparse1}.

\begin{remark}\label{rmk:1st_d}
    Note that the input dimension of the coupling layer can not be less than 2. 
    If \(\bm{x}^{(1)}=x_1\in\mathbb{R}\), the CDF coupling layer is not applicable to define an invertible mapping of $x_1$, meaning that \(f_{\text{KR}}\) consists of \(K-1\) bijections $f_{[k]}^{outer}$, i.e., 
    \begin{equation}\label{eqn:outer_K}
   \bm{z}=\left(\begin{array}{c}f^{outer}_{[K-1]}\\
\bm{x}^{(3:K)}\end{array}\right)\circ\cdots\circ\left(\begin{array}{c}f^{outer}_{[2]}\\
	\bm{x}^{(K)}\end{array}\right)\circ f^{outer}_{[1]}(\bm{x}).
  \end{equation}
 According to the KR rearrangement \cite{zech2022sparse1}, \(f^{outer}_{[K]}(x_{[K-1]}^{(1)})=2F(x_{[K-1]}^{(1)})-1\) if $z_1$ is uniform on $[-1,1]$. Note that normalizing flows introduce another convergence path to map $\bm{x}$ to $\bm{z}$, meaning that the output of $f_{[K-1]}^{outer}$ should converge to the corresponding components of $\bm{z}$. In other words, \(\bm{x}^{(1)}_{[K-1]}\) is close to a uniform random sample on \([-1,1]\), which indicates that
 \(f^{outer}_{[K]}(\cdot)=2F(\cdot)-1\) is close to an identity operator. In practice, we may let $f_{[K]}^{outer}$ be an identity mapping as in equation \eqref{eqn:outer_K}. 
 Another choice is to apply a trainable mapping to approximate $F(\cdot)$ for further refinement, which is not considered in this work.
 
\end{remark}

\subsubsection{Affine linear mapping}
In the previous sections, a transformation \(f_{\text{KR}}\) with descending active transformation dimensions is defined, mapping from \([-1,1]^d\) to \([-1,1]^d\). To tackle the general product domain $\Omega=\prod_{i=1}^d[a_i, b_i]$, a linear mapping is introduced, converting $\Omega$ to $[-1,1]^d$:
\begin{equation}
\bm{y}=\hat{\bm{a}}\odot \bm{x} + \hat{\bm{b}},
\label{eqn:affine_linear_layer}
\end{equation}
where $\odot$ denotes the Hadamard product or componentwise product, $\hat{a}_i=2/(b_i-a_i)$ and  $\hat{b}_i=-(b_i+a_i)/(b_i-a_i)$. 
By rescaling the data to the domain $[-1, 1]^d$, we take advantage of the symmetry around the origin, which aligns with the initialization of the neural network model parameters around zero.

\begin{figure}[ht]
	\centering 
	\includegraphics[scale=0.24]{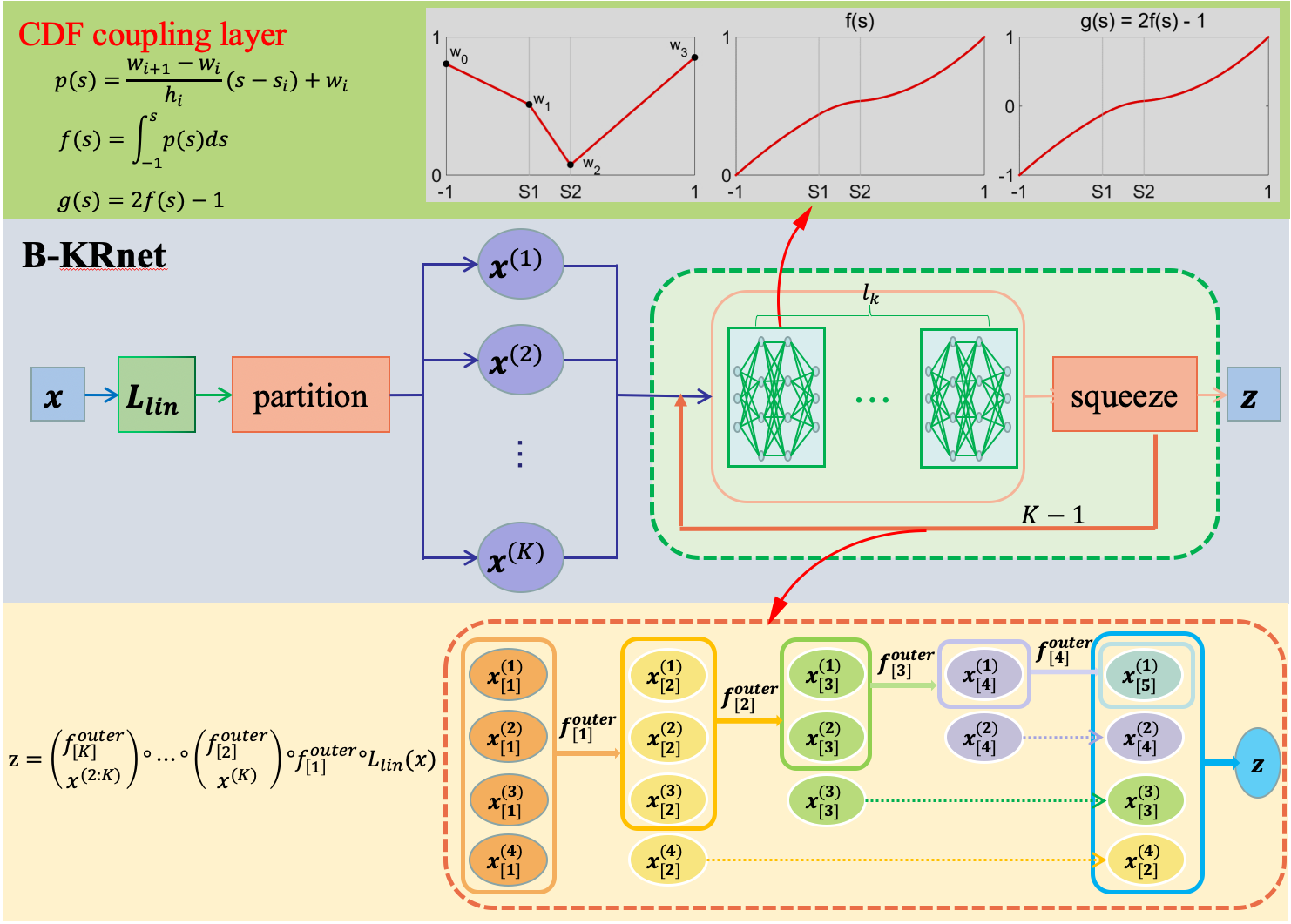}
	\caption{The schematic of B-KRnet. The input is passed to a linear layer that maps the input into $[-1,1]^d$, which is then partitioned into $K$ parts and passed to $K$ bijections $f^{outer}_{[k]}$. Each $f^{outer}_{[k]}$ consists of $l_k$ CDF coupling layers. }
	\label{fig:bd_KR_struc}
\end{figure}

The detailed schematic of our B-KRnet is shown in \cref{fig:bd_KR_struc}. 
The structure of B-KRnet is similar to KRnet \cite{tang2020deep}, which is defined on an unbounded domain. The scale and bias layer of KRnet has been removed in B-KRnet since the domain is bounded. Another difference is that B-KRnet, based on the CDF coupling layer, results in a piecewise linear PDF model. If such a PDF model is used to approximate a second-order differential equation, e.g., the Fokker-Planck equation, a mixed formulation must be used, in other words, the gradient of the solution should be approximated by an additional neural network.

\subsection{The complexity of B-KRnet}
In this part, we discuss the degrees of freedom (DOFs) of B-KRnet.  
It consists of \(K\) bijections $f^{outer}_{[k]}$ when \(d_1\neq 1\); otherwise it consists of \(K-1\) bijections.
Each $f^{outer}_{[k]}$ includes $l_k$ CDF coupling layers and all trainable parameters come from CDF coupling layers. We denote \(\text{DOFs}(f^{outer}_{[K]})=0\) for \(d_1=1\). Then the number of DOFs of B-KRnet is
\begin{equation}
\label{eqn:DOF}
\text{DOFs}(p_{\text{B-KRnet},\bm{\theta}})=\sum_{k=1}^{{K}}\text{DOFs}(f^{outer}_{[k]})=\sum_{k=1}^{{K}}\sum_{i=1}^{l_k}\text{DOFs}(L_{\text{CDF},[k,i]}).
\end{equation}

Assume that the neural network \eqref{eqn:theta_NN} for a CDF coupling layer of \(f^{outer}_{[k]}\) is fully connected and has {$D_k$} hidden layers, each with {$W_k$} neurons.
{For \(1\leq k\leq K-1\), \(L_{\text{CDF}, [k,i]}\)\text{is defined on }
\([-1,1]^{\sum^{K-k+1}_{i=1}d_i}\). 
The input of the \(L_{\text{CDF}, [k,i]}\) is divided into two parts \(\bm{y}_1\) and \(\bm{y}_2\) with \(\bm{y}_1=\bm{x}^{(1:K-k)}_{[k-1]},\; \bm{y}_2=\bm{x}^{(K-k+1)}_{[k-1]}\).
The DOFs of two adjacent CDF layers in \(f^{outer}_{[k]}\) with \( 1\leq k<K\) are
\begin{equation}
\label{eqn:DOF_two_CDFs_k}
\begin{aligned}
&\text{DOFs}(L_{\text{CDF},[k,i]})+\text{DOFs}(L_{\text{CDF},[k,i+1]})\\
=&~2(D_k-1)W_k^2+\left((2n+1)\sum^{K-k+1}_{i=1}d_i+2D_k\right)W_k+2n\sum^{K-k+1}_{i=1}d_i.
\end{aligned}
\end{equation}
If \(d_1>1\), the DOFs of two adjacent CDF layers in \(f^{outer}_{[K]}\) are
\begin{equation}
\label{eqn:DOF_two_CDFs_K}
\begin{aligned}
&\text{DOFs}(L_{\text{CDF},[K,i]})+\text{DOFs}(L_{\text{CDF},[K,i+1]})\\
=&~2(D_K-1)W_K^2+\left((2n+1)d_1+2D_k\right)W_k+2nd_1.
\end{aligned}
\end{equation}
For simplicity, let $K=d$, \(d_k=1\) for \(k=1,\dots, K\). 
Substituting equation \eqref{eqn:DOF_two_CDFs_k} 
into equation \eqref{eqn:DOF}, we can obtain the total number of model parameters is
 \begin{equation*}
\text{DOFs}(p_{\text{B-KRnet}})=\sum_{k=1}^{d-1}\frac{l_k}{2}\big(2(D_k-1)W_k^2+((2n+1)(d-k+1)+2D_k)W_k+2n(d-k+1)\big).
 \end{equation*}
{Let $p_{\text{NF}}$ be a normalizing flow with a regular structure, i.e., a sequence of coupling CDF layers subject to a fixed partition of the input. For the same $D_k$, $W_k$ and total depth, we have}
\begin{equation}
    \text{DOFs}(p_{\text{NF}})-\text{DOFs}(p_{\text{B-KRnet}})=\sum_{k=1}^{d-1}\frac{(k-1)l_k}{2}((2n+1)W_k+2n).
\end{equation}
}
\subsection{The capability of B-KRnet}
In this section, we briefly discuss the capability of B-KRnet. KR rearrangement guarantees the existence of a transformation \(T\) such that \(T_{\#}\rho=\pi\).  Meanwhile, under certain conditions, the discrepancy between transports defines the difference in the corresponding pushforward measures \cite{butler2018convergence, butler2022p, sagiv2022spectral, sagiv2019wasserstein, baptista2025approximation}. 
Consequently, an approximation of the transformation yields a corresponding approximation of the pushforward measure, which motivates us to approximate \(T\) rather than directly approximating the unknown distribution.
Note that the triangular structure of the KR rearrangement is a special case of the structure used in B-KRnet. Moreover, B-KRnet can be considered a direct approximation for KR rearrangement by letting \(l_k=1\), \(K=d\), and \(n\) go to infinity. The corresponding CDF coupling layer \eqref{eqn:CDF_layer} is then tasked with approximating the component \(T_k\) used in the KR rearrangement \eqref{eqn:KR_rearrangement}.

On the other hand, \cite{teshima2020coupling} establishes the universality of unbounded normalizing flows. They reduce the universality of the pushforward measures to that of single-coordinate transformations. A constructive proof then demonstrates that affine coupling layers can approximate these single-coordinate transformations. Following a similar line of reasoning, we can also reduce the problem to the universality of single-coordinate transformations in the compact domain setting. Since piecewise quadratic functions with sufficiently large \(n\) can approximate arbitrary functions, it is possible to approximate the corresponding PDF even without employing a coupling structure. The convergence rate in this case can be quantified using classical Lagrange interpolation error estimates. However, the role of coupling layers in the universality approximation remains an open question and presents a promising direction for future investigation.

\section{Application to density estimation}\label{sec:density_estimation}
Let $\bm{X}$ be a $d$-dimensional random variable with an unknown density function $p_{\bm{X}}(\cdot)$ defined on a bounded domain $\Omega$. Given the observations $\{\bm{x}_i\}_{i=1}^N$ that are drawn independently from $p_{\bm{X}}$, we aim to construct a generative model as a surrogate for $p_{\bm{X}}$. 
We train a B-KRnet model 
by minimizing the cross entropy between the data distribution and $p_{\text{B-KRnet}}$.
\begin{equation}
\label{eqn:LOSS_density_est}
\begin{aligned}
\mathcal{L}_{data}&=-\frac{1}{N}\sum_{i=1}^N\log(p_{\text{B-KRnet}}(\bm{x}_i)),\\
&=-\frac{1}{N}\sum_{i=1}^N\log\big(p_{\bm{Z}}\left(f_{\text{B-KRnet}}(\bm{x})\right)\big)+\log\left(\left| \nabla_{\bm{x}}f_{\text{B-KRnet}}(\bm{x})\right|\right),
\end{aligned}
\end{equation}
which is equivalent to maximizing the likelihood. Noting that the Kullback-Leibler (KL) divergence between $p_{\bm{X}}$ and the density model $p_{\text{B-KRnet}}$ can be written as
\begin{equation}
\label{eqn:KL_divergence}
\begin{aligned}
D_{\text{KL}}(p_{\bm{X}}\mid\mid p_{\text{B-KRnet}})
&=-H(p_{\bm{X}})+H(p_{\bm{X}},p_{\text{B-KRnet}}),
\end{aligned}
\end{equation}
and 
\begin{equation}
H(p_{\bm{X}},p_{\text{B-KRnet}})\approx -\frac{1}{N}\sum_{i=1}^N\log(p_{\text{B-KRnet}}(\bm{x}_i))=\mathcal{L}_{data},
\end{equation}
where $\bm{x}_i$ are drawn from $p_{\bm{X}}$, $H(p_{\bm{X}})$ is the entropy of $p_{\bm{X}}$ and $H(p_{\bm{X}},p_{\text{B-KRnet}})$ is the cross entropy between $p_{\bm{X}}$ and $p_{\text{B-KRnet}}$. We are trying to minimize $D_{\text{KL}}(p_{\bm{X}}\mid\mid p_{\text{B-KRnet}})$ 
since \(H(p_{\bm{X}})\) is a constant that depends on the target distribution.

To evaluate the accuracy of $p_{\text{B-KRnet}}$, we approximate the relative KL divergence by applying the Monte Carlo method on a validation set, i.e.,
$$\frac{D_{\mathrm{KL}}(p_{\bm{X}}\mid\mid p_{\text{B-KRnet}})}{H(p_{\bm{X}})}\approx \frac{\sum_{i=1}^{N_v}\big(\log(p_{\bm{X}}(\bm{x}_i)-\log p_{\text{B-KRnet}}(\bm{x}_i))\big)}{-\sum_{i=1}^{N_v}\log p_{\bm{X}}(\bm{x}_i)}.$$
Here $\{\bm{x}_i\}_{i=1}^{N_v}$ are drawn from the ground truth $p_{\bm{X}}$.  
Unless otherwise specified, the hyperbolic tangent function (tanh) is used as the activation function. The neural network introduced in \eqref{eqn:theta_NN} is a feed-forward neural network consisting of two hidden layers, each comprising \(32\) neurons. For the training procedure, we apply the Adam optimizer \cite{kingma2014adam}. All numerical tests are implemented with Pytorch.
\subsection{Example 1: Annulus}
Consider an annulus,
\begin{equation}
\left\{\enspace\begin{aligned}
&x=r\cos\theta,\\
&y=r\sin\theta,\\
&\theta\sim \text{Uniform}[0, 2\pi],\quad r\sim f(r)=\frac{1}{r}1_{[1, \,e]},
\end{aligned}\right.
\end{equation}
where $1_{[1, \, e]}$ is an indicator function with $1_{[1, \, e]}(r)=1$ if $r\in {[1,\, e]};\;0$ otherwise. $r$ can be sampled via
$\log r\sim \text{Uniform}[0,1].$
That is to say, given a sample $\tilde{r}$ generated from a uniformly distributed on $[0,1]$, then $r=\exp(\tilde{r})$ obeys the PDF $f(r)$. 
The ground truth PDF $p(x,y)$ has a form
\begin{equation}
p(x,y)=p(r,\theta)\left|\frac{\partial(x,y)}{\partial(r,\theta)}\right|^{-1}=\frac{1}{2\pi r^2}=\frac{1}{2\pi(x^2+y^2)},\quad 
 1\leq x^2+y^2\leq e^2.
\end{equation}
For this case, the entropy of $p$ can be computed analytically,
\begin{equation*}
H(p)=-\int_1^e\int_0^{2\pi}\frac{1}{2\pi r}\log\left(\frac{1}{2\pi r}\right)\mathrm{d}\theta \mathrm{d}r=\frac{1}{2}+\log(2\pi).
\end{equation*}

The number of training points is $2\times10^4$. For the B-KRnet, we take $8$ CDF coupling layers and apply \(3\) subintervals along each dimension. 
The Adam method is applied with an initial learning rate of $0.001$ and 
a batch size of $4096$.
A total of $4000$ epochs are considered and a validation dataset with $2\times10^4$ samples is used for computing the relative KL divergence throughout the whole training process. 

\begin{figure}[h!]
	\centering
	\begin{minipage}[b]{0.28\linewidth}
		\includegraphics[height=3cm,width=3.8cm]{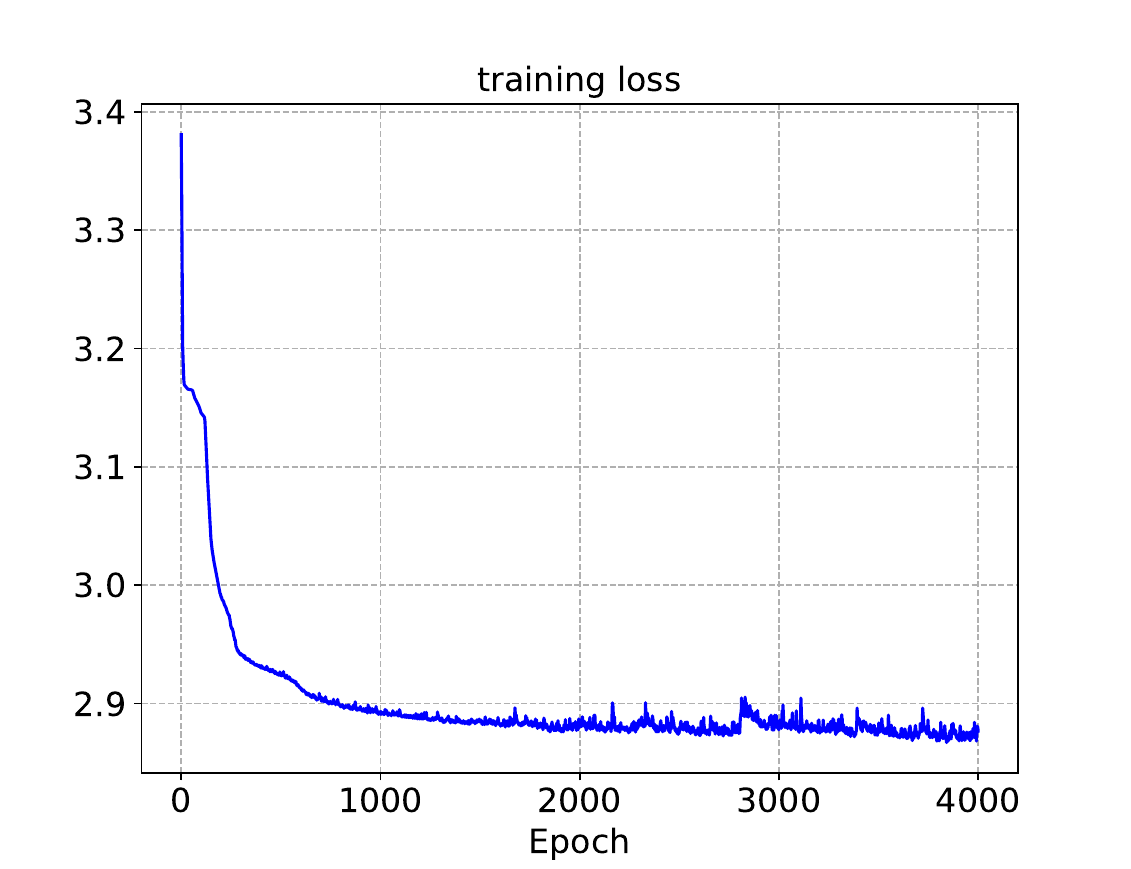}
	\end{minipage}
	\begin{minipage}[b]{0.28\linewidth}
		\includegraphics[height=3cm,width=3.8cm]{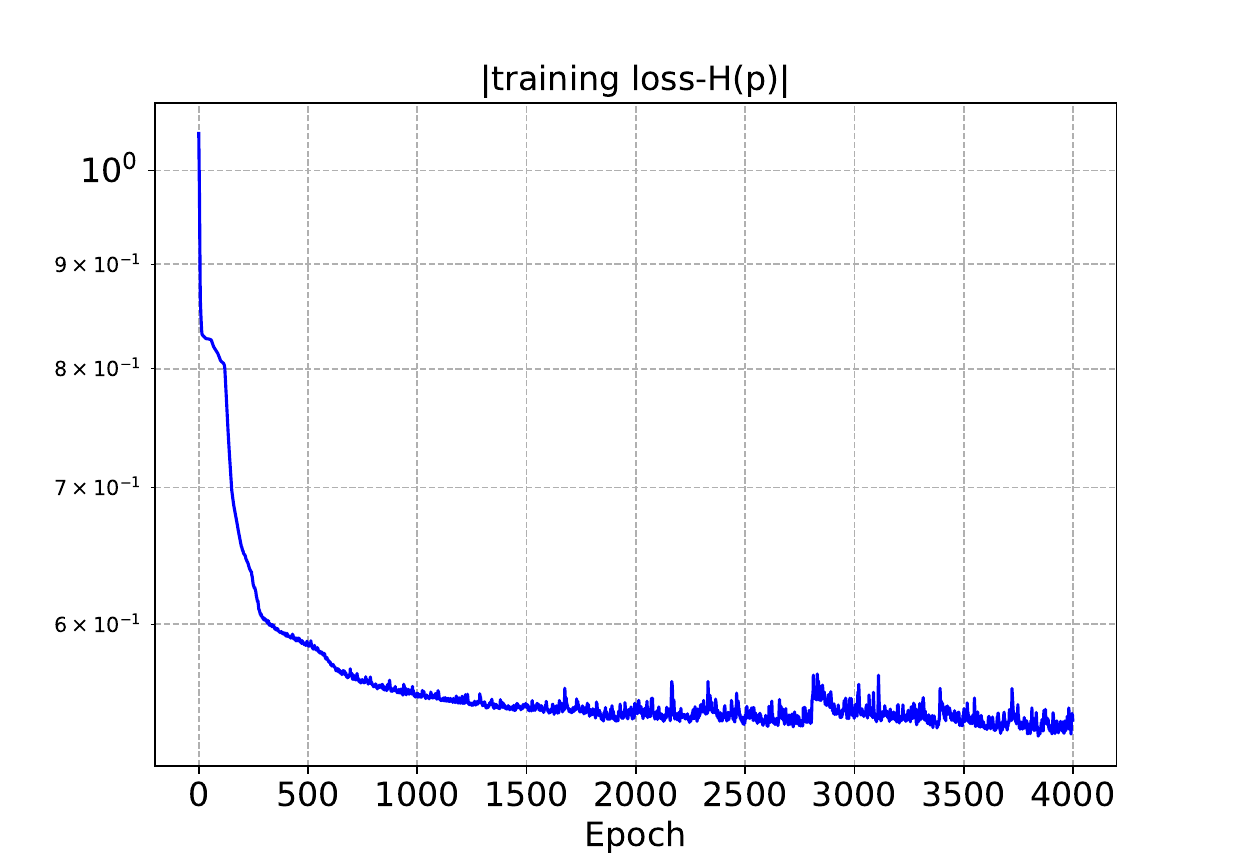}
	\end{minipage}
	\begin{minipage}[b]{0.28\linewidth}
		\includegraphics[height=3cm,width=3.8cm]{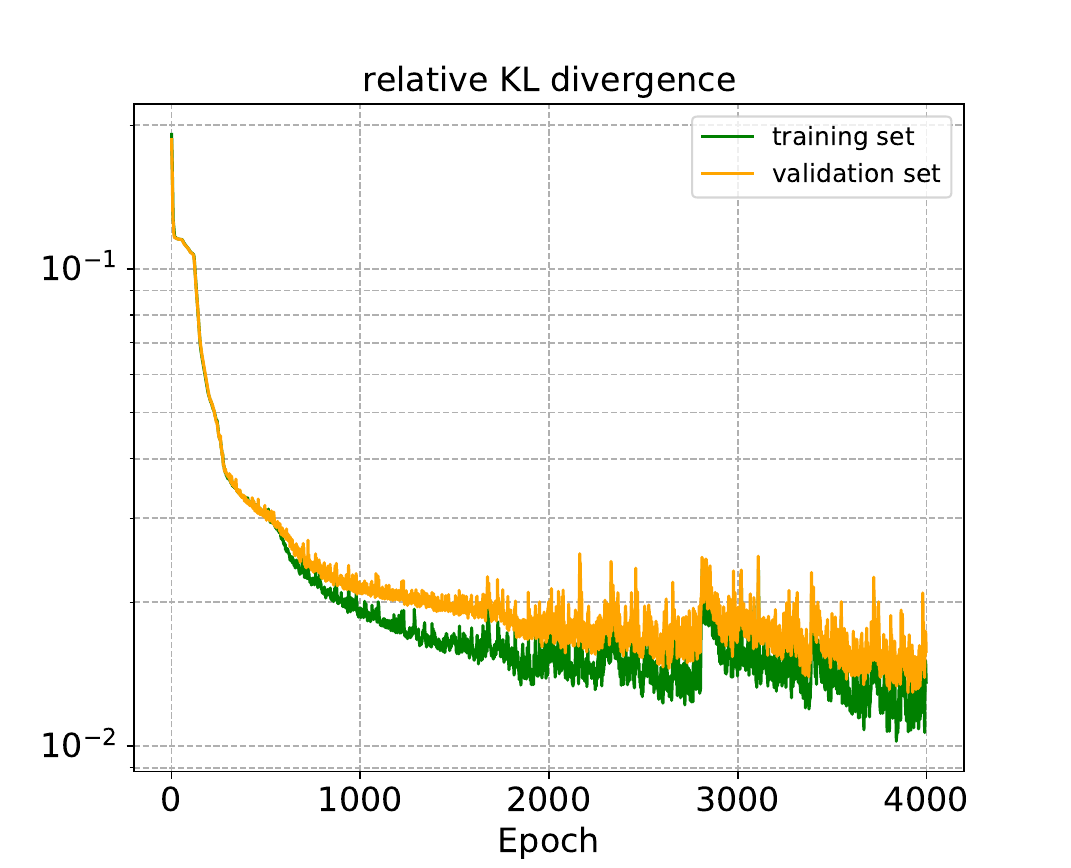}
	\end{minipage}
	\caption{Training procedure for the distribution on an annulus. Left: The training loss. Middle: The KL divergence on the training set. Right: The relative KL divergence on the training set and validation set.}
	\label{fig:circ_loss_kl}
\end{figure}

\begin{figure}[h!]
	\centering
	\begin{minipage}[b]{0.23\linewidth}
		\includegraphics[height=3cm,width=3cm]{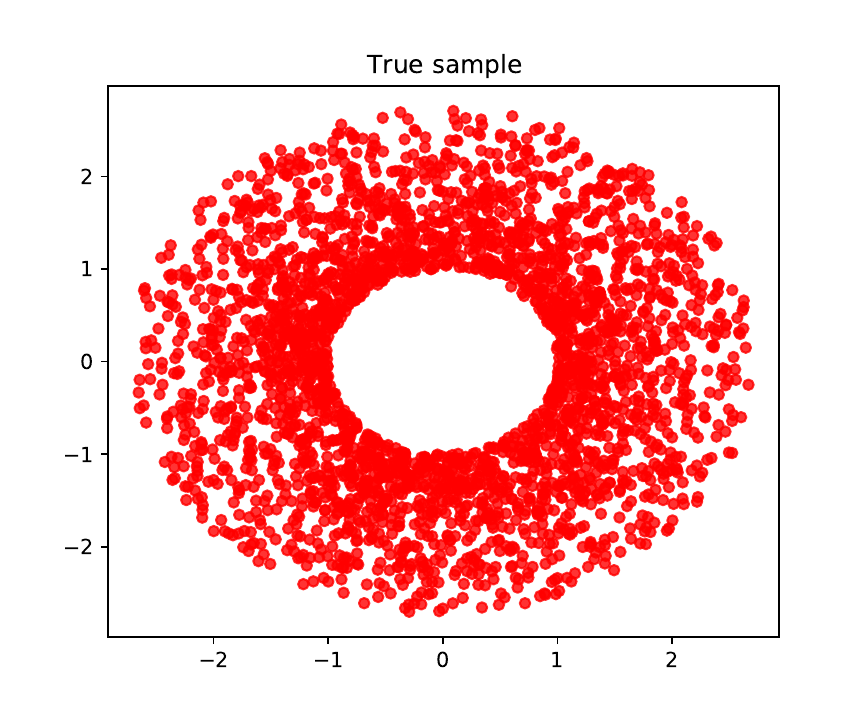}
	\end{minipage}
	\begin{minipage}[b]{0.23\linewidth}
		\includegraphics[height=3cm,width=3cm]{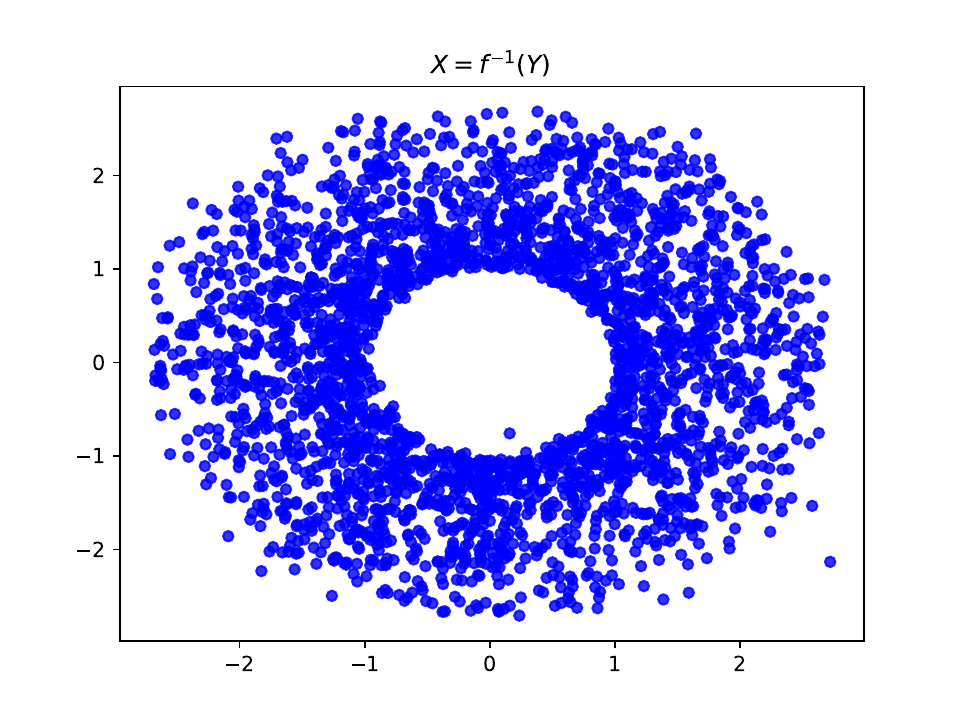}
	\end{minipage}
	\begin{minipage}[b]{0.23\linewidth}
		\includegraphics[height=3cm,width=3cm]{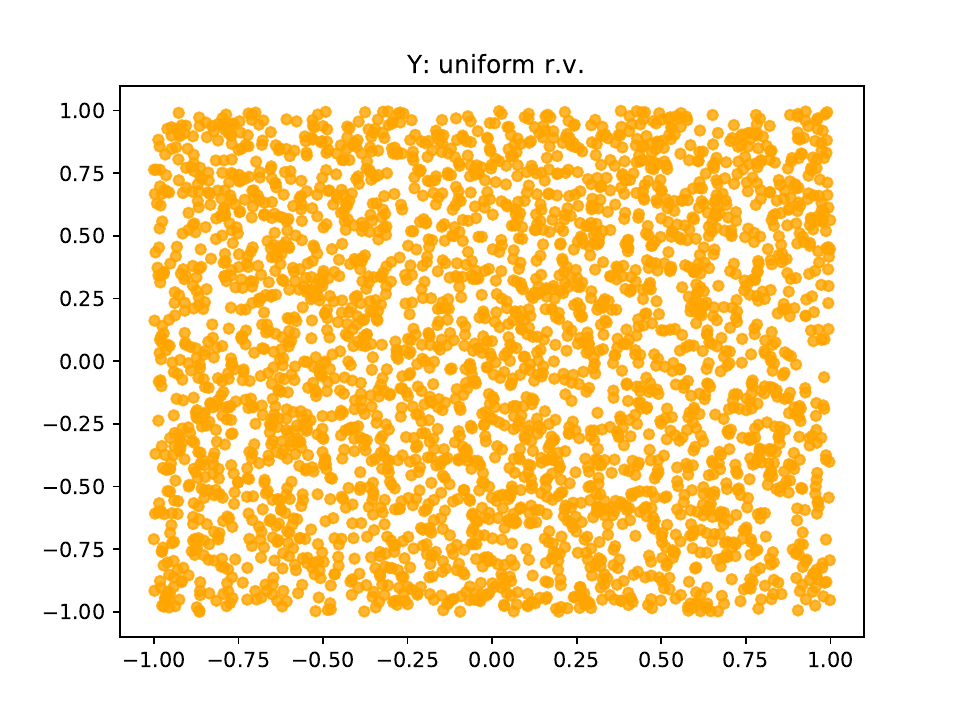}
	\end{minipage}
	\caption{Comparison between true samples and samples from B-KRnet for the distribution on an annulus. Left: True samples. Middle: Samples from B-KRnet. Right: Samples from the base distribution (uniform).}
	\label{fig:circ_sample}
\end{figure}

The training loss and the relative KL divergence are presented in \cref{fig:circ_loss_kl}. We also compare the samples from the true distribution and samples from the PDF model in \cref{fig:circ_sample}. The samples generated by B-KRnet agree very well with the true samples. In particular, the sharp boundary of $p(x,y)$ is well captured. 

\begin{figure}[h!]
\centering
\includegraphics[scale=0.25]{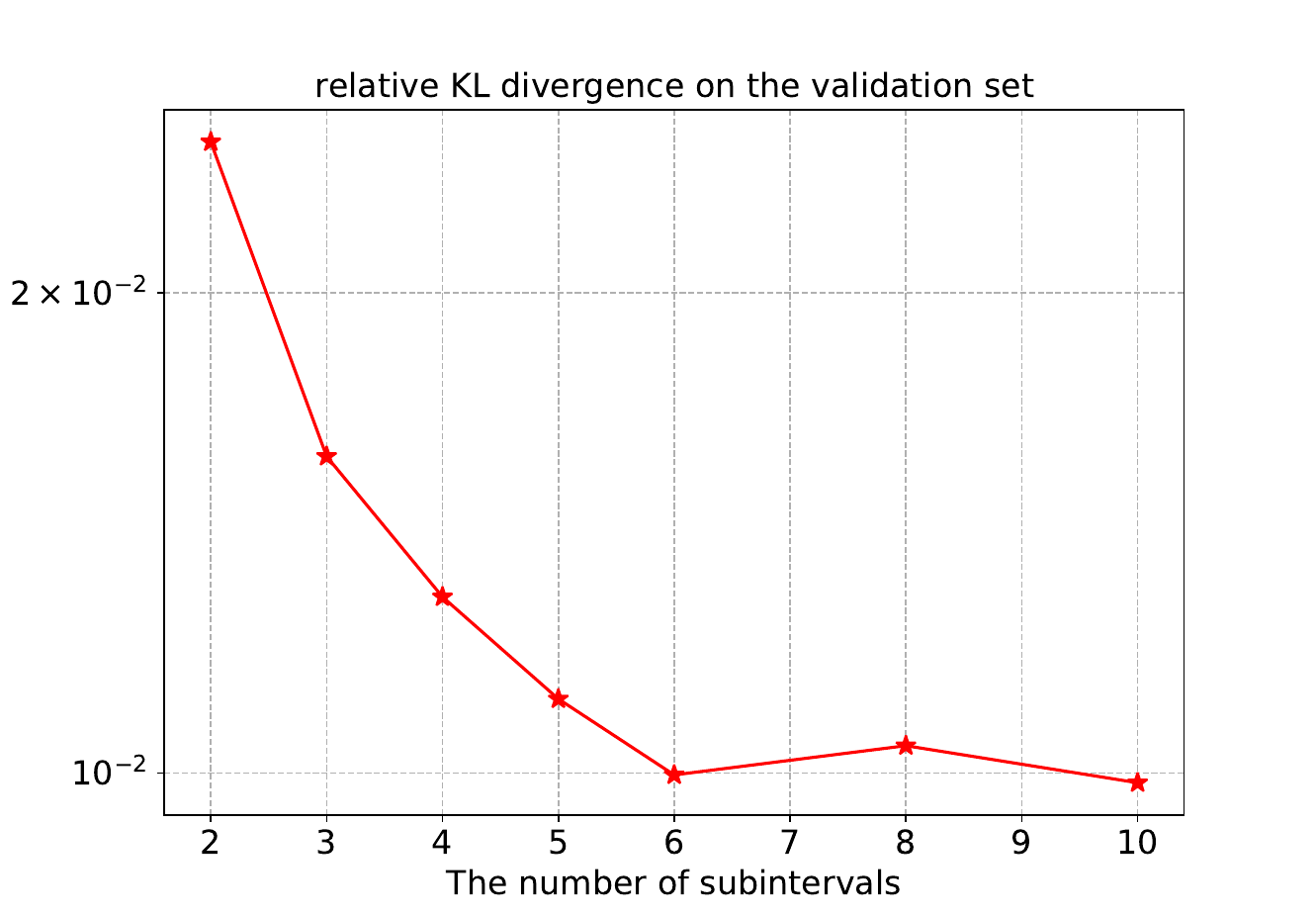}
\caption{The relative KL divergence for different numbers of subintervals.}
\label{fig:circ_vary_nodes}
\end{figure}

Moreover, we explore the effect of the number of subintervals used in the CDF layers. \Cref{fig:circ_vary_nodes} shows the relative KL divergence on the validation set for different numbers of subintervals. 
Increasing the number of subintervals improves performance, but only marginally, especially when the number of subintervals exceeds 6 in this case. 

\subsection{Example 2: Mixture of Gaussians}
We consider a mixture of Gaussians that is restricted to $[-1,1]^2$, i.e., 
\begin{equation}
p_{\bm{Y}}(\bm{y})=\frac{1}{6C}\sum\limits_{i=1}^6\text{Normal}(\bm{y}_i,\sigma^2\bm{I}_2)1_{[-1,1]^2}(\bm{y}),
\end{equation}
where $\bm{y}_i=(0.8\cos\frac{i\pi}{3}+0.3, 0.8\sin\frac{i\pi}{3}+0.3)$, $\sigma=0.15$, $\bm{I}_2$ is the \(2\times 2\) identity matrix, $\text{Normal}(\bm{\mu},\bm{\Sigma})$ represents a Gaussian distribution with mean \(\bm{\mu}\) and variance matrix \(\bm{\Sigma}\),
and $C$ is a normalization constant.
\begin{figure}[h!]
	\centering
	\begin{minipage}[b]{0.28\linewidth}
		\includegraphics[height=3cm,width=3.8cm]{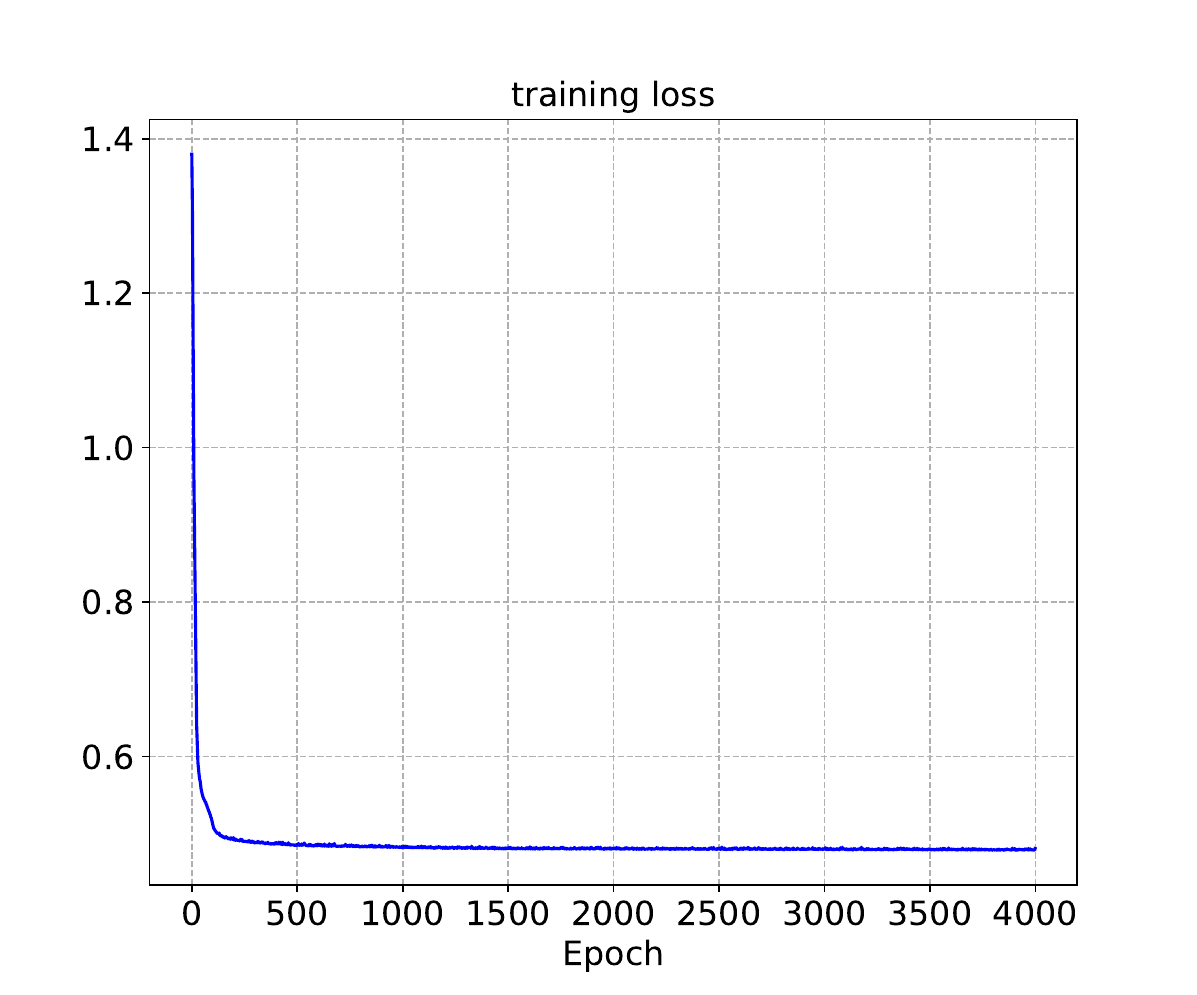}
	\end{minipage}
	\begin{minipage}[b]{0.28\linewidth}
		\includegraphics[height=3cm,width=3.8cm]{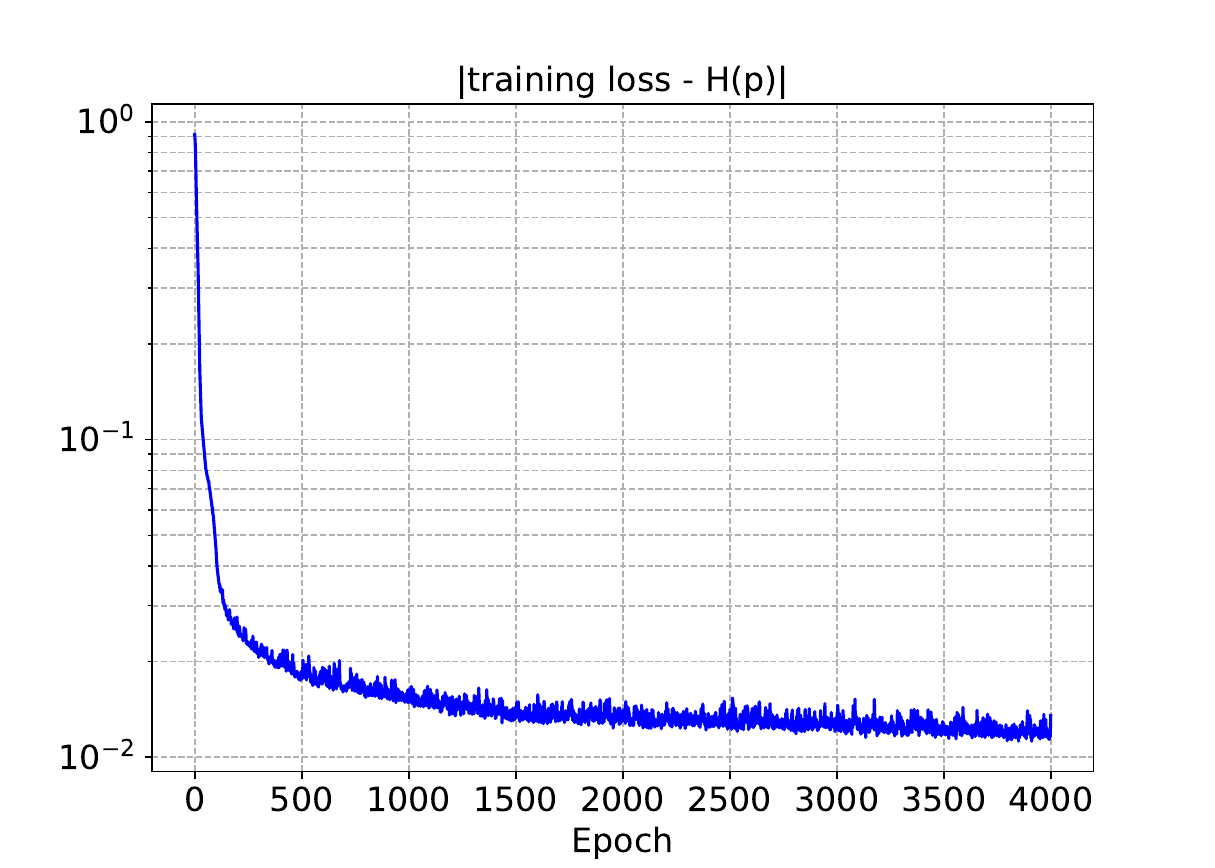}
	\end{minipage}
	\begin{minipage}[b]{0.28\linewidth}
		\includegraphics[height=3cm,width=3.8cm]{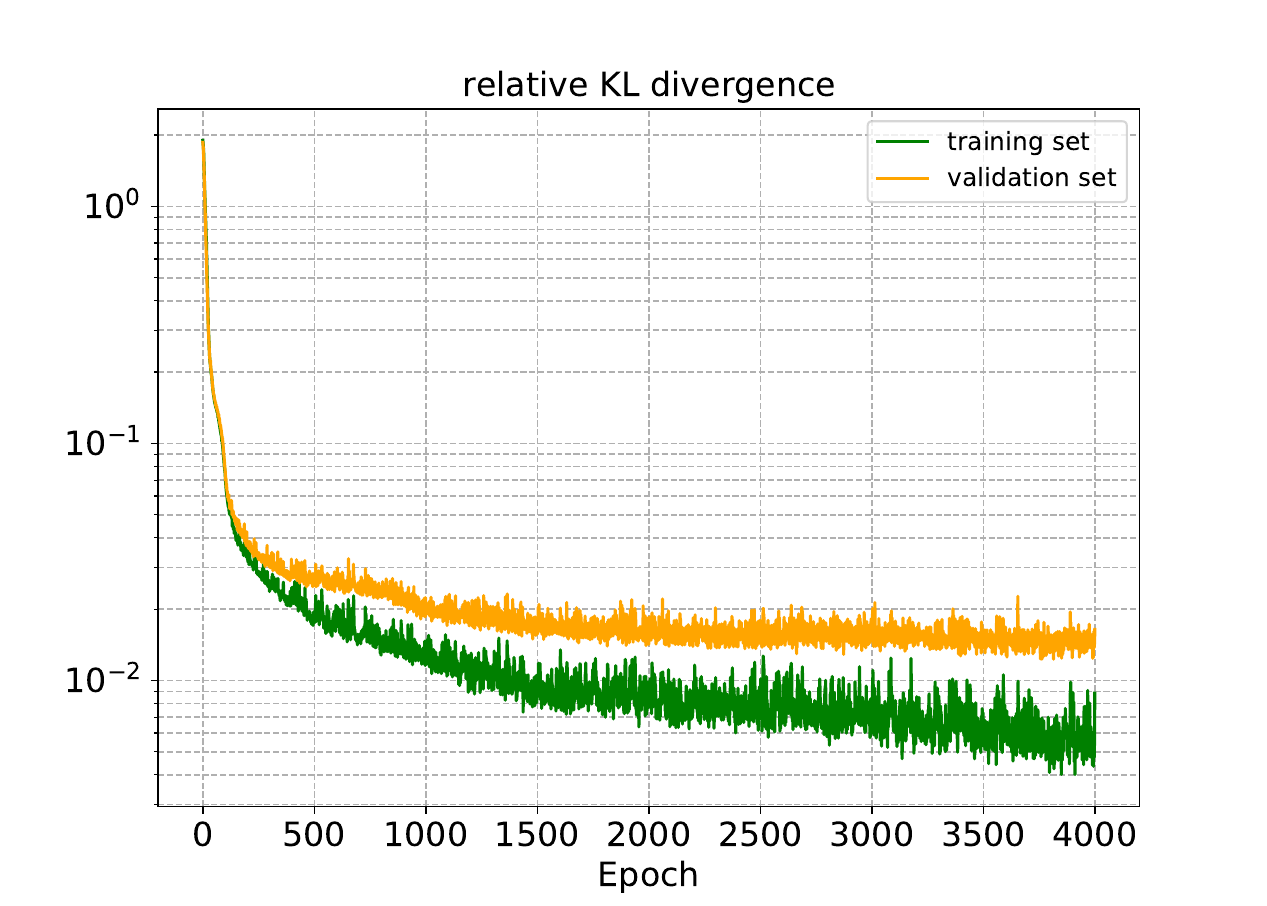}
	\end{minipage}
	\caption{Training procedure for the mixture of Gaussians on a box. Left: The training loss. Middle: The KL divergence on the training set. Right: The relative KL divergence on the training set and validation set.}
	\label{fig:mix_gauss_loss_kl}
\end{figure}

\begin{figure}[h!]
	\centering
	\begin{minipage}[b]{0.23\linewidth}
		\includegraphics[height=3cm,width=3cm]{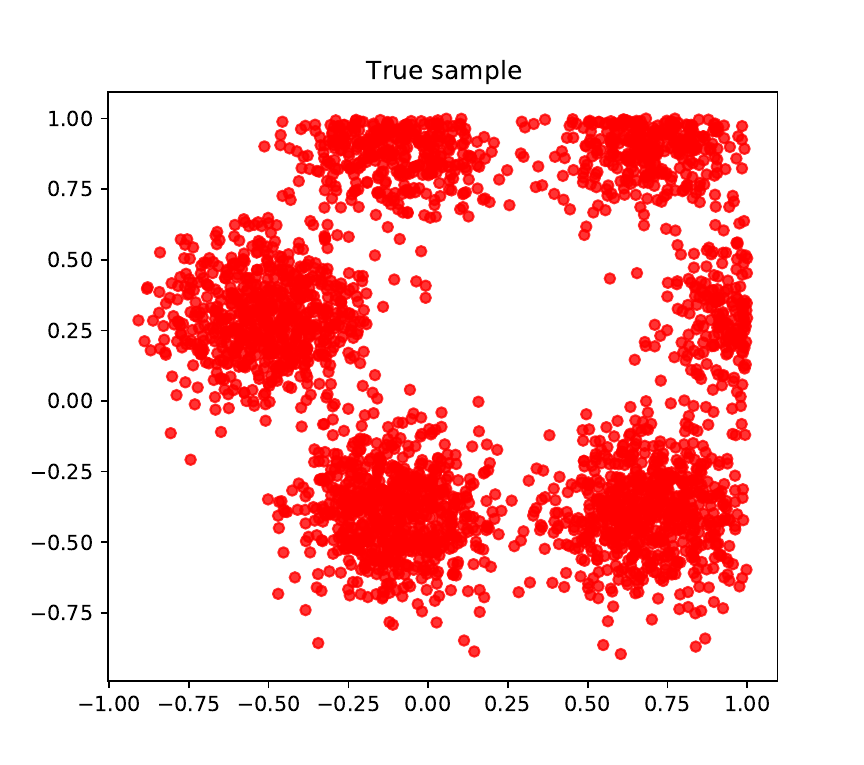}
	\end{minipage}
	\begin{minipage}[b]{0.23\linewidth}
		\includegraphics[height=3cm,width=3cm]{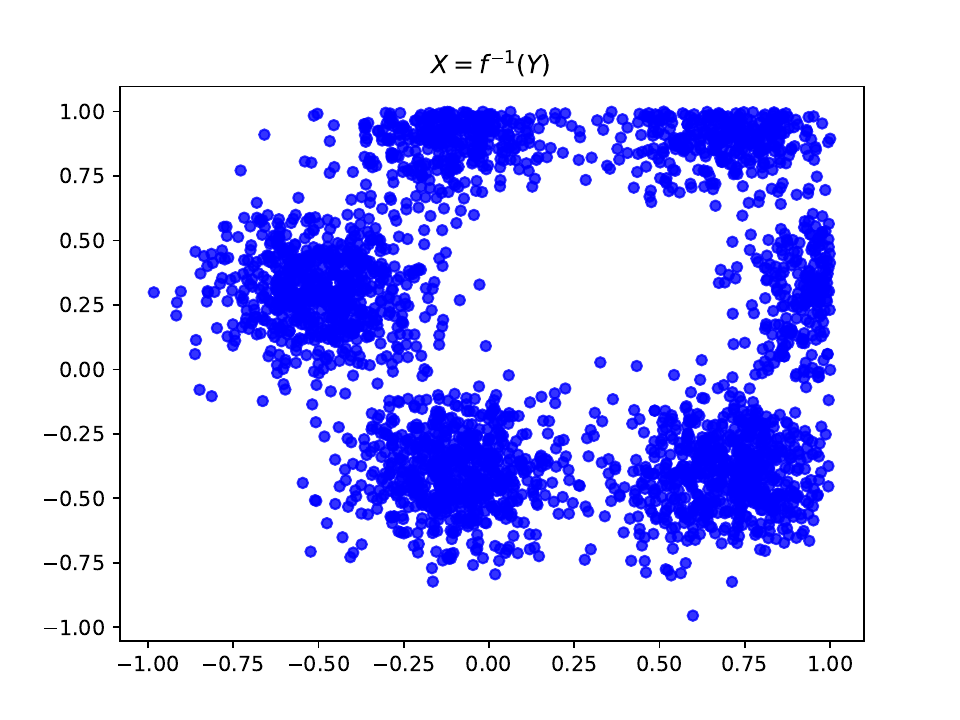}
	\end{minipage}
	\begin{minipage}[b]{0.23\linewidth}
		\includegraphics[height=3cm,width=3cm]{fig/circ/prior_sample.pdf}
	\end{minipage}
	\caption{Comparison between true samples and samples from B-KRnet for the mixture of Gaussians restricted to $[-1,1]^2$. Left: True samples. Middle: Samples from B-KRnet. Right: Samples from the base distribution (uniform).}
	\label{fig:mix_gauss_sample}
\end{figure}

The B-KRnet is built by $8$ CDF coupling layers with \(3\) subintervals along each dimension. \(2\times 10^4\) samples are applied with a batch size of 4096. The Adam optimizer with an initial learning of 0.001 is conducted.
We obtain the approximation of $H(p)$ by generating $10^6$ samples from the true distribution. 
The training loss and the relative KL divergence are presented in \cref{fig:mix_gauss_loss_kl}. We also compare the samples from the true distribution and samples from the PDF model in \cref{fig:mix_gauss_sample}. The B-KRnet model yields an excellent agreement with the ground truth distribution.

\begin{figure}[h!]
\centering
\includegraphics[scale=0.25]{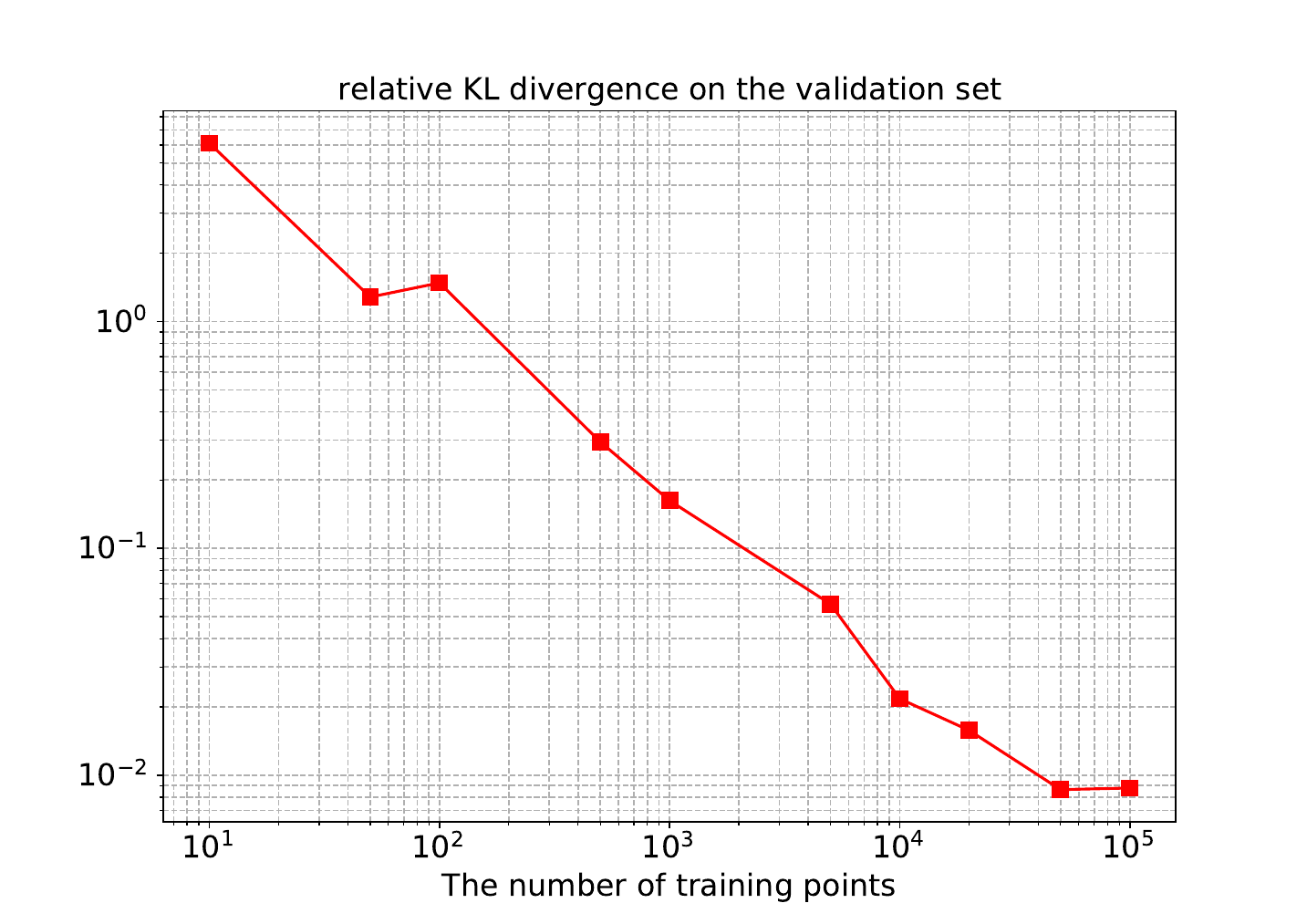}
\caption{The relative KL divergence for different numbers of the training points.}
\label{fig:gauss_kl_vary_samples}
\end{figure}

\begin{figure}[h!]
	\centering
	\begin{minipage}[b]{0.23\linewidth}
		\includegraphics[height=3cm,width=3cm]{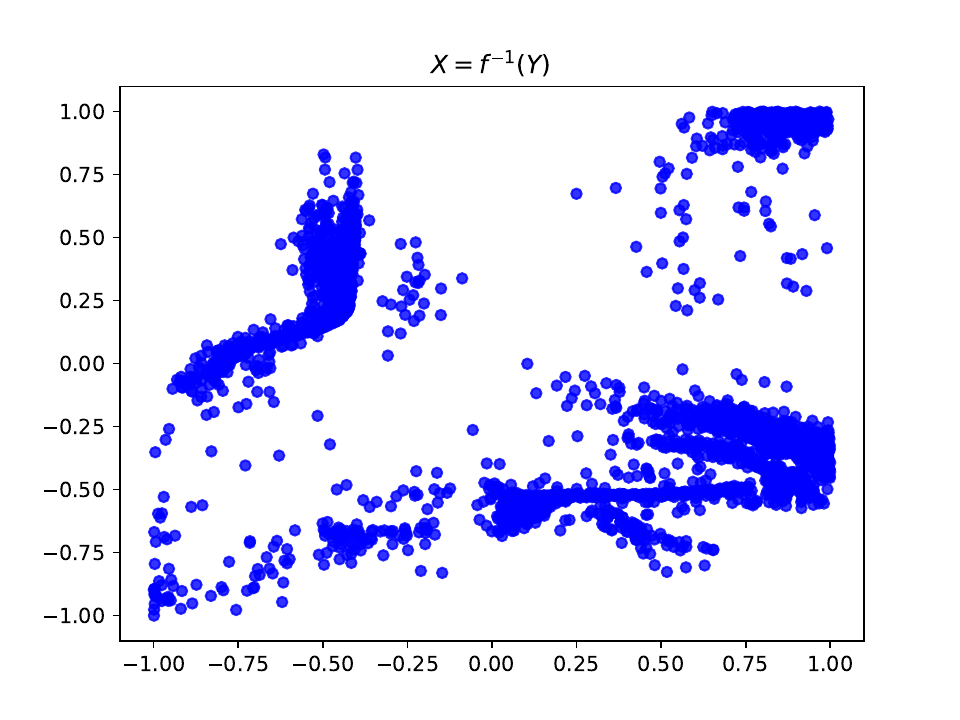}
        \subcaption{{N=10}}
	\end{minipage}
	\begin{minipage}[b]{0.23\linewidth}
		\includegraphics[height=3cm,width=3cm]{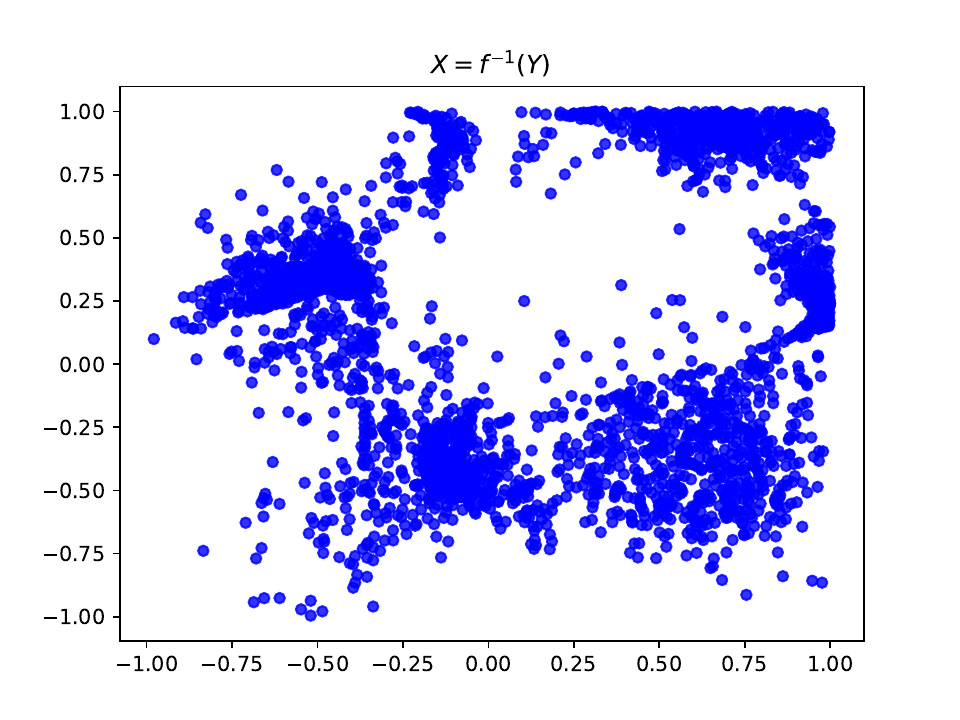}
         \subcaption{N=50}
	\end{minipage}
	\begin{minipage}[b]{0.23\linewidth}
		\includegraphics[height=3cm,width=3cm]{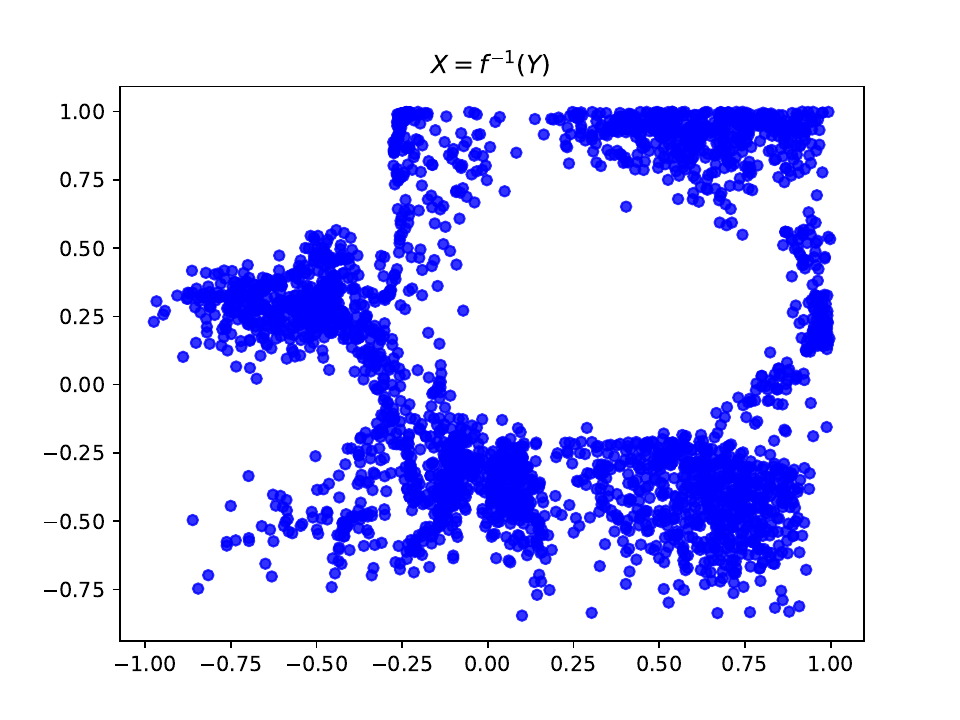}
            \subcaption{N=100}
	\end{minipage}
 \begin{minipage}[b]{0.23\linewidth}
		\includegraphics[height=3cm,width=3cm]{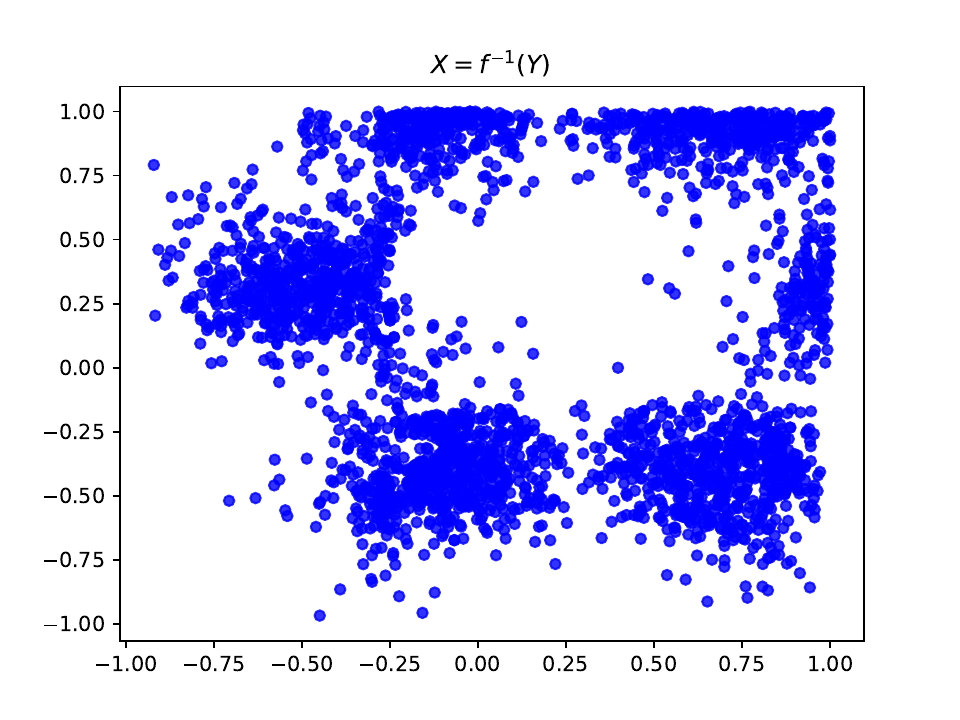}
            \subcaption{N=500}
	\end{minipage}
	\begin{minipage}[b]{0.23\linewidth}
		\includegraphics[height=3cm,width=3cm]{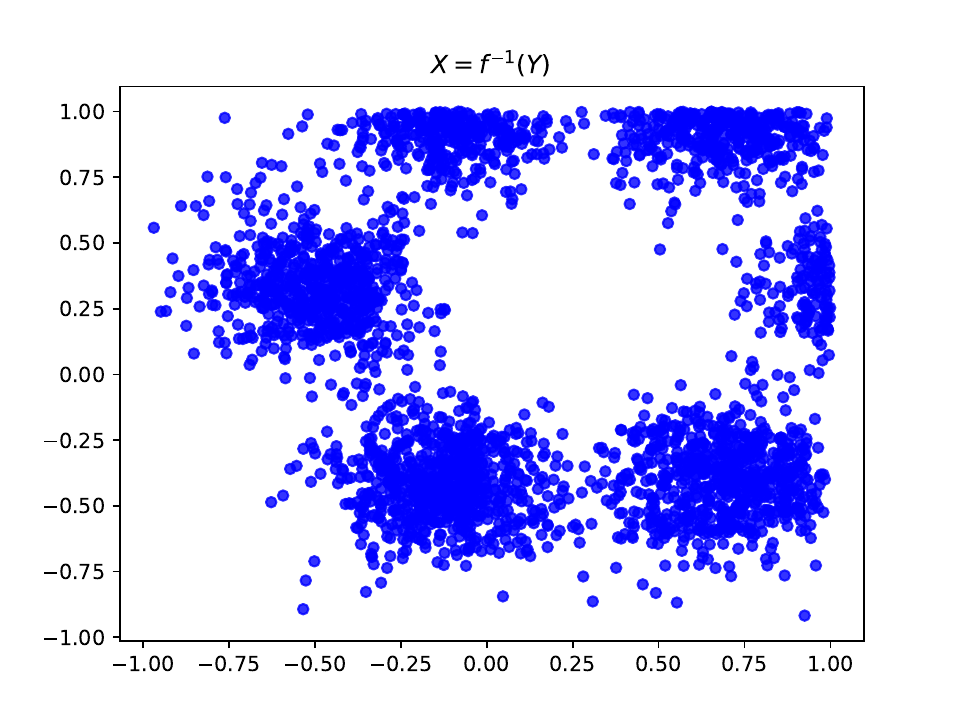}
            \subcaption{N=1000}
	\end{minipage}
	\begin{minipage}[b]{0.23\linewidth}
		\includegraphics[height=3cm,width=3cm]{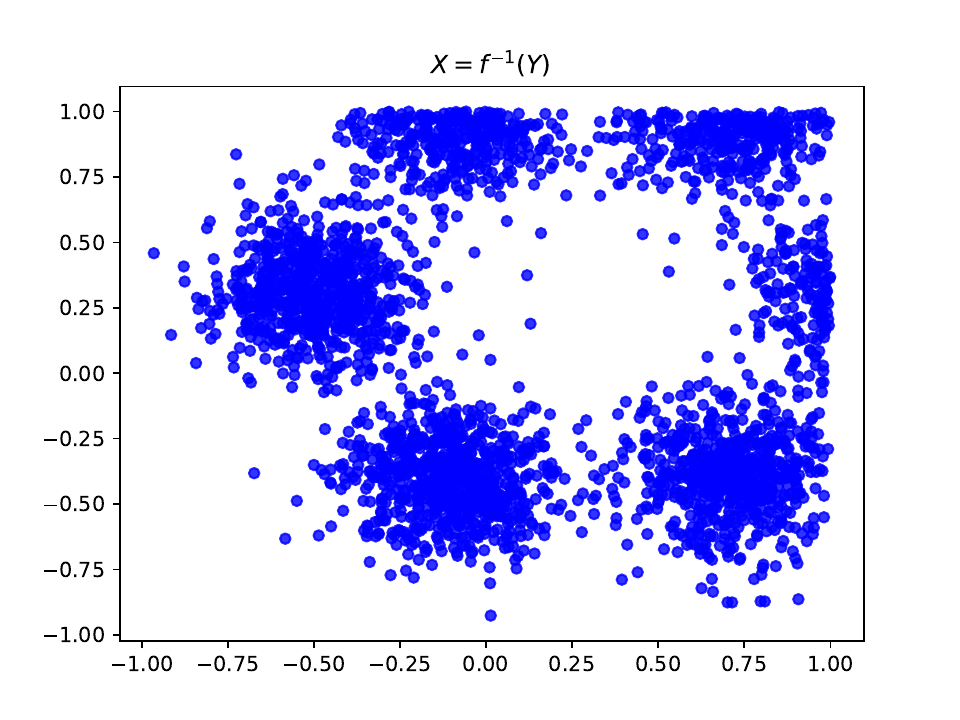}
            \subcaption{N=10000}
	\end{minipage}
	\begin{minipage}[b]{0.23\linewidth}
		\includegraphics[height=3cm,width=3cm]{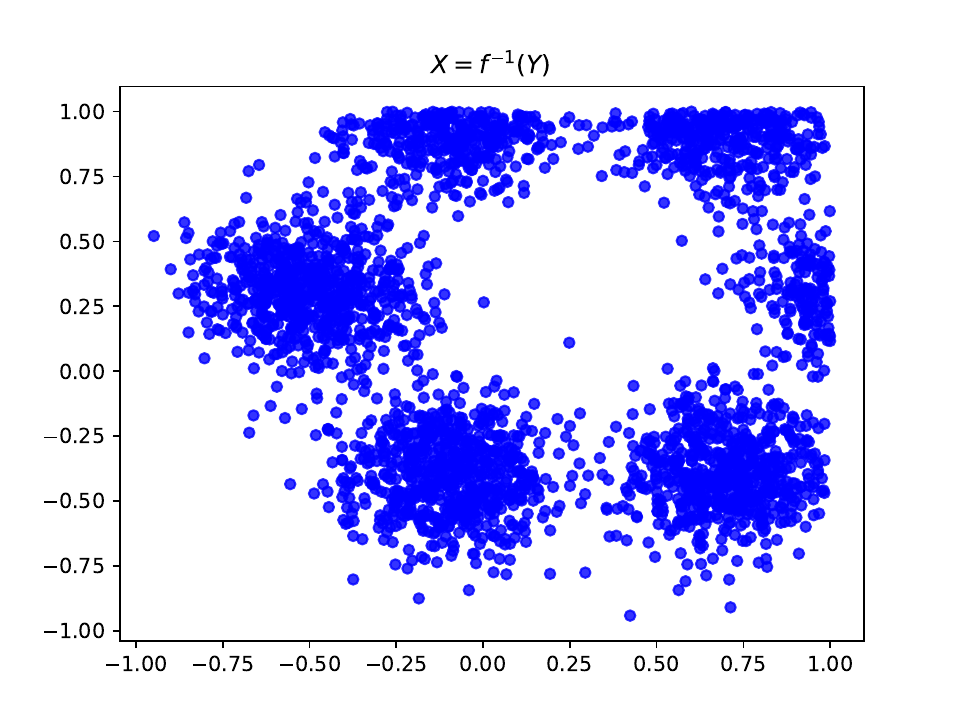}
          \subcaption{N=50000}
	\end{minipage}
	\begin{minipage}[b]{0.23\linewidth}
		\includegraphics[height=3cm,width=3cm]{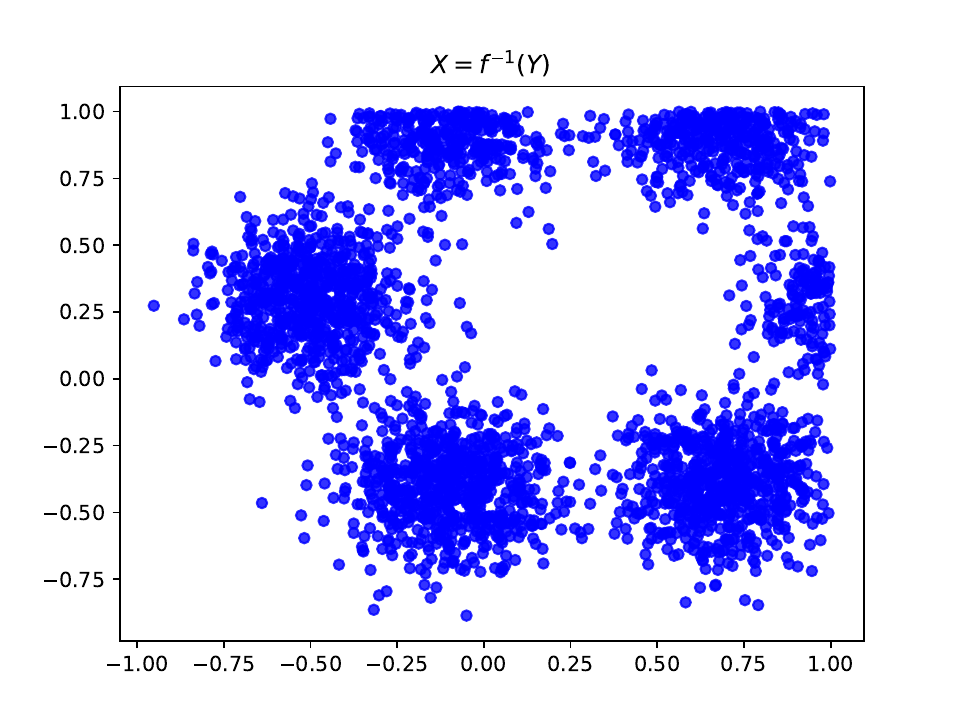}
            \subcaption{N=100000}
	\end{minipage}
	\caption{Samples generated by B-KRnets corresponding to different numbers of training points.}
	\label{fig:mix_gauss_sample_for_different_num}
\end{figure}

In addition, we investigate the proposed method by varying the training sample size. Batch training is employed when the number of training samples exceeds \(5000\) to maintain consistent optimization steps across different sample sizes. The relative KL divergence against the number of training points is shown in \cref{fig:gauss_kl_vary_samples}. It can be observed that the relative KL divergence decreases as the number of training points grows. However, after reaching a certain threshold, adding more training points does not further improve performance. \Cref{fig:mix_gauss_sample_for_different_num} presents the samples from B-KRnets corresponding to different training sample sizes.

\subsection{Example 3: Logistic distribution with holes}
We now consider density estimation for a logistic distribution with holes on a bounded domain $\Omega=[-10, 10]^d$. The training dataset $S=\{\bm{y}^{(i)}\}^{N}_{i=1}$ is generated from $p_{\bm{Y}}(\bm{y})$, where the components of $\bm{Y}=(Y_1, Y_2, \dots, Y_d)$ are i.i.d and each component $Y_i\sim$Logistic$(0,s)$ with a PDF $\rho(y_i;0,s)$. We 
restrict $\bm{Y}$ to $[-10,10]^d$ and require that it also satisfies the condition that the 2-norm of the vector, obtained by multiplying two adjacent components with the matrix by the matrix \(R_{\gamma,\theta_j}\) is larger than a constant \(C\):
\begin{equation}
\label{eqn:constraint2}
\left\|R_{\gamma,\theta_j}\left[y_j^{(i)}, y^{(i)}_{j+1}\right]^{\rm{T}}\right\|_2\geq C, \quad j=1,\dots,d-1,
\end{equation}
where 
\begin{equation}
R_{\gamma, \theta_j}=\left[\begin{array}{cc}
\gamma & 0\\
0 & 1\\
\end{array}\right]\left[\begin{array}{cc}
\cos\theta_j &-\sin\theta_j\\
\sin\theta_j & \cos\theta_j \\
\end{array}\right], \quad \theta_j=\left\{\begin{array}{cc}
\frac{\pi}{4}, &\text{  if } j \text{ is even; }\\
\frac{3\pi}{4}, &\text{otherwise}.\end{array}\right.
\end{equation}
Then an elliptic hole is generated for two adjacent dimensions. The reference PDF takes the form 
\begin{equation}
p_{\bm{Y}}(\bm{y})=\frac{1_{B}(\bm{y})\prod_{i=1}^d\rho(y_i;0,s)}{\mathbb{E}[1_{B}(\bm{Y})]},
\end{equation}
where $B\subset\Omega$ is the domain admitting constraint \eqref{eqn:constraint2}, and $1_B(\cdot)$ is an indicator function with $1_B(y)=1$ if $\bm{y}\in B;\; 0$ otherwise.

Let $d=8$, $C=5$, $\gamma=3$ and $s=2$.
 To demonstrate the effectiveness of the structure with descending active transformation dimensions, we compare the performance of B-KRnet models subject to either 
 a structure with decreasing active dimensions or the half-half partition used by classical flow-based models. For the sake of convenience, we call the latter Classical-NF. In fact, Classical-NF can be considered as $f^{outer}_{[1]}\circ L_{lin}$ without the outer loop. In B-KRnet, we deactivate the dimensions by one, i.e., $K=8$. 
 Let $l_k=l_{k-1}-2$ if $l_{k-1}\geq 6$ and $l_1=16$. 
 To roughly match the DOFs of B-KRnet, 
 Classical-NF consists of \(62\) CDF coupling layers. B-KRnet and Classical-NF share the same structure for the CDF coupling layer, with three subintervals along each dimension.
 The resulting number of parameters in Classical-NF is slightly larger than that in B-KRnet. $10^6$ training points are 
 used, with a batch size of $2 \times 10^5$. The Adam optimizer with an initial learning rate of $0.001$ is applied.  
 Moreover, we reduce the learning rate to half of its initial value after every $5000$ epochs. A total of $10000$ epochs are conducted and a validation dataset with $10^6$ samples is used for computing the relative KL divergence throughout the training process. 
 
\begin{figure}[h!]
	\centering
	\begin{minipage}[b]{0.28\linewidth}
		\includegraphics[height=3cm,width=3.8cm]{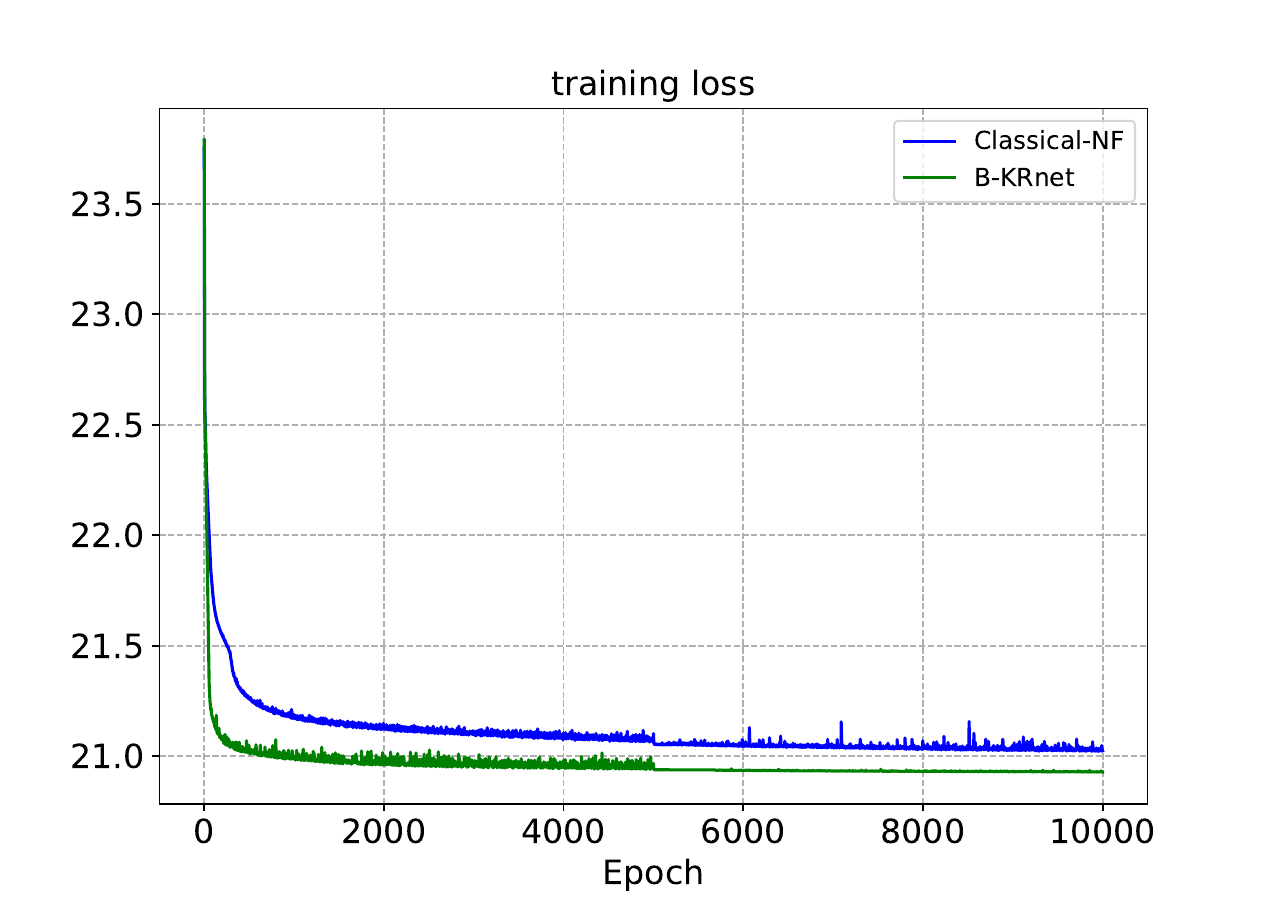}
	\end{minipage}
	\begin{minipage}[b]{0.28\linewidth}
		\includegraphics[height=3cm,width=3.8cm]{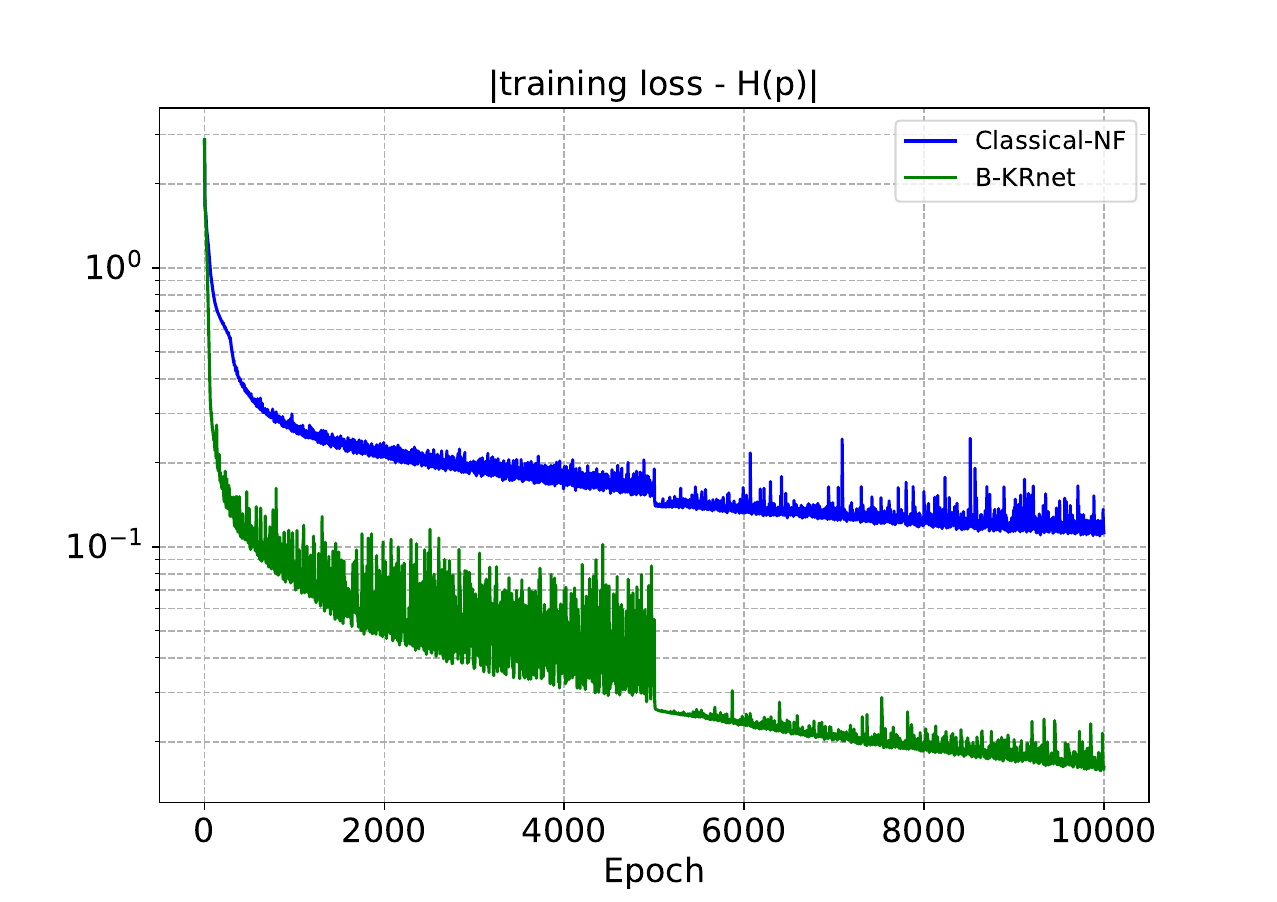}
	\end{minipage}
	\begin{minipage}[b]{0.28\linewidth}
		\includegraphics[height=3cm,width=3.8cm]{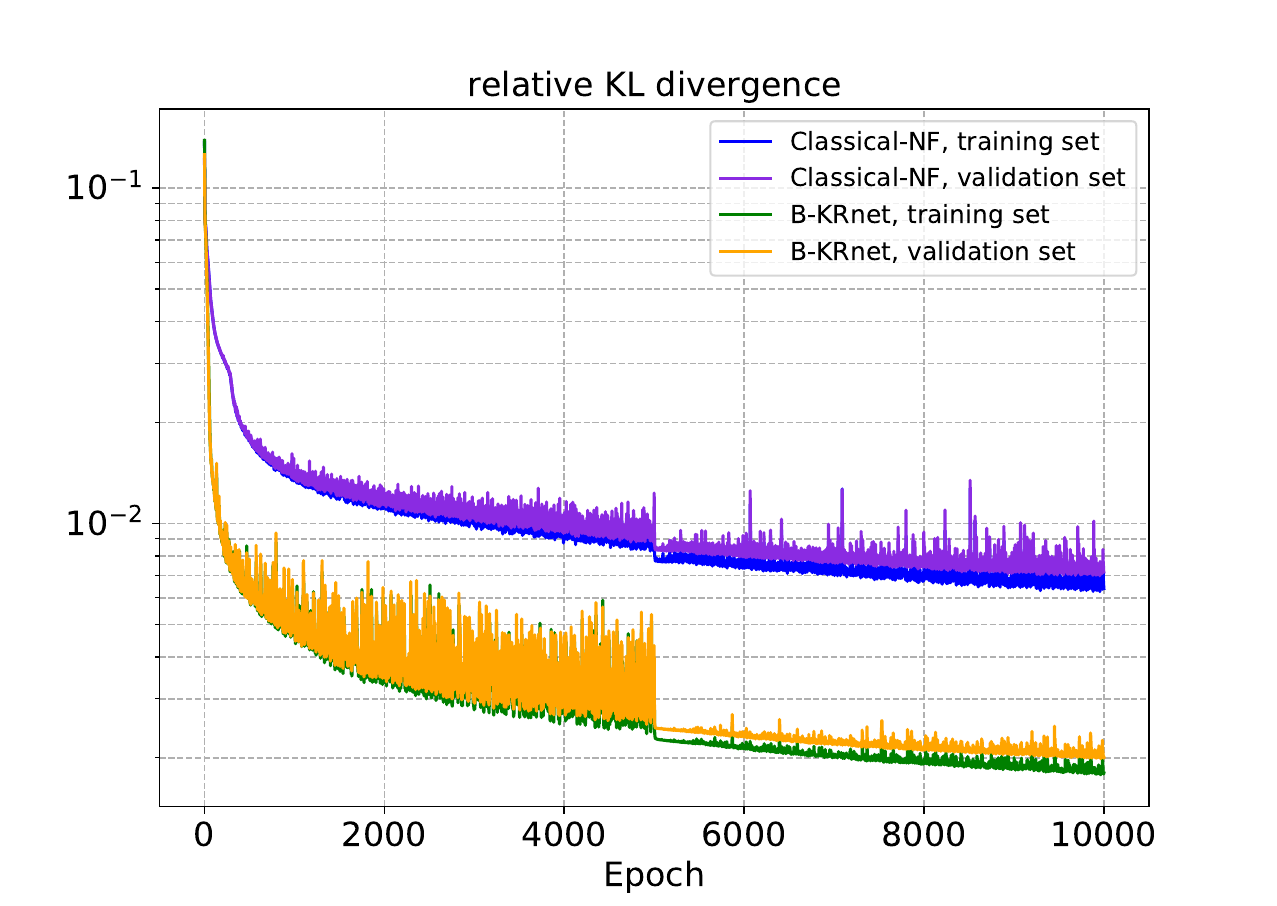}
	\end{minipage}
	\caption{Comparison between Classical-NF and B-KRnet for $d=8$. Left: The training loss. Middle: The KL divergence on the training set. Right: The relative KL divergence on the training set and validation set.}
	\label{fig:d8_log_loss_kl}
\end{figure}

\begin{figure}[h!]
	\centering
\begin{minipage}[b]{0.23\linewidth}
	\includegraphics[height=3cm,width=3cm]{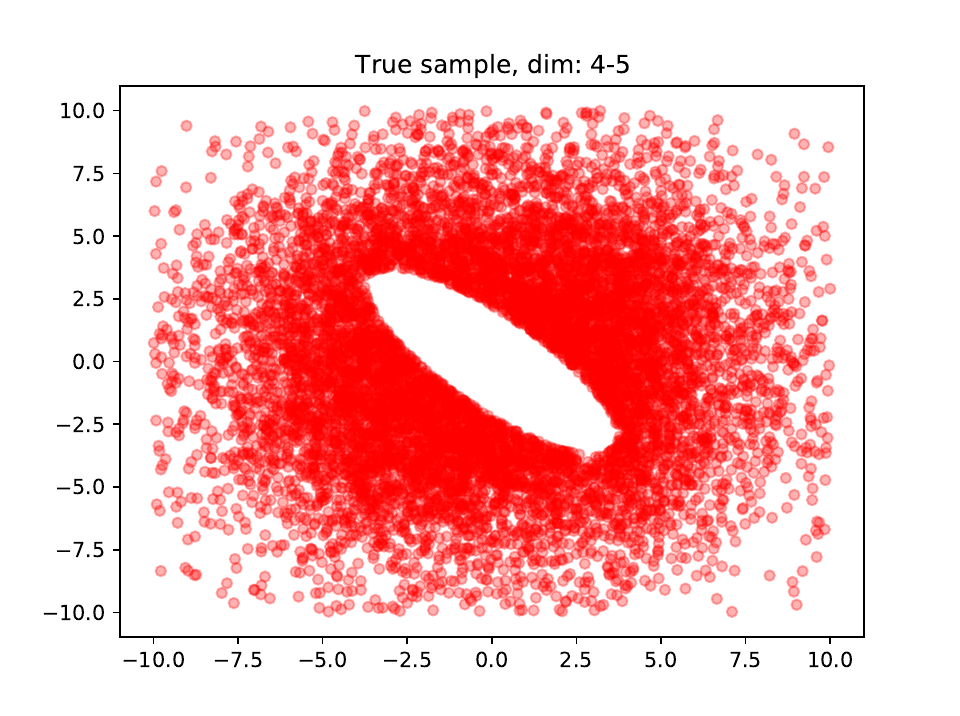}
\end{minipage}
	\begin{minipage}[b]{0.23\linewidth}
	\includegraphics[height=3cm,width=3cm]{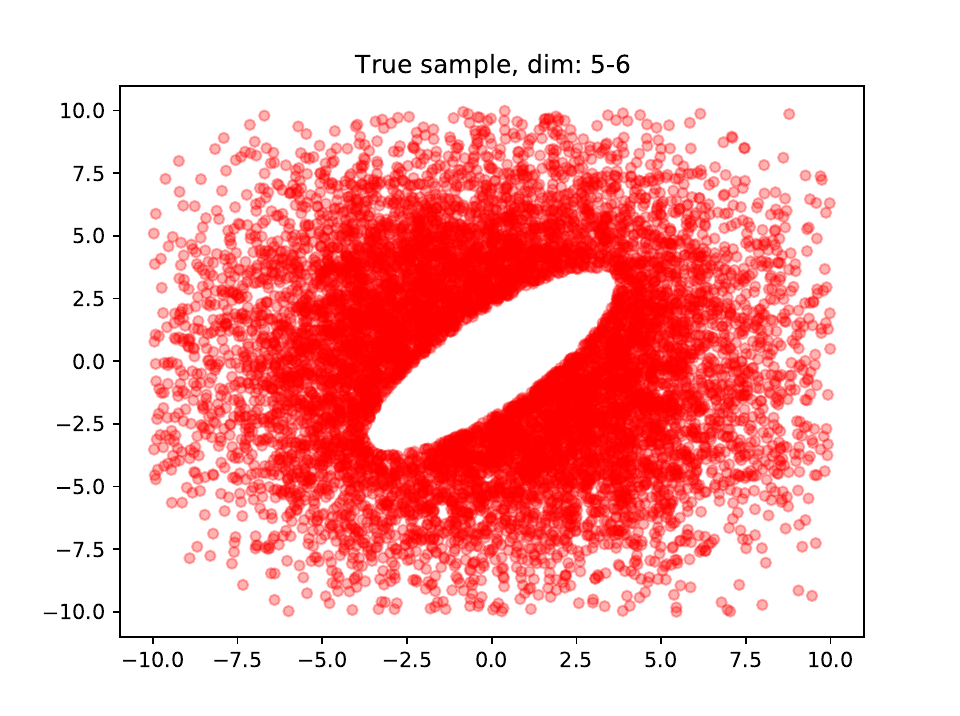}
\end{minipage}
\begin{minipage}[b]{0.23\linewidth}
	\includegraphics[height=3cm,width=3cm]{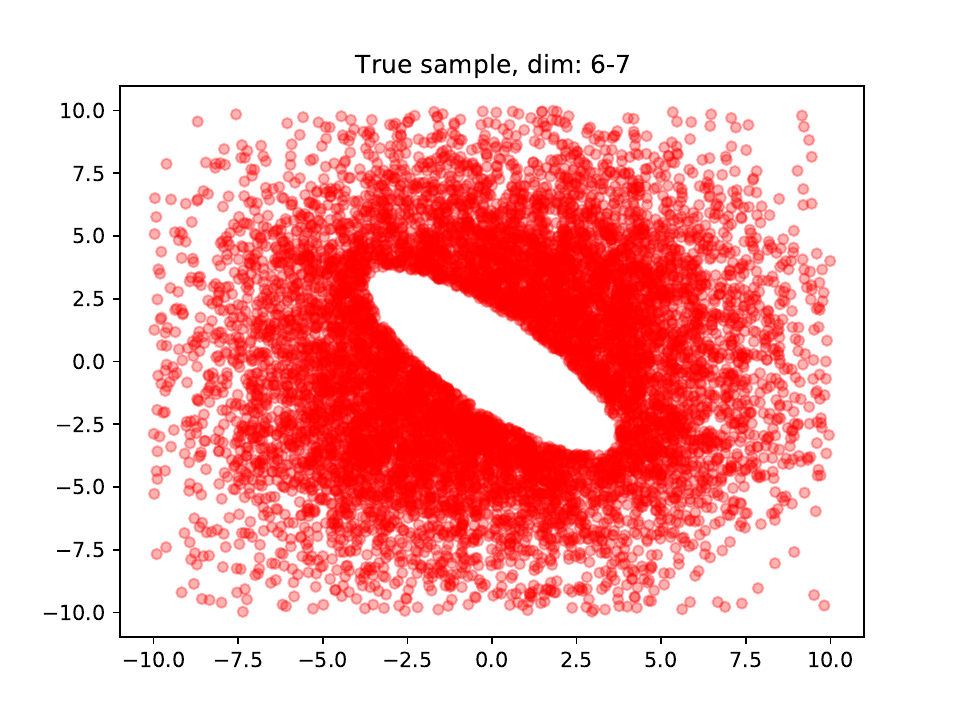}
\end{minipage}
\begin{minipage}[b]{0.23\linewidth}
	\includegraphics[height=3cm,width=3cm]{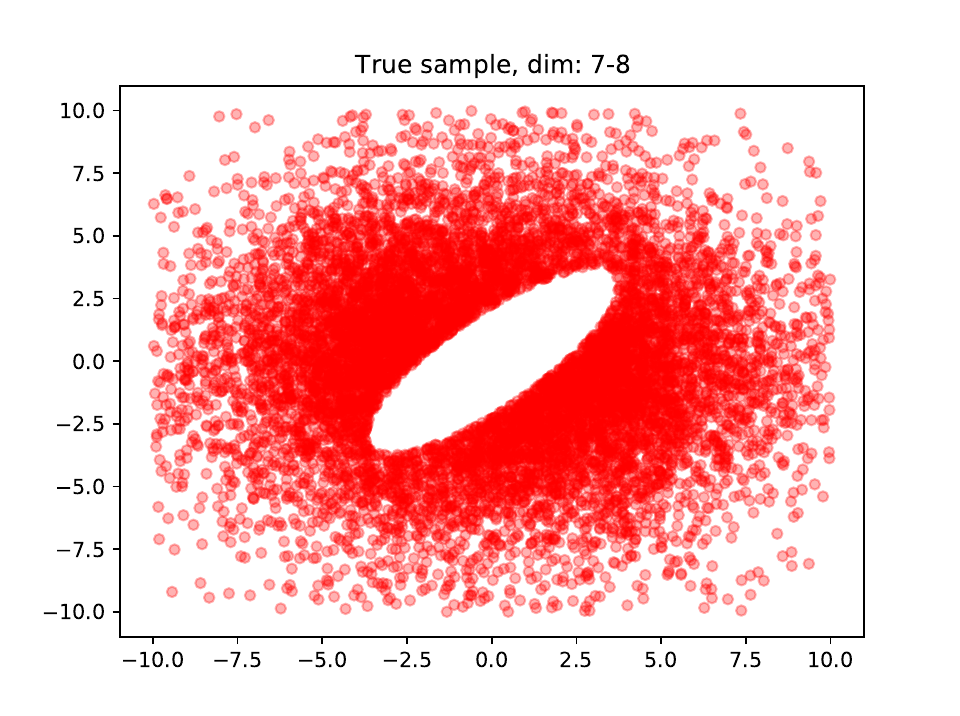}
\end{minipage}

	\begin{minipage}[b]{0.23\linewidth}
	\includegraphics[height=3cm,width=3cm]{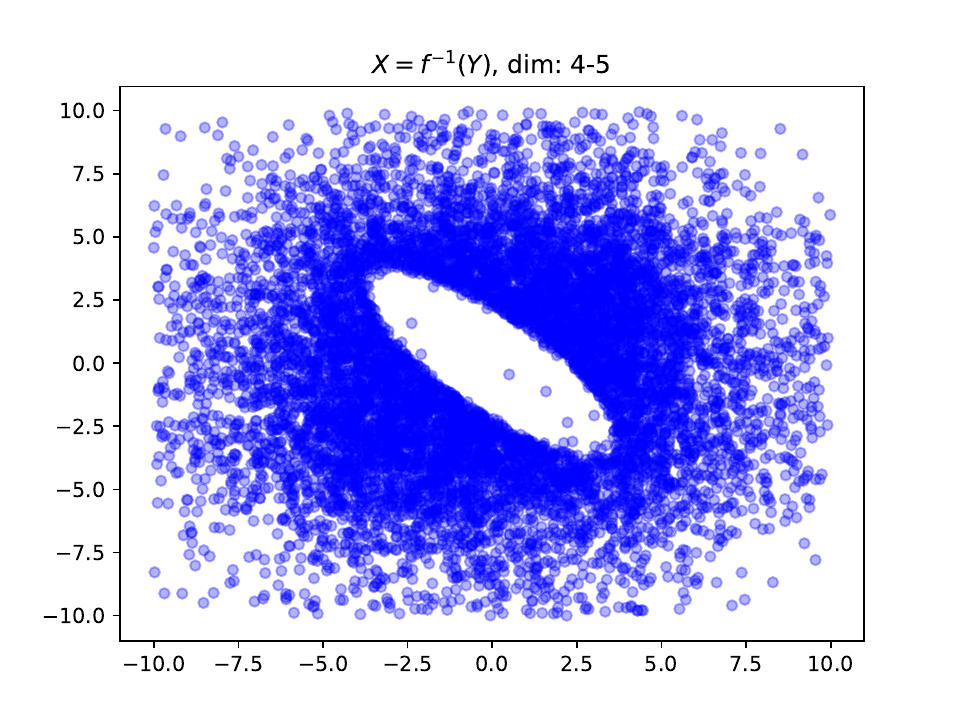}
\end{minipage}
\begin{minipage}[b]{0.23\linewidth}
	\includegraphics[height=3cm,width=3cm]{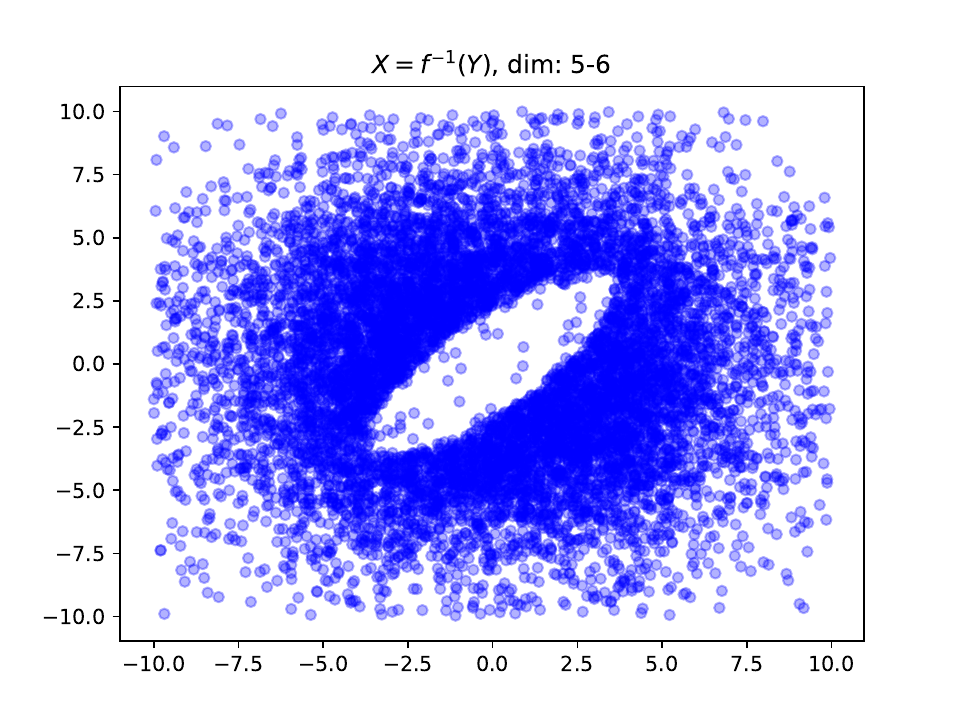}
\end{minipage}
\begin{minipage}[b]{0.23\linewidth}
	\includegraphics[height=3cm,width=3cm]{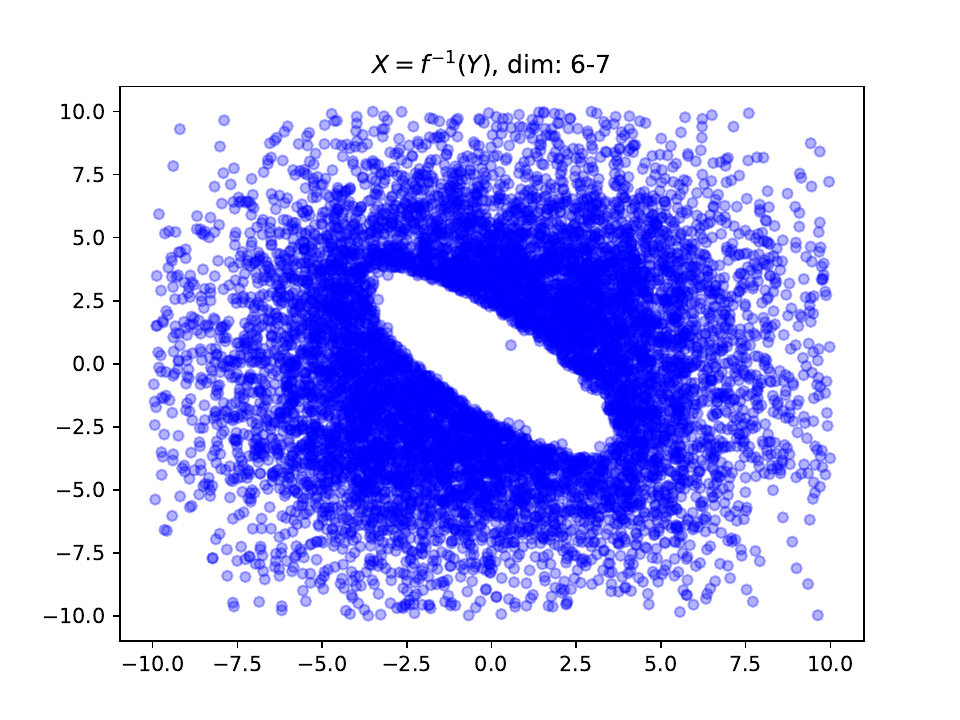}
\end{minipage}
\begin{minipage}[b]{0.23\linewidth}
	\includegraphics[height=3cm,width=3cm]{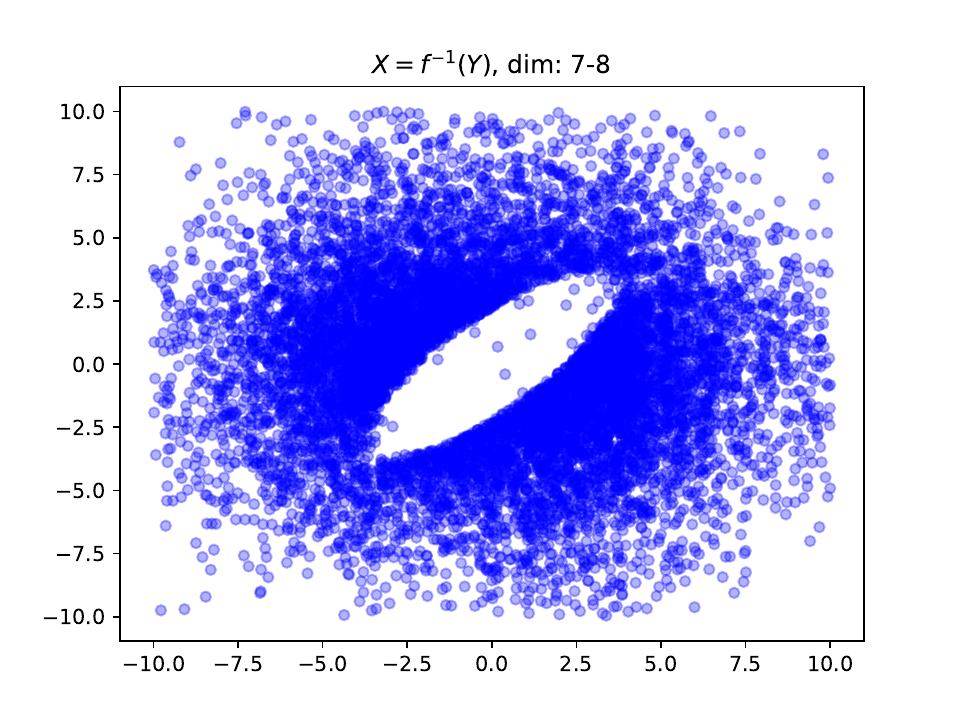}
\end{minipage}

	\begin{minipage}[b]{0.23\linewidth}
		\includegraphics[height=3cm,width=3cm]{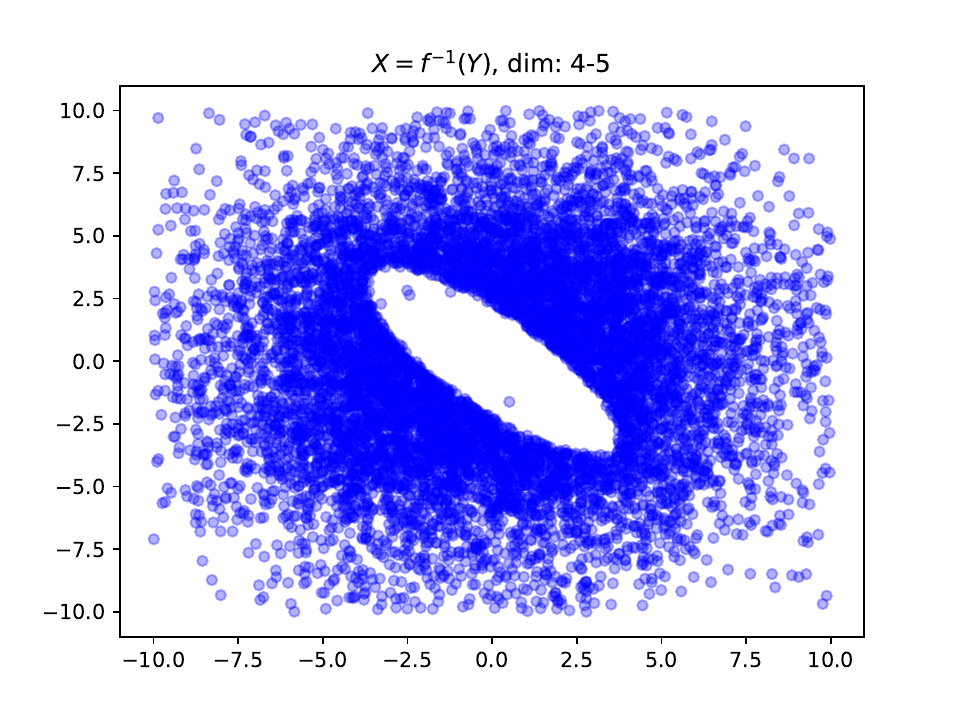}
	\end{minipage}
\begin{minipage}[b]{0.23\linewidth}
	\includegraphics[height=3cm,width=3cm]{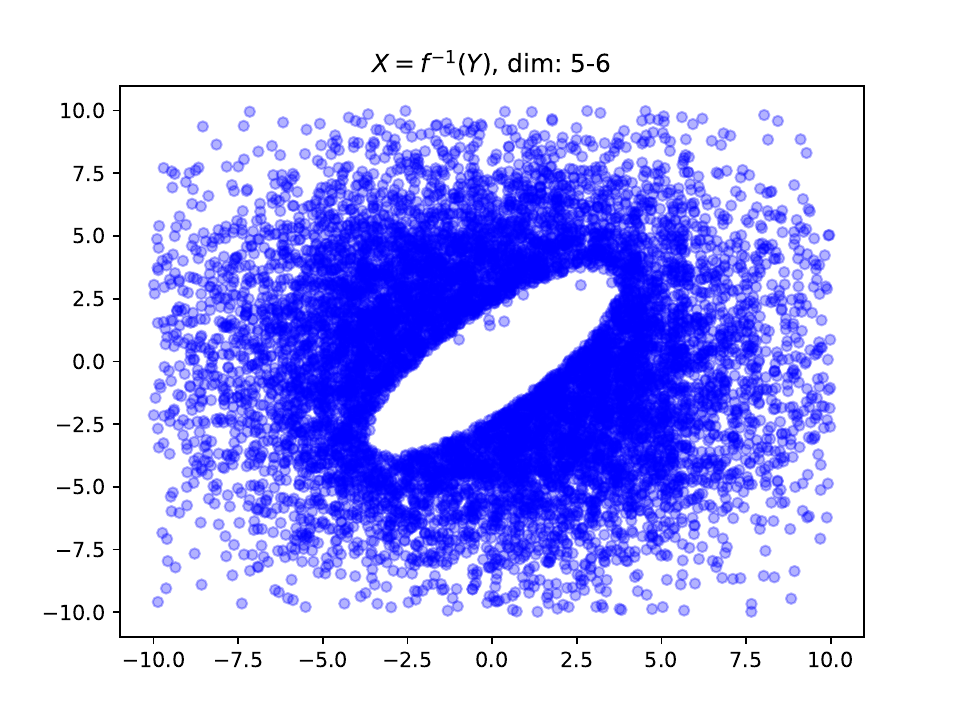}
\end{minipage}
\begin{minipage}[b]{0.23\linewidth}
	\includegraphics[height=3cm,width=3cm]{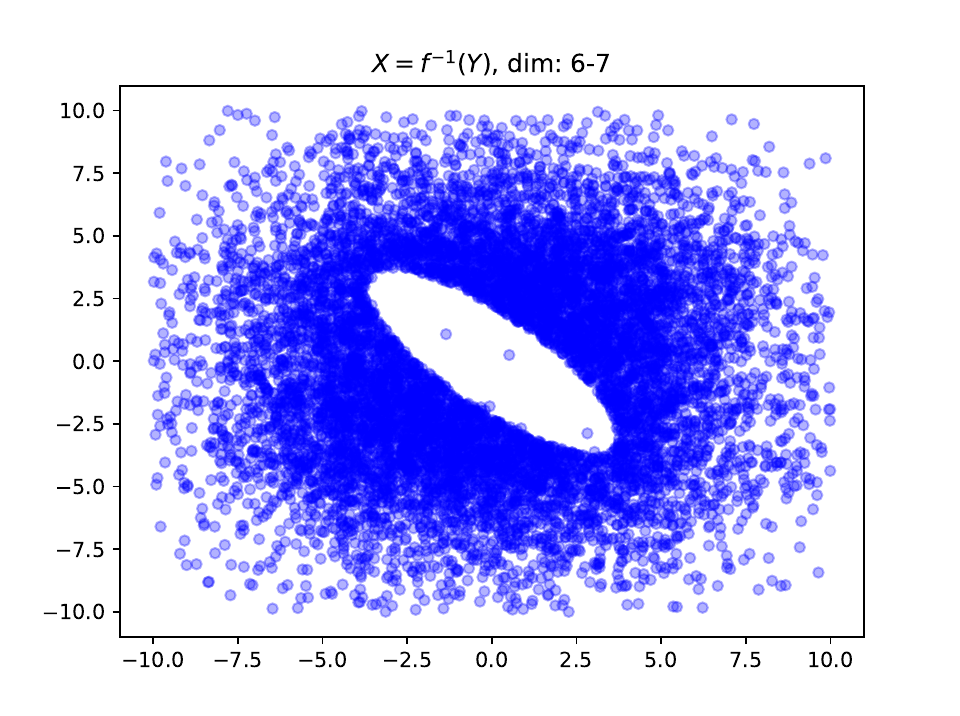}
\end{minipage}
\begin{minipage}[b]{0.23\linewidth}
	\includegraphics[height=3cm,width=3cm]{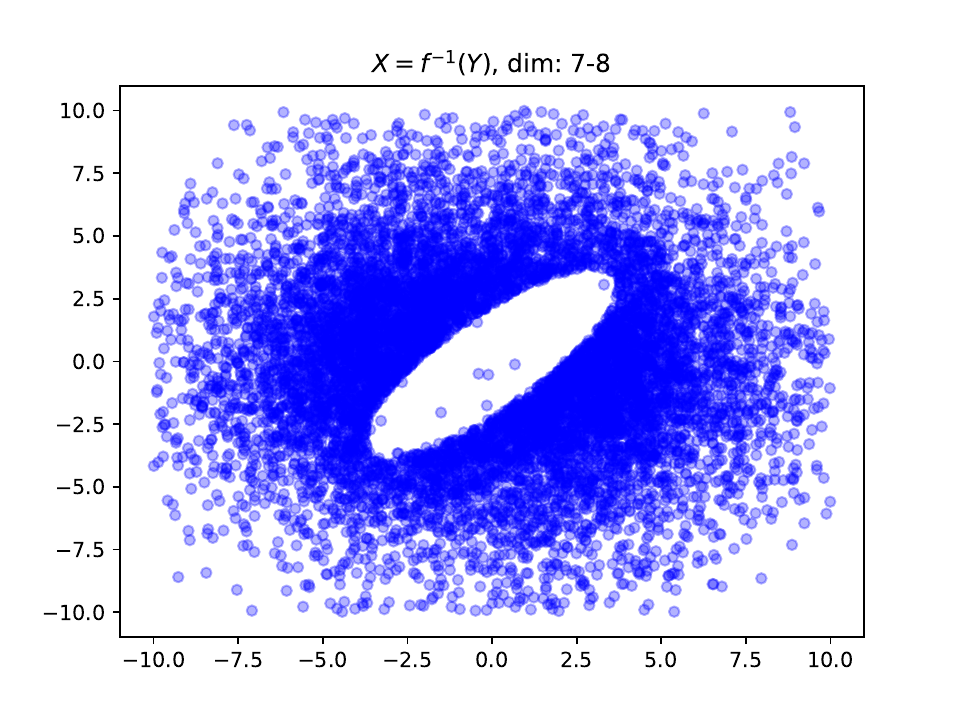}
\end{minipage}
\caption{Comparison between true samples and samples from PDF models for $d=8$. Top: True samples. Middle: Samples from Classical-NF. Right: Samples from B-KRnet.}
\label{fig:d8_log_sample}
\end{figure}

The training loss and relative KL divergence are presented in \cref{fig:d8_log_loss_kl}, which indicate that KR rearrangement achieves higher accuracy than the half-half partition. We also compare the samples from the true distribution and samples from PDF models in \cref{fig:d8_log_sample}. The samples generated from B-KRnet show a better agreement with the true samples than Classical-NF.

\section{Application to solving PDEs on a bounded domain}\label{sec:adaptive_method}
In this part, we employ B-KRnet to approximate PDEs 
related to density and we elaborate on the adaptive learning procedure in detail. Let $\Omega\subset\mathbb{R}^d$ be a hyperrectangle domain and $\bm{p}: \Omega\to\mathbb{R}_+$. 
Denote \(\partial^{\bm{\alpha}}_{\bm{x}}f(\bm{x})=\frac{\partial^{|\alpha|}}{\partial^{\alpha_1}_{x_1}\cdots\partial^{\alpha_d}_{x_d}}f(\bm{x})\) for \(\bm{\alpha}=(\alpha_1,\dots,\alpha_d)\in\mathbb{N}^d_0\) and \(|\bm{\alpha}|=\alpha_1+\cdots+\alpha_d\).
We consider the following 
second-order PDE
\begin{equation}
\label{eqn:pde_problem}
\left\{\,\begin{aligned}
\mathcal{N}\left[\bm{x};
\left\{\partial^{\bm{\alpha}}_{\bm{x}} p,\, |\bm{\alpha}|\leq2\right\}\right]=0, &\quad\bm{x}\in\Omega,\\
\mathcal{B}\left[\bm{x};
\left\{\partial^{\bm{\alpha}}_{\bm{x}} p,\, |\bm{\alpha}|\leq 2\right\}\right]=0, &\quad\bm{x}\in\partial\Omega,
\end{aligned}
\right.
\end{equation}
where $\mathcal{N}$ is a linear or nonlinear second-order differential  
operator acting on \(p\). $\mathcal{B}$ is a boundary operator. We assume that the solution $p(\bm{x})$ is non-negative and mass-conservative, which is very common in physical and biological modeling. That is to say,
\begin{equation*}
    \int_{\Omega}{p}(\bm{x}){\rm{d}}\bm{x}=M, \quad {p}(\bm{x})\geq 0, \quad \bm{x}\in\Omega,
\end{equation*}
where \(M\) is a constant. Then a trainable parameter \(\zeta\) can be introduced to approximate \(M\), resulting in \(\zeta \hat{p}\) as an approximator for \(p\) where \(\hat{p}\) is a PDF.

For simplicity, we consider \(p\) to be a PDF, i.e.,
\begin{equation}
\label{eqn:pdf_constraint}
\int_{\Omega}{p}(\bm{x}){\rm{d}}\bm{x}=1, \quad {p}(\bm{x})\geq 0, \quad \bm{x}\in\Omega.
\end{equation}
 We map $\bm{X}$ to a random vector $\bm{Z}$ which is uniformly distributed on $[-1,1]^d$ by B-KRnet and use the induced density model $p_{\text{B-KRnet},\bm{\theta}}(\bm{x})$ as a surrogate for $p(\bm{x})$. 
 Since $p_{\text{B-KRnet}, \bm{\theta}}$ is a PDF, the non-negativity and conservation constraints \eqref{eqn:pdf_constraint} naturally hold. Note that B-KRnet is limited to first-order differentiation since the CDF coupling layer is piecewise quadratic. To deal with a second-order PDE, we convert it into a first-order system by introducing an auxiliary vector function to represent the gradient $\nabla p$.
 
 \subsection{Tackle second order differential operator: auxiliary function}\label{sec:aux_nn}
 
 We 
 introduce an auxiliary vector function $\bm{g}(\bm{x})=\nabla p$. Then $\frac{\partial^2p}{\partial x_i\partial x_j}$  can be approximated by $\frac{\partial\bm{g}_j}{\partial x_i}$.   
  In this way, we can rewrite equation \eqref{eqn:pde_problem} as a first-order system
\begin{equation}
\label{eqn:pde_problem_aux}
\left\{\,\begin{aligned}
\mathcal{N}\left[\bm{x};{p}, \left\{\partial^{\bm{\alpha}}_{\bm{x}} p, \, \partial^{\bm{\beta}}_{\bm{x}} \bm{g},\, |\bm{\alpha}|, |\bm{\beta}|=1\right\}\right]=0, &\quad\bm{x}\in\Omega,\\
\mathcal{B}\left[\bm{x}; p,\left\{\partial^{\bm{\alpha}}_{\bm{x}} p, \, \partial^{\bm{\beta}}_{\bm{x}} \bm{g},\, |\bm{\alpha}|, |\bm{\beta}|=1\right\}\right]=0, &\quad\bm{x}\in\partial\Omega,\\
\bm{g}=\nabla p, &\quad\bm{x}\in\Omega.
\end{aligned}
\right.
\end{equation}

For simplicity, we abbreviate \(\mathcal{N}\left[\bm{x};{p}, \left\{\partial^{\bm{\alpha}}_{\bm{x}} p, \, \partial^{\bm{\beta}}_{\bm{x}} \bm{g},\, |\bm{\alpha}|, |\bm{\beta}|=1\right\}\right]\) as \(\mathcal{N}[\bm{x};p,\bm{g}]\), and \(\mathcal{B}\left[\bm{x}; p,\left\{\partial^{\bm{\alpha}}_{\bm{x}} p, \, \partial^{\bm{\beta}}_{\bm{x}} \bm{g},\, |\bm{\alpha}|, |\bm{\beta}|=1\right\}\right]\) as \(\mathcal{B}[\bm{x};p,\bm{g}]\).
We employ a neural network $\bm{g}_{\text{NN}}$ to approximate $\bm{g}$.
Following the physics-informed neural networks (PINNs), we consider the loss, 
\begin{equation}
\label{eqn:PINN}
\mathcal{L}(p_{\text{B-KRnet},\bm{\theta}}, \bm{g}_{\text{NN}})=\lambda_{{pde}}\mathcal{L}_{{pde}}
+\lambda_{b}\mathcal{L}_b
+\lambda_{\bm{g}}\mathcal{L}_{\bm{g}}
,
\end{equation}
where $\{\lambda_{pde},\lambda_{b}, \lambda_{\bm{g}}\}$ are hyperparameters to balance  the losses $\mathcal{L}_{pde}$, $\mathcal{L}_{b}$ and $\mathcal{L}_{\bm{g}}$ induced by the governing equations, the boundary conditions and $\bm{g}_{\text{NN}}$ respectively. More specifically,
\begin{align}
&\mathcal{L}_{pde}
=\mathbb{E}_{\bm{x}\sim\rho}\big[\left|\mathcal{N}[\bm{x}; p_{\text{B-KRnet},\bm{\theta}}, \bm{g}_{\text{NN}}]\right|^2\big], \label{eqn:Loss_pde}\\ &\mathcal{L}_{b}
=\mathbb{E}_{\bm{x}\sim\rho_b}\big[\left|\mathcal{B}[\bm{x}; p_{\text{B-KRnet},\bm{\theta}}, \bm{g}_{\text{NN}}]\right|^2\big],\\
&\mathcal{L}_{\bm{g}}
=\mathbb{E}_{\bm{x}\sim\rho}\big[\left\|\bm{g}_{\text{NN}}(\bm{x})-\nabla p_{\text{B-KRnet},\bm{\theta}}(\bm{x})\right\|^2_2\big],\label{eqn:Loss_g}
\end{align}
where $\rho(\bm{x})$ and $\rho_b(\bm{x})$ are positive PDFs on $\Omega$ and $\partial\Omega$ respectively.  
Given training data $C_{pde}=\{\bm{x}^i\}_{i=1}^{N_{pde}}$, $C_b=\{\bm{x}_{b}^i\}_{i=1}^{N_b}$, i.e., random samples of $\rho(\bm{x})$ and $\rho_b(\bm{x})$, the losses $\mathcal{L}_{pde}$, $\mathcal{L}_b$ and $\mathcal{L}_{\bm{g}}$can be approximated 
by
\begin{align}
&\mathcal{L}_{pde}
\approx\widehat{\mathcal{L}}_{pde}
\coloneqq\frac{1}{N_{pde}}\sum_{i=1}^{N_{pde}}\big(\mathcal{N}[\bm{x}^i;p_{\text{B-KRnet},\bm{\theta}}(\bm{x}^i),\bm{g}_{\text{NN}}(\bm{x}^i)]\big)^2,\\
&\mathcal{L}_b
\approx\widehat{\mathcal{L}}_b
\coloneqq\frac{1}{N_b}\sum\limits_{i=1}^{N_b}\big(\mathcal{B}[\bm{x}_b^i;p_{\text{B-KRnet},\bm{\theta}}(\bm{x}^i_b),\bm{g}_{\text{NN}}(\bm{x}^i_b)]\big)^2,\\
&\mathcal{L}_{\bm{g}}
\approx\widehat{\mathcal{L}}_{\bm{g}}
\coloneqq\frac{1}{N_{pde}}\sum\limits_{i=1}^{N_{pde}}\big\|\bm{g}_{\text{NN}}(\bm{x}^i)-\nabla p_{\text{B-KRnet},\bm{\theta}}(\bm{x}^i)\big\|^2_2,\label{eqn:def_g}
\end{align}
from which we define the total empirical loss $\widehat{\mathcal{L}}(p_{\text{B-KRnet},\bm{\theta}},\bm{g}_{\text{NN}})$ as
\begin{equation}
\label{eqn:PINN_loss}
\widehat{\mathcal{L}}(p_{\text{B-KRnet},\bm{\theta}},\bm{g}_{\text{NN}} )
=\lambda_{{pde}}\widehat{\mathcal{L}}_{{pde}}
+\lambda_{b}\widehat{\mathcal{L}}_b
+\lambda_{\bm{g}}\widehat{\mathcal{L}}_{\bm{g}}
.
\end{equation}
The optimal parameters $\bm{\theta}^*$, \(g^*_{\text{NN}}\) can be obtained as
\begin{equation}
\label{eqn:theta_optimal}
\bm{\theta}^*, g^*_{\text{NN}}=\arg\min_{\bm{\theta}, g_{\text{NN}}}\widehat{\mathcal{L}}(p_{\text{B-KRnet},\bm{\theta}}, \bm{g}_{\text{NN}}).
\end{equation} 

\subsection{Adaptive sampling procedure}\label{sec:adaptive_sampling_procedure}
The choice of training points plays an important part in achieving good numerical accuracy. We pay particular attention to the choice of $C_{pde}$ here. Typically, the training points are generated from a uniform distribution on a finite domain $\Omega$ which is inefficient when $p(\cdot)$ is concentrated in a small (yet known) region. Recently, adaptive sampling procedures have been successfully applied to solve Fokker-Planck equations defined on the whole domain \cite{tang2022adaptive, feng2021solving, zeng2023adaptive}. Noting that the residual $\left|\mathcal{N}[\bm{x}; p_{\text{B-KRnet},\bm{\theta}},\bm{g}_{NN}]\right|$ is larger more likely in the region of high probability density, their choice for $\rho$ in equation \eqref{eqn:Loss_pde} is $p_{\text{B-KRnet},\bm{\theta}}$, which also takes advantage of the fact that flow-based models can generate exact samples easily. 
 We adopt a similar procedure to solve PDEs on bounded domains.

The crucial idea is to update the training set by samples from the current optimal B-KRnet and then continue the training process. Since 
there is no prior knowledge of the solution at the beginning, we let $\rho$ be a uniform distribution, i.e., the initial set of collocation points $C_{pde}$ are drawn from a uniform distribution in $\Omega$. $C_b$ is also drawn from a uniform distribution on $\partial\Omega$. Then we solve the optimization problem \eqref{eqn:theta_optimal} via the Adam optimizer and obtain optimal $\bm{\theta}^{*,0}$, which corresponds to a bounded mapping $f_{\bm{\theta}^{*,0}}$ and a PDF $p_{\text{B-KRnet},\bm{\theta}^{*,0}}$. We update part of the collocation points in $C_{pde}$ by drawing samples from $p_{\text{B-KRnet},\bm{\theta}^{*,0}}$, i.e., update $\rho=(1-\gamma)\rho+\gamma p_{\text{B-KRnet},\bm{\theta}^{*,0}}$ where $0<\gamma\leq1$. More specifically, we form the new training set $C_{pde}^1$ using a mixture distribution given by the current training set and samples from $\bm{X}=(f_{\bm{\theta}^{*,0}})^{\rm{-1}}(\bm{Z})$, where $\bm{Z}$ is a uniform random variable. Then we continue to train the B-KRnet with $C_{pde}^1$ and $C_b$ to obtain a new set of optimal parameters, $\bm{\theta}^{*,1}$. Repeat this procedure until the maximum number of updates is reached. In this way, more collocation points will be chosen in the region of higher density while fewer collocation points in the region of lower density. Such a strategy can be concluded as follows.

\begin{enumerate}
	\item[1.] Let $\rho(\bm{x})=\frac{1}{|\Omega|}$. Generate an initial training set with samples uniformly distributed in $\Omega$:
	$$C_{pde}^0=\{\bm{x}^{i,0}\}_{i=1}^{N_{pde}}\subset \Omega, \quad \bm{x}^{i,0}\sim {\mathrm{Uniform}}\;\Omega,$$
		$$C_{b}=\{\bm{x}^{i}_{b}\}_{i=1}^{N_{b}}\subset \partial\Omega, \quad \bm{x}^{i}_{b}\sim {\mathrm{Uniform}}\; \partial\Omega.$$
	\item[2.] Train B-KRnet by solving optimization problem \eqref{eqn:theta_optimal} with training datasets $C_{pde}^0$ and $C_b$ to obtain 
	$\bm{\theta}^{*,0}$.
	\item[3.] $\rho(\bm{x})=(1-\gamma)\rho(\bm{x})+\gamma p_{\text{B-KRnet},\bm{\theta}^{*,0}}(\bm{x})$, $N_{new}=\gamma * N_{pde}$. Generate samples from $p_{\text{B-KRnet},\bm{\theta}^{*,0}}(\cdot)$ to get a new training set $C_{pde}^1=\{\bm{x}^{i,1}\}_{i=1}^{N_{new}}$. The new collocation point $\bm{x}^{i,1}$ can be obtained by transforming the uniformly distributed sample via the inverse normalizing flow,
	\begin{align*}
	\bm{z}^{i,1}\sim \mathrm{Uniform}(-1,1)^d, \quad \bm{x}^{i,1}=\big(f_{\bm{\theta}^{*,0}}\big)^{\rm{-1}}(\bm{z}^{i,1}).
	\end{align*}
		Set $C_{pde}^0=C_{pde}^1\cup\{\bm{x}^{i,0}\}_{i=N_{new}+1}^{N_{pde}}$. 
	\item[4.] Repeat steps 2-3 for $N_{\mathrm{adaptive}}$ times to get a convergent approximation.
\end{enumerate}

Our algorithm is summarized in Algorithm \cref{alg:1}. 

\begin{algorithm}
	\caption{Solving PDEs involving density}
	\label{alg:1}
	\begin{algorithmic}
		\STATE \textbf{Input:} maximum epoch number $N_e$, maximum iteration number $N_{\mathrm{adaptive}},$ update rate $\gamma$, initial training data $C_{pde}=\{\bm{x}^i\}_{i=1}^{N_{pde}}$, $C_b=\{\bm{x}^i_b\}_{i=1}^{N_b}$, initial learning rate $l_r$, decay rate $\eta$, step size $n_s$.
		\FOR{$k=0,\cdots,N_{\mathrm{adaptive}}$}
		\FOR{$j=1,\cdots,N_e$}
            \IF{$((k-1)N_e + j)\%n_s==0$}
		\STATE $l_r=\eta*l_r$.
		\ENDIF 
		\STATE Divide $C_{pde}$, $C_b$ into $n$ mini-batches  $\{C^{ib}_{pde}\}_{ib=1}^n$, $\{C^{ib}_{b}\}_{ib=1}^n$ randomly.
		\FOR{$ib = 1,\cdots,n$}
		\STATE Compute the loss function \eqref{eqn:PINN_loss} ${\widehat{\mathcal{L}}^{ib}}$ for mini-batch data $C^{ib}_{pde}$ and $C^{ib}_b$,
		\STATE Update $\bm{\theta}$ using the Adam optimizer.
		\ENDFOR 
		\ENDFOR 
		\STATE $N_{new}=\gamma*N_{pde}$. Sample from $p_{\text{B-KRnet},\bm{\theta}}(\cdot)$ and update training set $C_{pde}$.
		\ENDFOR 
		\STATE \textbf{Output:} The predicted solution $p_{\text{KRnet},\bm{\theta}}(\bm{x})$.
	\end{algorithmic}
\end{algorithm}

\section{Application to solving PDEs over mixed domains}\label{sec:mixed_domain}
In practical applications, we often encounter mixed regions. For example, the solution of kinetic Fokker-Planck equation is a PDF of location and velocity, where in general the location is bounded and the velocity is unbounded. We consider the following PDE,
\begin{equation}\label{eqn:mix_PDE}
\left\{\,
\begin{aligned}
&\widetilde{\mathcal{N}}[\bm{x},\bm{v}; p(\bm{x},\bm{v})] =  f(\bm{x},\bm{v}),\quad \bm{x}\in \Omega, \quad\bm{v}\in\mathbb{R}^d,\\
&\widetilde{\mathcal{B}}[\bm{x},\bm{v}; p(\bm{x},\bm{v})]=0,\qquad\quad \bm{x}\in \partial\Omega,\quad\bm{v}\in\mathbb{R}^d,
\end{aligned}\right.
\end{equation}
where $p(\bm{x};\bm{v})$ is assumed to be a PDF, \(\widetilde{\mathcal{N}}\) and \(\widetilde{B}\) respectively represent a differential operator and a boundary operator acting on \(p\).
To deal with problems defined on mixed regions, we develop a conditional flow, 
combining the original unbounded KRnet and B-KRnet.

For $\bm{v}\in\mathbb{R}^d$, $d>1$, and $\bm{x}\in\Omega$, we rewrite the PDF as 
\begin{equation}
p(\bm{x},\bm{v})=h(\bm{v}|\bm{x})\cdot q(\bm{x}),
\end{equation}
where $q(\bm{x})$ denotes the marginal distribution of spatial variable $\bm{x}$ and $h(\bm{v}|\bm{x})$ denotes the probability of velocity $\bm{v}$ at the position $\bm{x}$. We model $q(\bm{x})$ by B-KRnet and $h(\bm{v}|\bm{x})$ by original KRnet, i.e.,
\begin{equation}
\label{eqn:cond_KRnet}
p_{\bm{\theta}}(\bm{x},\bm{v})=h_{\text{KRnet},\bm{\theta}_1}(\bm{v}|\bm{x})\cdot q_{\text{B-KRnet},\bm{\theta}_2}(\bm{x}).
\end{equation}
{\color{black} The construction of the conditional PDF $h_{\text{KRnet},\bm{\theta}_1}(\bm{v}|\bm{x})$ can be defined as in \cite{feng2021solving}, where the only difference is that the conditional PDF in \cite{feng2021solving} is in terms of time instead of location.}
Following the similar discussion in \cref{sec:aux_nn}, we can easily derive the loss function for problem \eqref{eqn:mix_PDE}. We can also use the procedure described in \cref{sec:adaptive_sampling_procedure} to develop an adaptive sampling strategy. 

\section{Numerical results}\label{sec:numerical_experiments_for_pdes}
In this section, we provide several numerical experiments to demonstrate the effectiveness of the proposed approach presented in \cref{alg:1}. 
We first consider a four-dimensional test problem and then apply our method to the Keller-Segel equations which involve two PDFs. To further demonstrate the efficiency of B-KRnet, we combine B-KRnet and KRnet to solve the kinetic Fokker-Planck equation. We employ the hyperbolic tangent function (Tanh) as the activation function. Three subintervals are used in each CDF coupling layer. $\text{NN}(\bm{y}_1)$ in equation \eqref{eqn:theta_NN} is a fully connected network with two hidden layers. The algorithms are implemented with Adam optimizer \cite{kingma2014adam} in Pytorch.  
The update rate for a new training set is set to $\gamma=0.8$. 

\subsection{Four-dimensional test problem}
We start with the following four-dimensional problem, 
\begin{equation}
\label{eqn:toy_ex}
\left\{\enspace\begin{aligned}
&-\Delta p(\bm{x})+p(\bm{x})=f(\bm{x}), \quad \bm{x}\in\Omega=(0, \pi)^4,\\
& \quad\frac{\partial p(\bm{x})}{\partial \bm{n}}\bigg|_{\partial\Omega}=0,\\
&\int_{\Omega}p(\bm{x}){\rm d}\bm{x}=1, \quad p(\bm{x})\geq0,
\end{aligned}\right.
\end{equation}
where $f(\bm{x})=\frac{8}{9\pi^4}(5\cos{x_1}\cos{x_2}\cos{x_3}\cos{x_4}+\frac{9}{8})$. The exact solution is $p(\bm{x})=\frac{8}{9\pi^4}(\cos{x_1}\cos{x_2}\cos{x_3}\cos{x_4}+\frac{9}{8})$. 
We introduce an auxiliary function $\bm{g}$ to represent the gradient of $p$ and rewrite the equation \eqref{eqn:toy_ex} as
\begin{equation}
\label{eqn:toy_ex_g}
\left\{\enspace\begin{aligned}
&-\nabla\cdot \bm{g}(\bm{x})+p(\bm{x})=f(\bm{x}), \quad \bm{x}\in\Omega=(0, \pi)^4,\\
&  \bm{g}(\bm{x})\cdot \bm{n}(\bm{x})\big|_{\partial\Omega}=0,\\
&\nabla p(\bm{x})=\bm{g}(\bm{x}), \quad \bm{x}\in\Omega,\\
&\int_{\Omega}p(\bm{x}){\rm d}\bm{x}=1, \quad p(\bm{x})\geq0.
\end{aligned}\right.
\end{equation}
In the framework of PINN, we approximate the gradient by neural network $\bm{g}_{\text{NN}}$ and solution $p$ by B-KRnet $p_{\text{B-KRnet},\bm{\theta}}$. 
To enforce $\bm{g}_{\text{NN}}$ to satisfy the Neumann boundary condition exactly\cite{sheng2021pfnn, sheng2022pfnn}, we consider the following surrogate
\begin{equation}
\bm{g}_{\text{NN}}(\bm{x})=\text{diag}(\bm{x}(\pi-\bm{x}))\tilde{\bm{g}}(\bm{x}),
\end{equation}
where $\tilde{\bm{g}}_{NN}$ is a fully connected neural network with the structure 4-64-32-32-32-4. Therefore the final loss function admits
\begin{equation}
\widehat{\mathcal{L}}(p_{\text{B-KRnet},\bm{\theta}}, g_{\text{NN}})=\lambda_{pde}\widehat{\mathcal{L}}_{pde}
+\lambda_{\bm{g}}\widehat{\mathcal{L}}_{\bm{g}}.
\end{equation}

Let \(\lambda_{pde}=1, \lambda_{\bm{g}}=2\). For the B-KRnet, we deactivate the dimensions by one, i.e., $K=4$, and let $l_1=8$, $l_2=l_3=l_4=6$. The neural network introduced in \eqref{eqn:theta_NN} has two fully connected hidden layers with $32$ hidden neurons.  The number of epochs is $500$, and four adaptivity iterations are conducted. The Adam method with an initial learning rate of $0.001$ is applied. 
The learning rate is halved every 500 epochs. The number of training points is $4000$ and the batch size is $2000$. A validation dataset with $5\times10^4$ samples is used for calculating the relative errors throughout the entire training process. 

\begin{figure}[h!]
	\centering
	\begin{minipage}[b]{0.35\linewidth}
           \hspace{.2cm}
		\includegraphics[height=3cm,width=3.8cm]{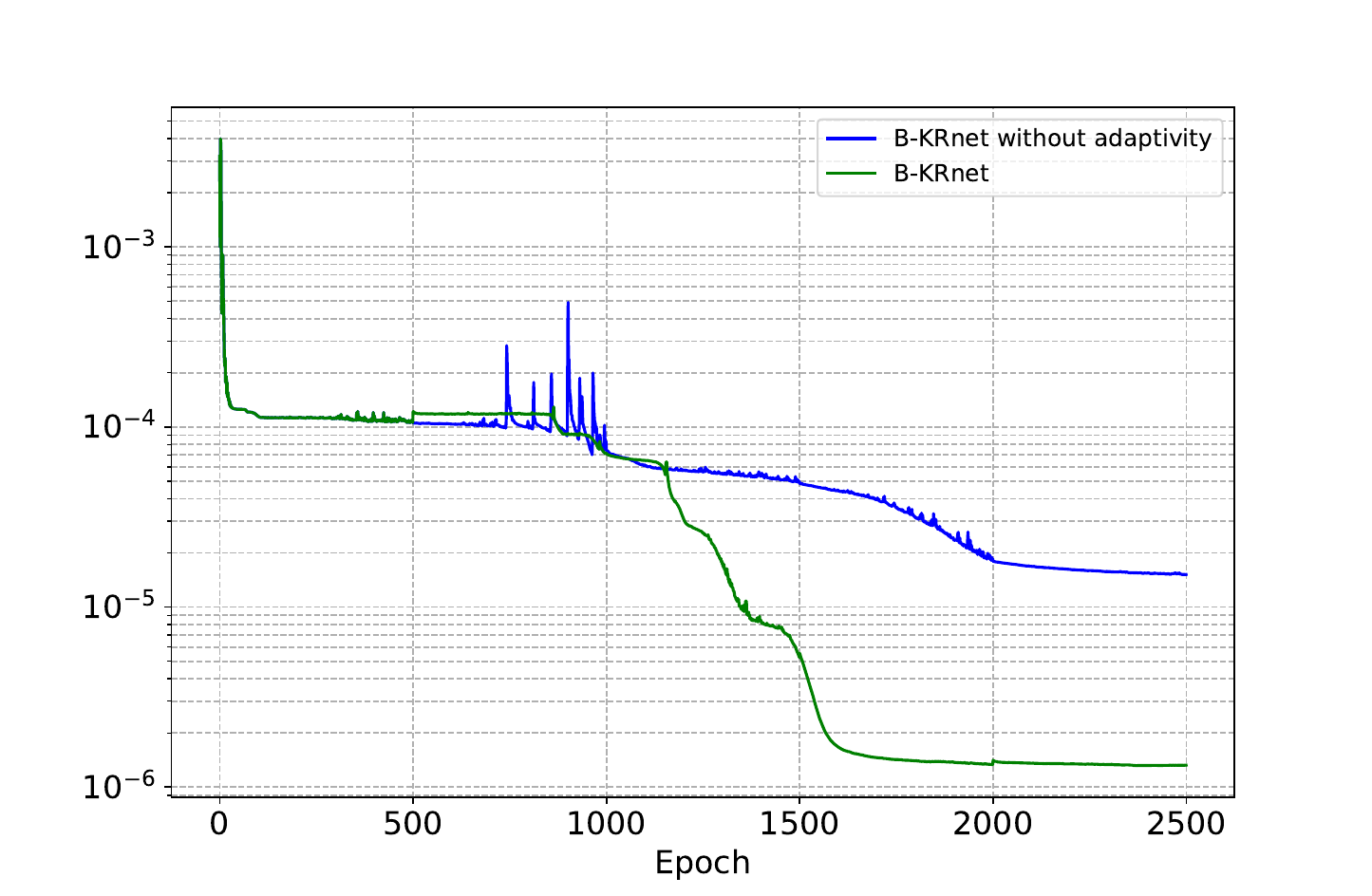}
		\subcaption{The training loss}
	\end{minipage}
	\begin{minipage}[b]{0.5\linewidth}
            \hspace{1cm}
		\includegraphics[height=3cm,width=3.8cm]{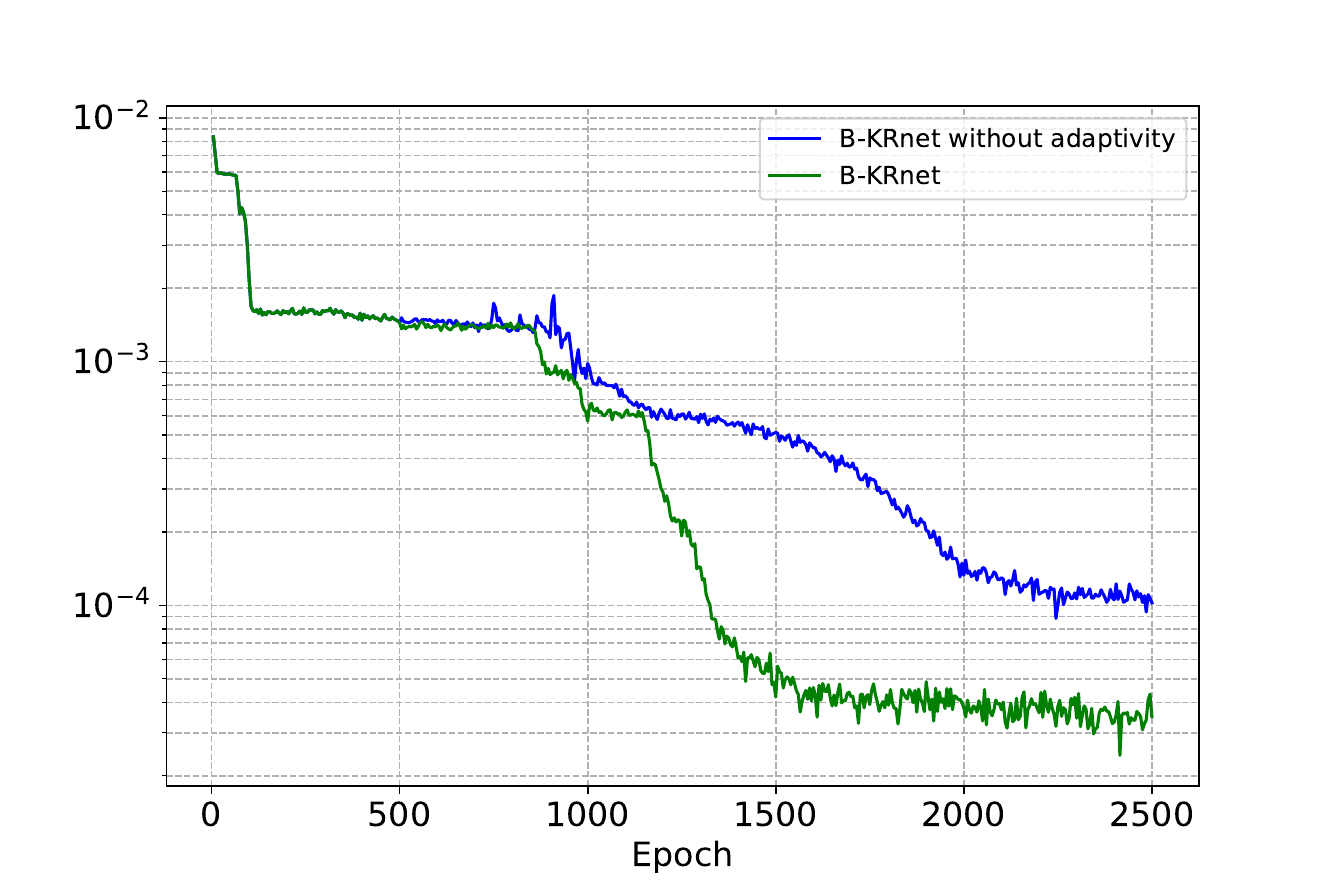}
		\subcaption{The relative KL divergence}
		\label{fig:ex0_kl}
	\end{minipage}

	\begin{minipage}[b]{0.35\linewidth}
           \hspace{.2cm}
		\includegraphics[height=3cm,width=3.8cm]{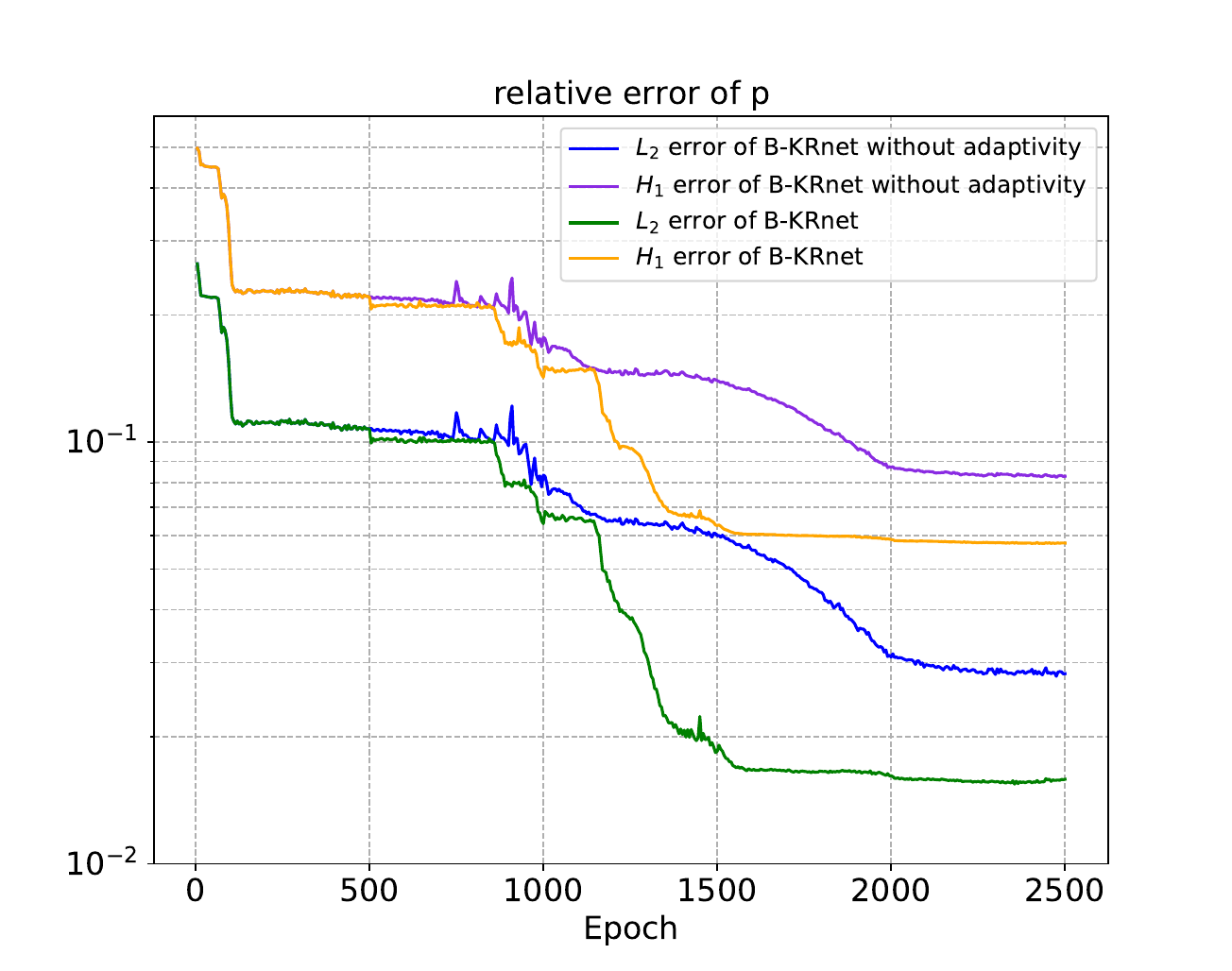}
		\subcaption{The relative error of $p$}
		\label{fig:ex0_l2}
	\end{minipage}
	\begin{minipage}[b]{0.5\linewidth}
        \hspace{1cm}
	\includegraphics[height=3cm,width=3.8cm]{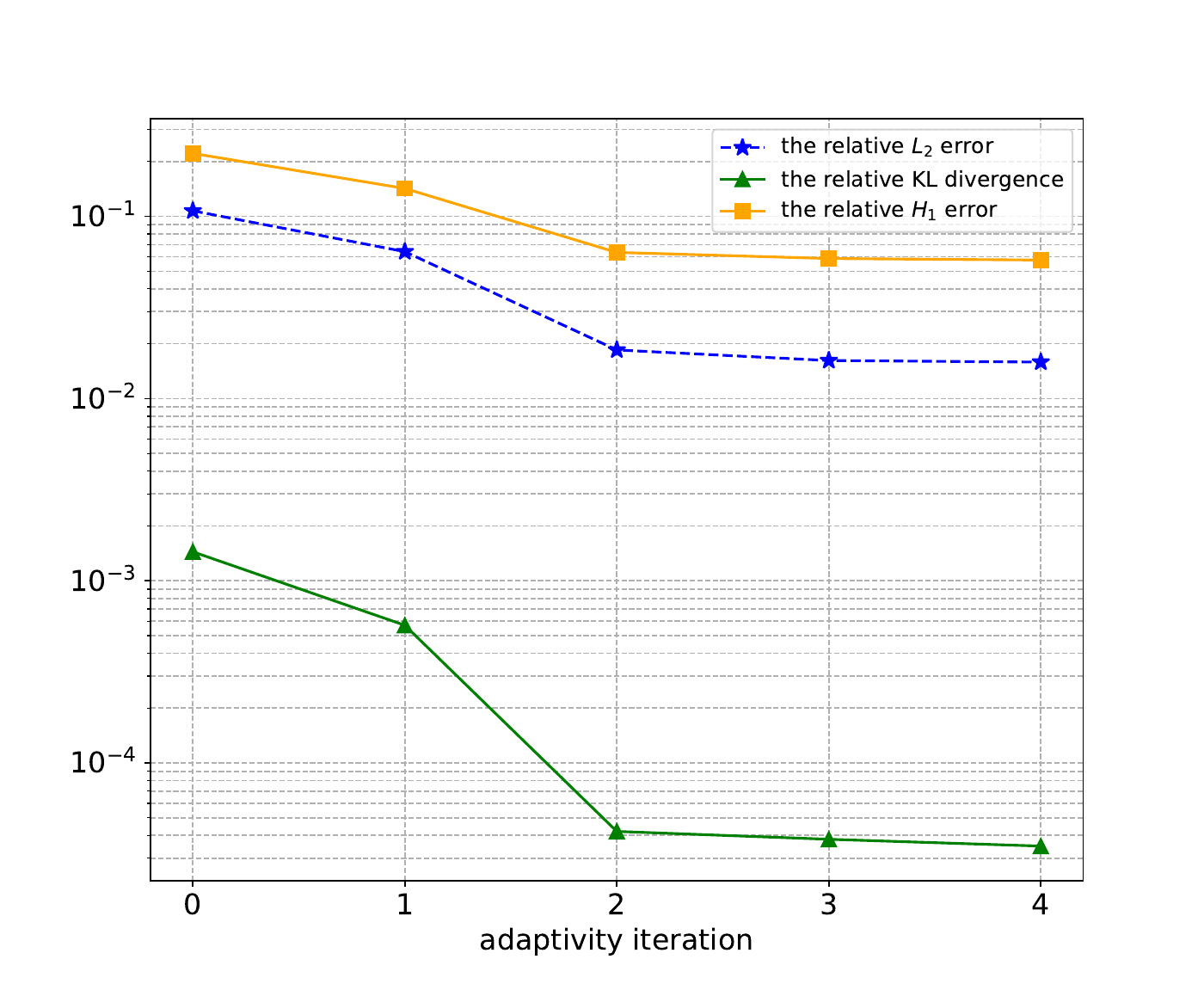}
	\subcaption{The relative error w.r.t. adaptivity iteration step}
 \label{fig:ex0_ada}
\end{minipage}
	\caption{Comparison between B-KRnet and B-KRnet without adaptivity for the four-dimensional problem. }
	\label{fig:Ex0_d4}
\end{figure}

  \Cref{fig:Ex0_d4} records the results of training processes with and without adaptive sampling. From \cref{fig:ex0_kl} and \cref{fig:ex0_l2}, we can observe that the relative error of adaptive B-KRnet is smaller than that of B-KRnet without adaptivity. Moreover, the convergence of B-KRnet is much faster. \Cref{fig:ex0_ada} plots the relative errors with respect to the adaptivity iteration steps.

\subsection{Stationary Keller-Segel system}
We consider the following stationary Keller-Segel equation \cite{hu2021positivity},
\begin{equation}
\left\{\enspace\begin{aligned}
&\Delta u(\bm{x})-\nabla\cdot(u\nabla v)(\bm{x})+f(\bm{x})=0, & \bm{x}\in\Omega,\\
&-\Delta v(\bm{x})+v(\bm{x})=u(\bm{x}), & \bm{x}\in\Omega,\\
& \quad\frac{\partial u}{\partial \bm{n}}\bigg|_{\partial\Omega}=0, \quad\frac{\partial v}{\partial \bm{n}}\bigg|_{\partial\Omega}=0,\\
&\int_{\Omega}u(\bm{x}){\rm d}\bm{x}=1, \quad u(\bm{x})\geq0,\quad \bm{x}\in\Omega,\\
&\int_{\Omega}v(\bm{x}){\rm d}\bm{x}=1, \quad v(\bm{x})\geq0,\quad \bm{x}\in\Omega,
\end{aligned}\right.
\label{eqn:sta_KS}
\end{equation}
where $\Omega=(0, \pi)^2$, $\bm{x}=(x,y)$, $u(\bm{x})$ is the unknown density of some bacteria, $v(\bm{x})$ is the density of chemical attractant and 
$$f(x,y)=\frac{1}{6\pi^2}\left(-4(-1+3\pi^2)\cos x\cos y+\cos^2x(1+3\cos 2y)-2\cos^2y\sin^2x\right).$$
The corresponding true solution is 
\begin{equation}
u(x,y)=\frac{1}{\pi^2}(\cos x\cos y+1),\quad v(x,y)=\frac{1}{3\pi^2}(\cos x\cos y+3).
\end{equation}
In this case,  two B-KRnets are required to represent $u$ and $v$ respectively. 
We introduce $\bm{\phi},\bm{\psi}$ to represent the derivative of $u,v$ respectively, i.e. $\bm{\phi}=\nabla u$, $\bm{\psi}=\nabla v$, and rewrite equation \eqref{eqn:sta_KS} as

\begin{equation}
\left\{\enspace\begin{aligned}
&\nabla\cdot \bm{\phi}(\bm{x})-\nabla u(\bm{x})\cdot\nabla v(\bm{x}) -(\nabla\cdot\bm{\psi})u(\bm{x})+f(\bm{x})=0,& \quad \bm{x}\in\Omega,\\
&-\nabla\cdot\bm{\psi}(\bm{x}) +v(\bm{x})=u(\bm{x}), &\bm{x}\in\Omega,\\
& \quad\bm{\phi}\cdot\bm{n}\big|_{\partial\Omega}=0, \quad\bm{\psi}\cdot\bm{n}\big|_{\partial\Omega}=0,\\
&\bm{\phi}(\bm{x})=\nabla u(\bm{x}), \quad \bm{\psi}(\bm{x})=\nabla v(\bm{x}),\quad \bm{x}\in\Omega,\\
&\int_{\Omega}u(\bm{x}){\rm d}\bm{x}=1, \quad u(\bm{x})\geq0,\quad\bm{x}\in\Omega,\\
& \int_{\Omega}v(\bm{x}){\rm d}\bm{x}=1, \quad v(\bm{x})\geq0, \quad\bm{x}\in\Omega.
\end{aligned}\right.
\end{equation}
We use $\bm{\phi}_{\text{NN}}$ and $\bm{\psi}_{\text{NN}}$ to approximate $\bm{\phi}$ and $\bm{\psi}$ respectively, which are defined as
\begin{equation}
\begin{aligned}
&\bm{\phi}_{\text{NN}}(x,y)=\text{diag}(x(\pi-x),\, y(\pi-y))\widetilde{\bm{\phi}}_{\text{NN}}(x,y),\quad \\
&\bm{\psi}_{\text{NN}}(x,y)=\text{diag}(x(\pi-x), \,y(\pi-y))\widetilde{\bm{\psi}}_{\text{NN}}(x,y),
\end{aligned}
\end{equation}
where $\widetilde{\bm{\phi}}_{NN}$ and $\widetilde{\bm{\psi}}_{NN}$ are both neural networks with four hidden layers.
Then our final loss function admits
\begin{equation}
\begin{aligned}
\widehat{\mathcal{L}}(u_{\text{B-KRnet},\bm{\theta}_1}, v_{\text{B-KRnet},\bm{\theta}_2}, \bm{\phi}_{\text{NN}},\bm{\psi}_{\text{NN}})
=\lambda_{pde, u}\widehat{\mathcal{L}}_{{pde, u}}+\lambda_{\bm{\phi}}\widehat{\mathcal{L}}_{\bm{\phi}}+\lambda_{pde,v}\widehat{\mathcal{L}}_{{pde,v}}\lambda_{\bm{\psi}}\widehat{\mathcal{L}}_{\bm{\psi}},
\end{aligned}
\end{equation}
where $\widehat{\mathcal{L}}_{\bm{\phi}}$ and  $\widehat{\mathcal{L}}_{\bm{\psi}}$ are defined by equation \eqref{eqn:def_g}.

\begin{figure}[h!]
	\centering
	\begin{minipage}[b]{0.25\linewidth}
		\includegraphics[height=2.4cm,width=2.7cm]{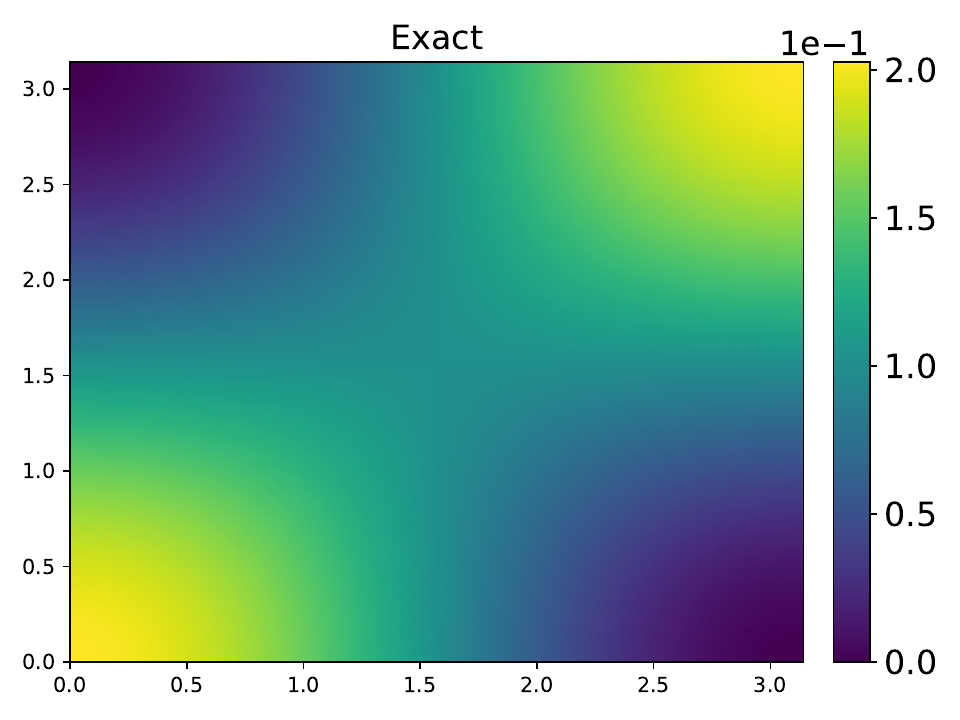}
	\end{minipage}
	\begin{minipage}[b]{0.25\linewidth}
	\includegraphics[height=2.4cm,width=2.7cm]{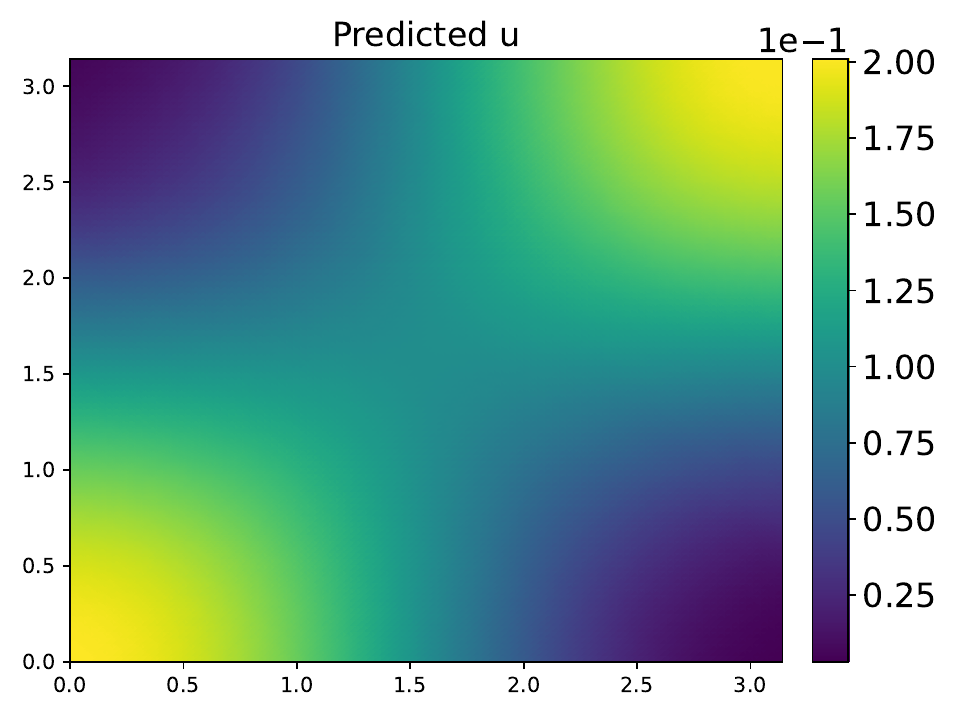}
\end{minipage}
\begin{minipage}[b]{0.25\linewidth}
	\includegraphics[height=2.4cm,width=2.7cm]{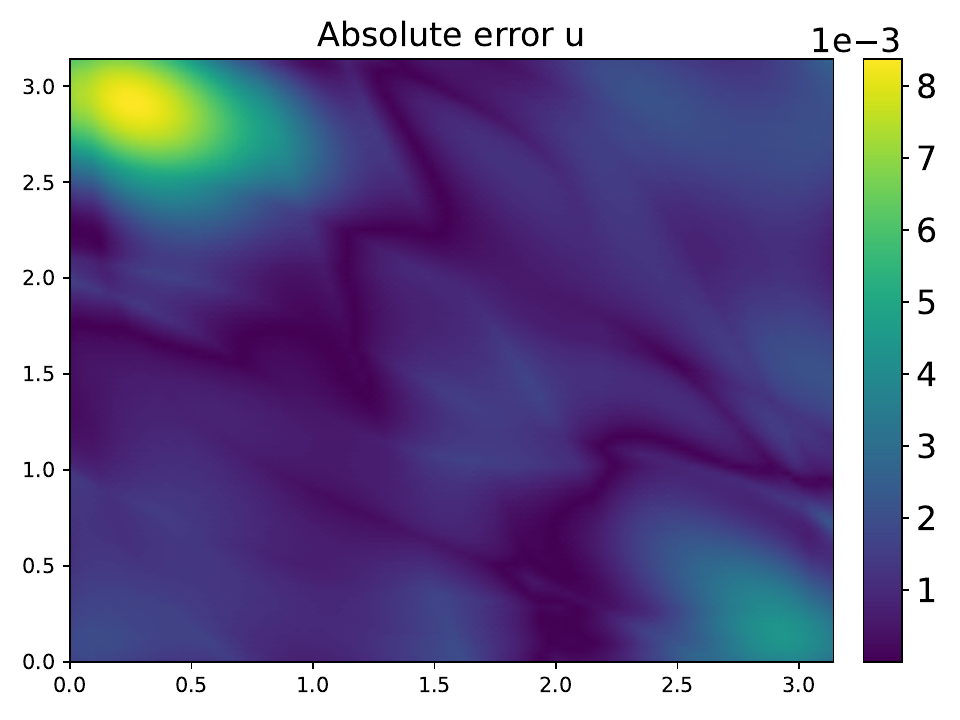}
\end{minipage}
	\caption{Comparison between true solution $u$ and numerical solution $u_{\text{B-KRnet},\bm{\theta}_1}$ of Keller-Segel equation. Left: The ground truth of $u$. Middle: Predicted solution $u_{\text{B-KRnet},\bm{\theta}_1}$. Right: $|u-u_{\text{B-KRnet},\bm{\theta}_1}|.$}
	\label{fig:KS_u_com}
\end{figure}

\begin{figure}[h!]
	\centering
	\begin{minipage}[b]{0.25\linewidth}
		\includegraphics[height=2.4cm,width=2.7cm]{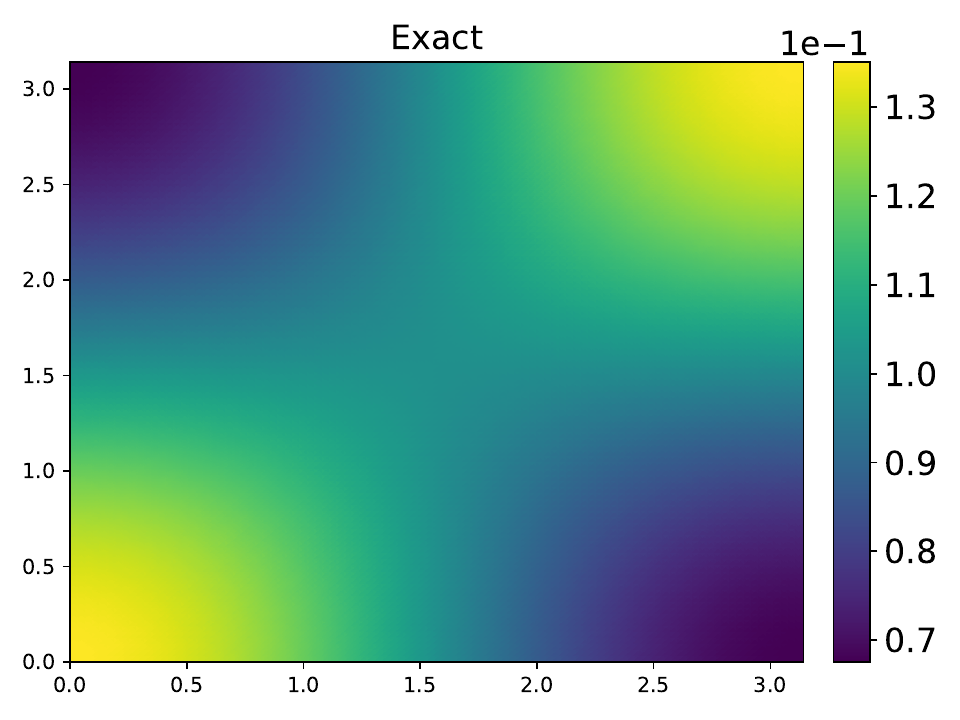}
	\end{minipage}
	\begin{minipage}[b]{0.25\linewidth}
	\includegraphics[height=2.4cm,width=2.7cm]{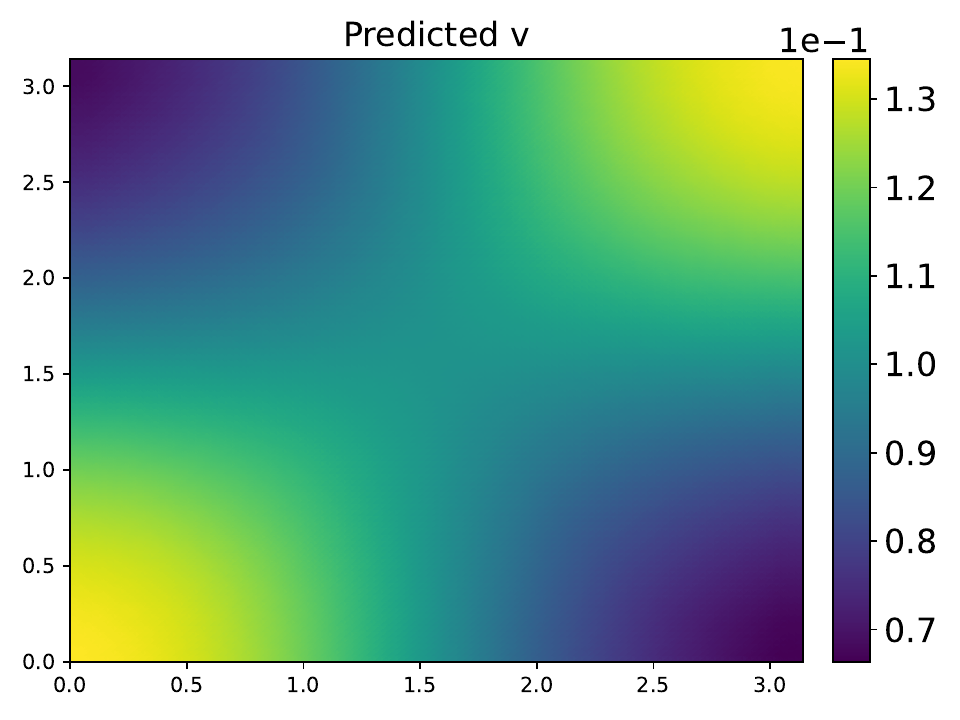}
	\end{minipage}
	\begin{minipage}[b]{0.25\linewidth}
	\includegraphics[height=2.4cm,width=2.7cm]{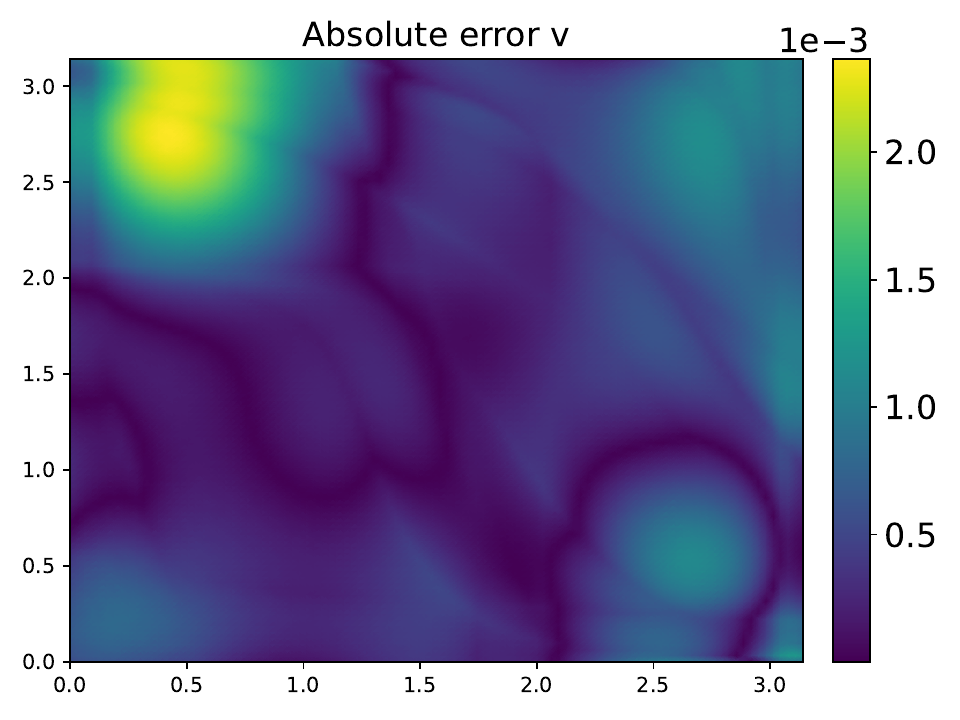}
	\end{minipage}
	\caption{Comparison between true solution $v$ and numerical solution $v_{\text{B-KRnet},\bm{\theta}_2}$ of Keller-Segel equation. Left: The ground truth of $v$. Middle: Predicted solution $v_{\text{B-KRnet},\bm{\theta}_2}$. Right: $|v-v_{\text{B-KRnet},\bm{\theta}_2}|$.}
	\label{fig:KS_v_com}
\end{figure}

\begin{figure}[h!]
	\centering
	\begin{minipage}[b]{0.28\linewidth}
		\includegraphics[height=3cm,width=3.8cm]{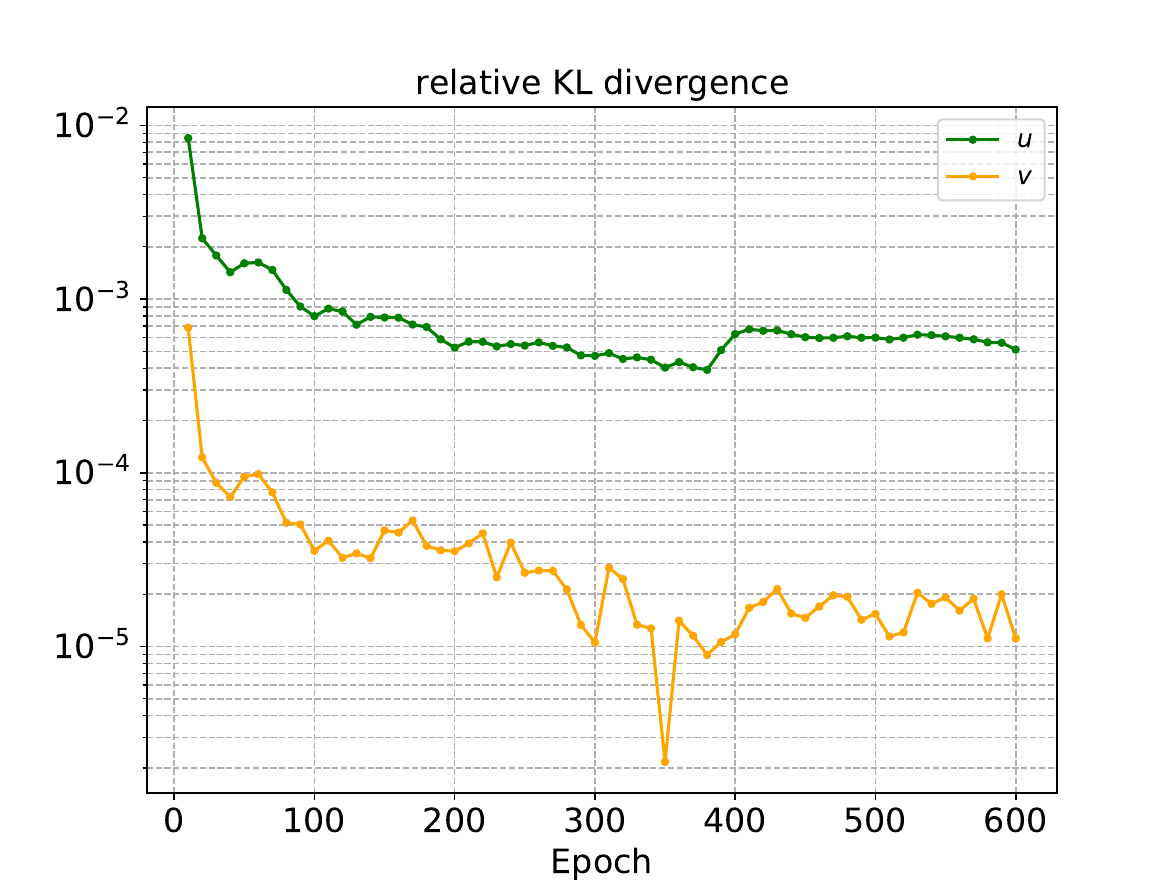}
	\end{minipage}
	\begin{minipage}[b]{0.28\linewidth}
		\includegraphics[height=3cm,width=3.8cm]{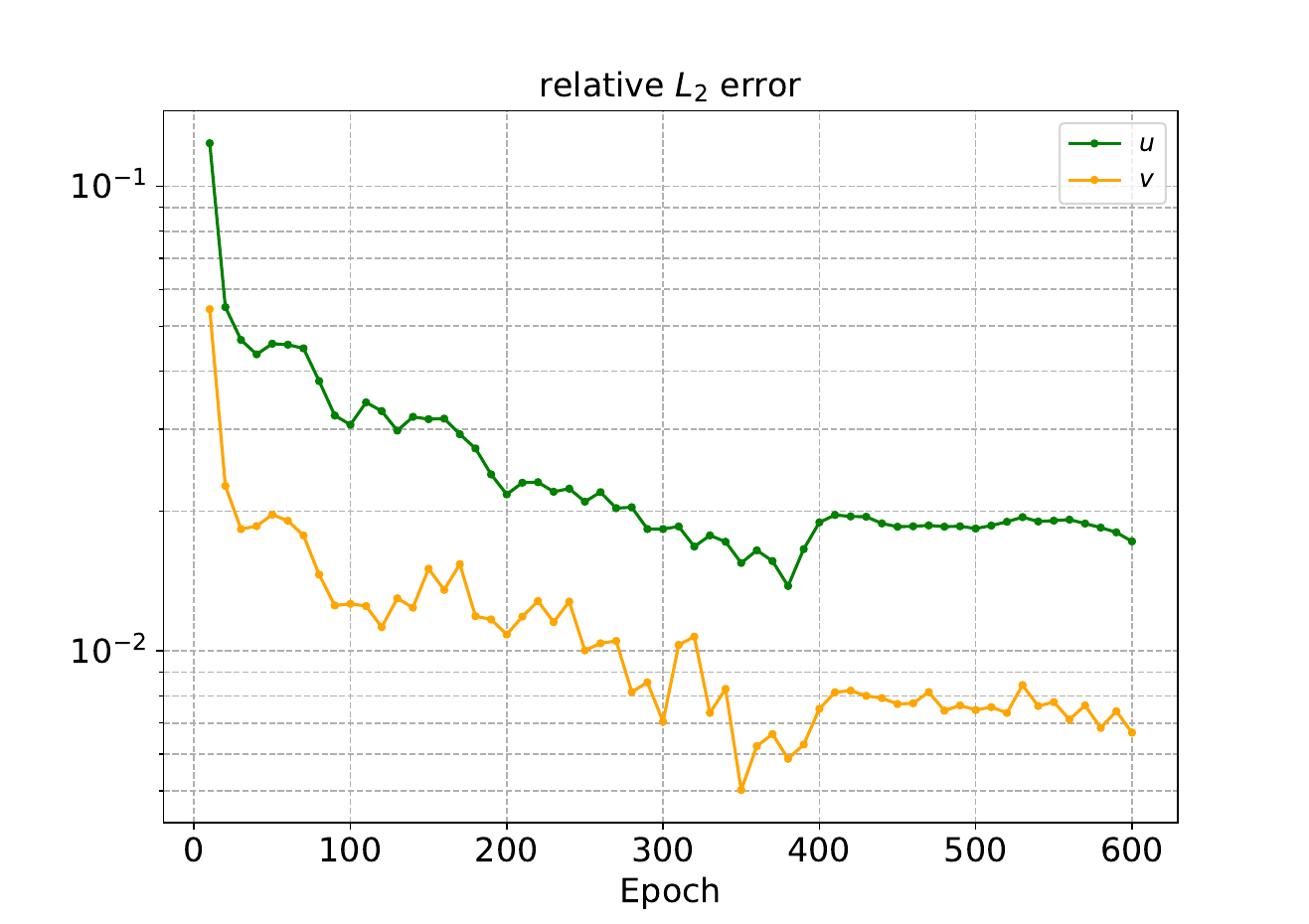}
	\end{minipage}
\begin{minipage}[b]{0.28\linewidth}
	\includegraphics[height=3cm,width=3.8cm]{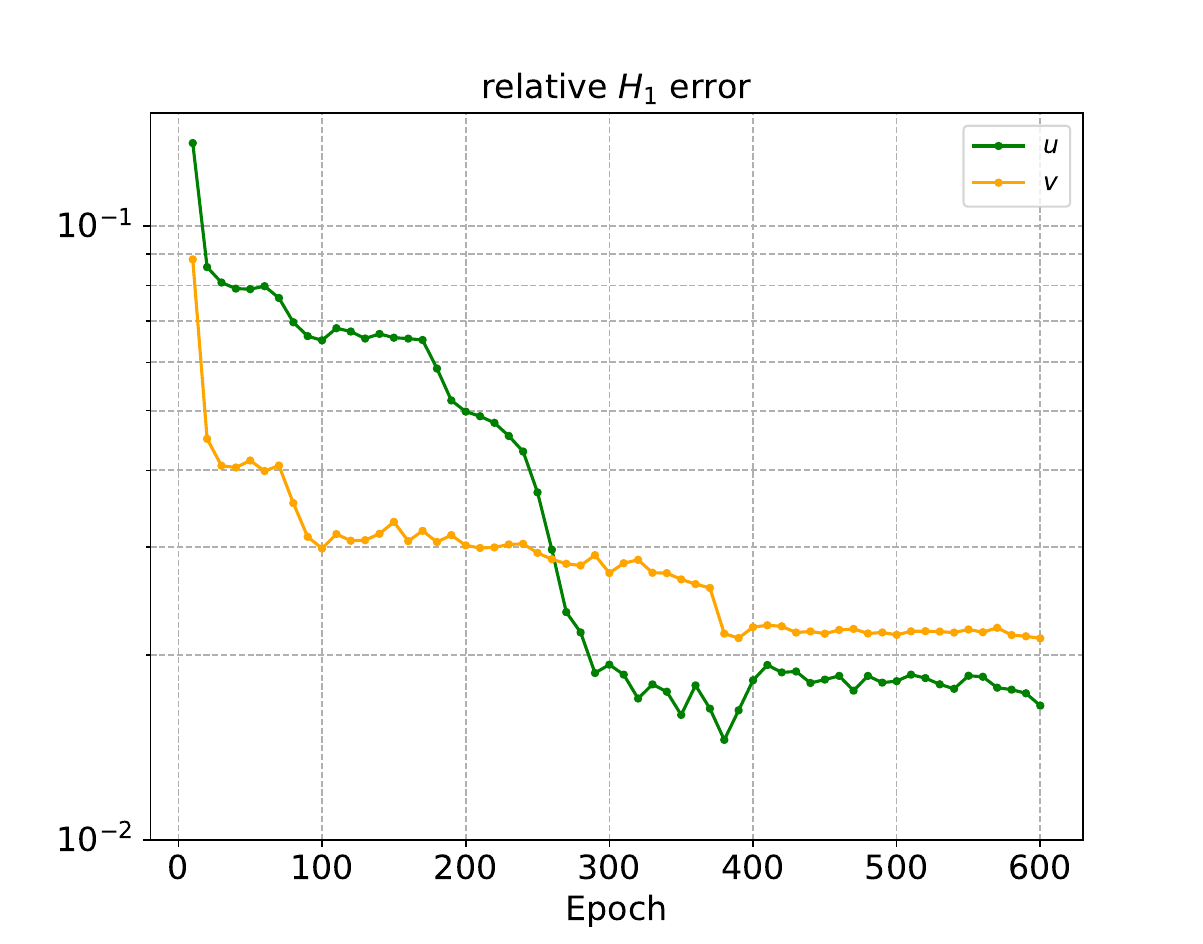}
\end{minipage}
	\caption{The relative error for Keller-Segel equation.}
	\label{fig:KS_error}
\end{figure}
For the B-KRnet, we take $8$ CDF coupling layers with $32$ hidden neurons. We use 2-64-32-32-32-2 network for $\widetilde{\bm{\phi}}_{\text{NN}}$ and $\widetilde{\bm{\psi}}_{\text{NN}}$.
$\lambda_u=\lambda_v=\lambda_{\bm{\phi}}=\lambda_{\bm{\psi}}=1$. We set the batch size to $1024$ and the number of training points to $10^4$. 
The initial learning rate is $0.001$, with a decay factor of half every $200$ epochs.
Five adaptivity iterations with 100 epochs are conducted. A validation dataset with $10^6$ samples is used for calculating the relative errors 
throughout the entire training process. 
The comparisons between the true solution and the predicted solution are presented in \cref{fig:KS_u_com} and \cref{fig:KS_v_com}.
The relative errors are presented in \cref{fig:KS_error}. The numerical solution provided by B-KRnet yields a good agreement with the analytic solution.

\subsection{2D stationary kinetic Fokker-Planck equation}
To illustrate the feasibility of the combination of KRnet and B-KRnet, we consider the following two-dimensional stationary kinetic Fokker-Planck equation with Dirichlet boundary conditions, i.e., 
\begin{equation}
\label{eqn:2D_sta_KFP}
\left\{\enspace\begin{aligned}
&-\bm{v}\cdot\nabla _{\bm{x}} p+\nabla_{\bm{v}} \cdot(\sigma\nabla _{\bm{v}} p+ \bm{v}p)=f, & (\bm{x},\bm{v})\in\Omega\times\mathbb{R}^2,\\
& p(\bm{x},\bm{v})\big|_{\partial \Omega}=0, & \bm{v}\in\mathbb{R}^2,\\
&\int_{\Omega\times\mathbb{R}}p(\bm{x},\bm{v})d\bm{x}d\bm{v}=1,  ~~p(\bm{x},\bm{v})\geq0, & (\bm{x},\bm{v})\in\Omega\times\mathbb{R}^2,
\end{aligned}\right.
\end{equation}
where $\Omega=(0, 2)^2$, \[f(\bm{x},\bm{v})=\frac{p}{\sigma}\|\bm{x}+\bm{v}\|^2-\frac{9}{16\sigma\pi}((1-x_1)x_2(2-x_2)v_1-(1-x_2)x_1(2-x_1)v_2)\exp\left(-\frac{\|\bm{v}+\bm{x}\|_2^2}{2\sigma}\right).\]
The ground truth is 
\begin{equation}
p(\bm{x},v)=\frac{9x_1x_2(2-x_1)(2-x_2)}{16\pi\sigma}\exp\left(-\frac{\|\bm{v}+\bm{x}\|_2^2}{2\sigma}\right).
\end{equation}

We consider the case that $\sigma=4$. This system is of second order with respect to $\bm{v}$ but of first order with respect to $\bm{x}$. We approximate the unknown $p$ by the formula \eqref{eqn:cond_KRnet}. Then 
 $$\frac{\partial^2 p_{\bm{\theta}}(\bm{x},\bm{v})}{\partial v_iv_j}=\frac{\partial ^2h_{\text{KRnet},\bm{\theta}_1}(\bm{v}|\bm{x})}{\partial v_iv_j}\cdot q_{\text{B-KRnet},\bm{\theta}_2}(\bm{x}).$$
 Since KRnet is second-order differentiable, we can directly obtain the loss function via PINN without converting the above system \eqref{eqn:2D_sta_KFP} into a first-order system.

\begin{figure}[h!]
	\centering
	\begin{minipage}[b]{0.25\linewidth}
		\includegraphics[height=2.4cm,width=2.7cm]{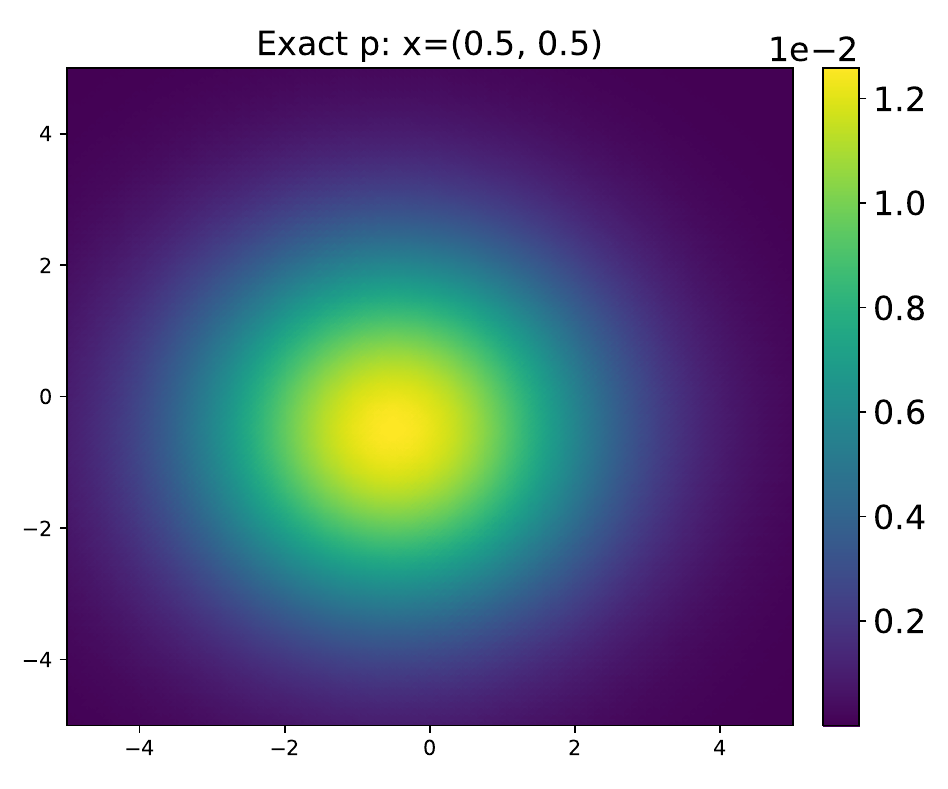}
	\end{minipage}
	\begin{minipage}[b]{0.25\linewidth}
		\includegraphics[height=2.4cm,width=2.7cm]{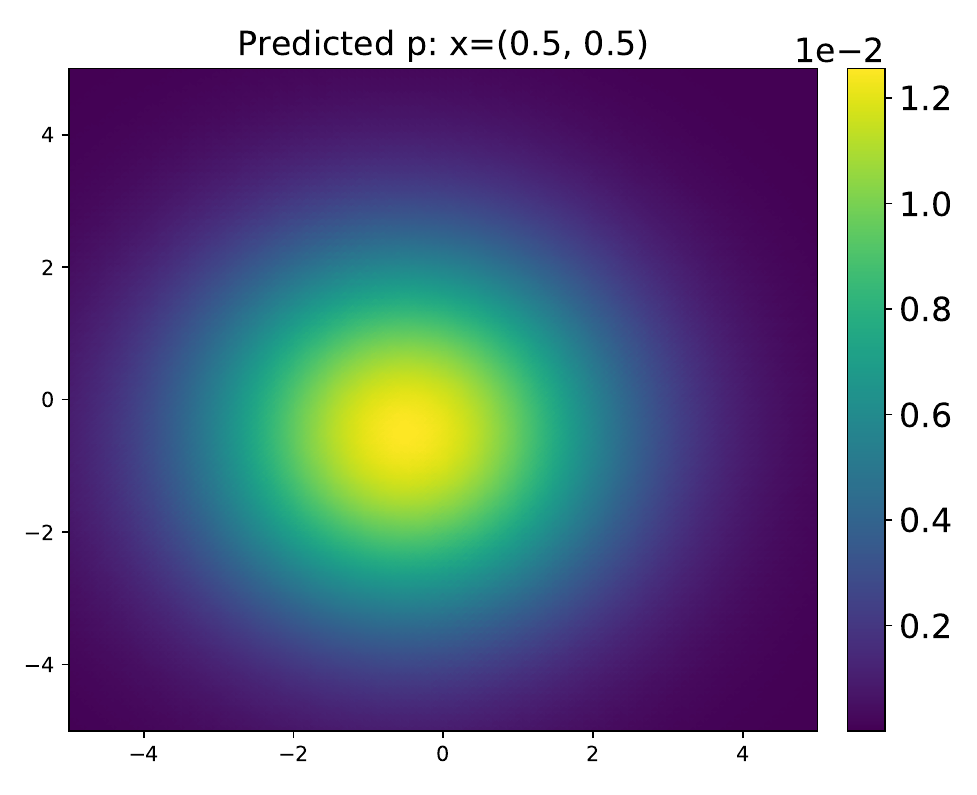}
	\end{minipage}
	\begin{minipage}[b]{0.25\linewidth}
		\includegraphics[height=2.4cm,width=2.7cm]{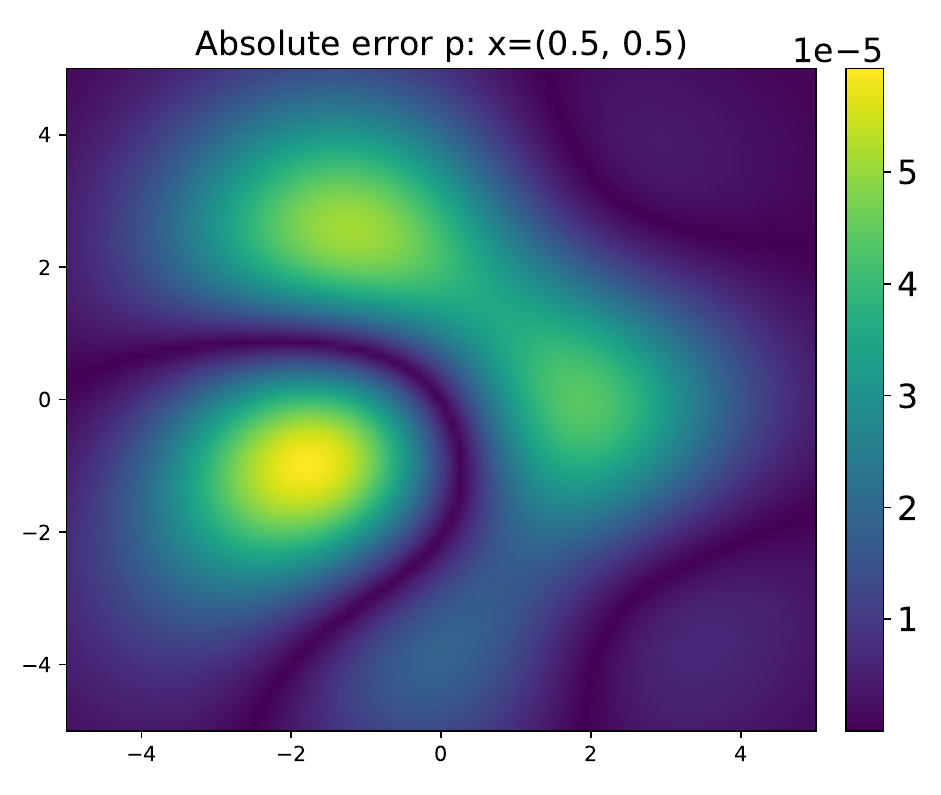}
	\end{minipage}
	
	\begin{minipage}[b]{0.25\linewidth}
		\includegraphics[height=2.4cm,width=2.7cm]{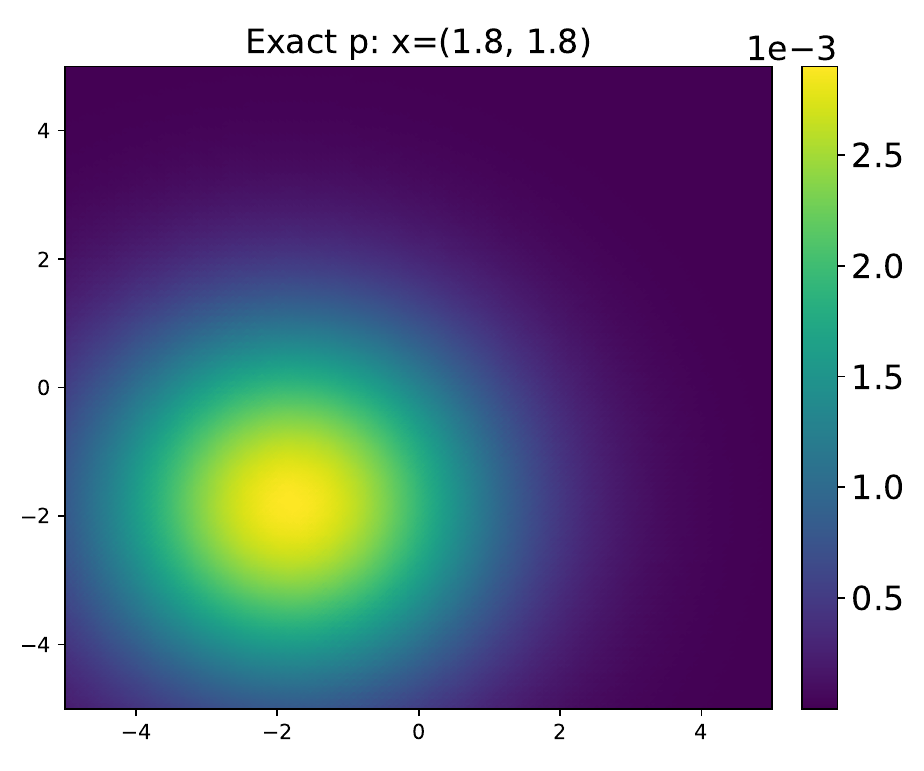}
	\end{minipage}
	\begin{minipage}[b]{0.25\linewidth}
		\includegraphics[height=2.4cm,width=2.7cm]{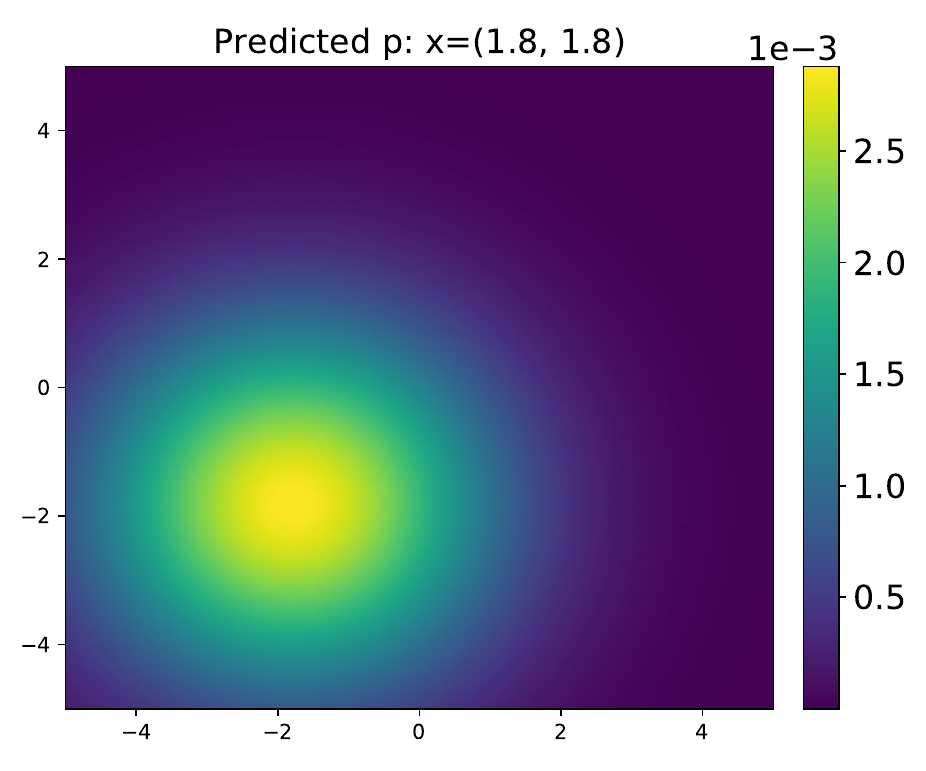}
	\end{minipage}
	\begin{minipage}[b]{0.25\linewidth}
		\includegraphics[height=2.4cm,width=2.7cm]{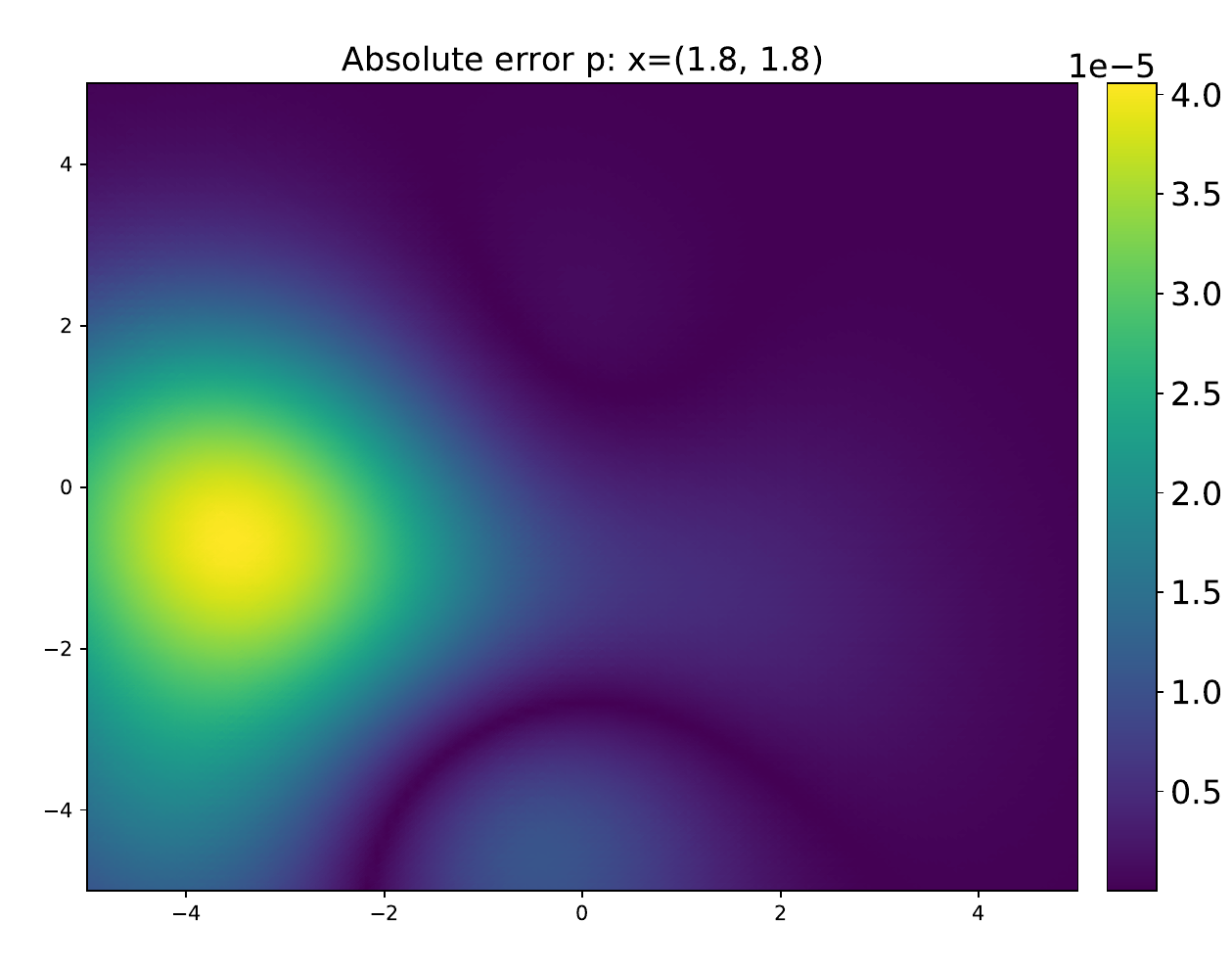}
	\end{minipage}
	\caption{$\sigma=4$. Comparisons between the predicted solutions and the reference solutions. Top row: $\bm{x}=(0.5, 0.5)$. Bottom row: $\bm{x}=(1.8, 1.8)$.}
	\label{fig:2d_sta_KFP_solu}
\end{figure}

\begin{figure}[h!]
	\centering
	\begin{minipage}[b]{0.28\linewidth}
		\includegraphics[height=3cm,width=3.8cm]{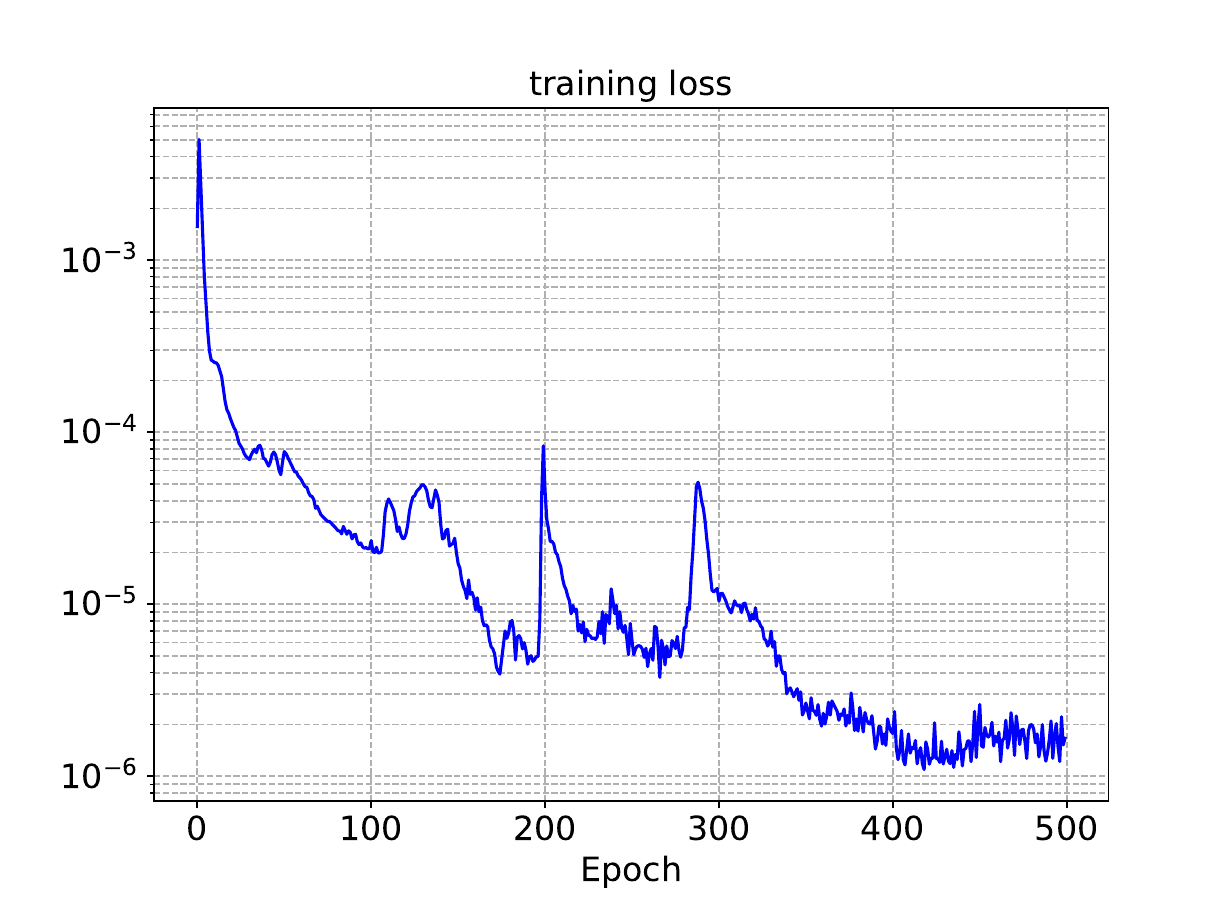}
	\end{minipage}
	\begin{minipage}[b]{0.28\linewidth}
		\includegraphics[height=3cm,width=3.8cm]{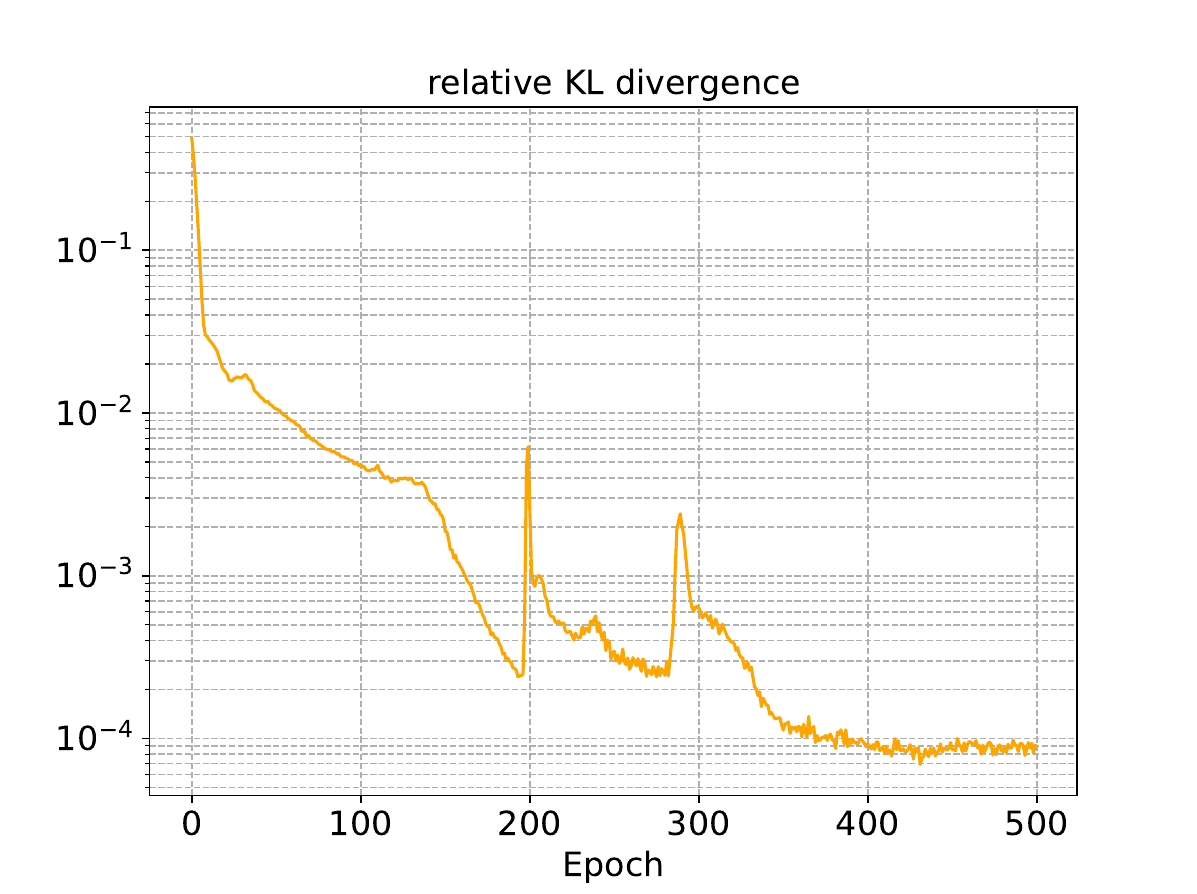}
	\end{minipage}
	\begin{minipage}[b]{0.28\linewidth}
		\includegraphics[height=3cm,width=3.8cm]{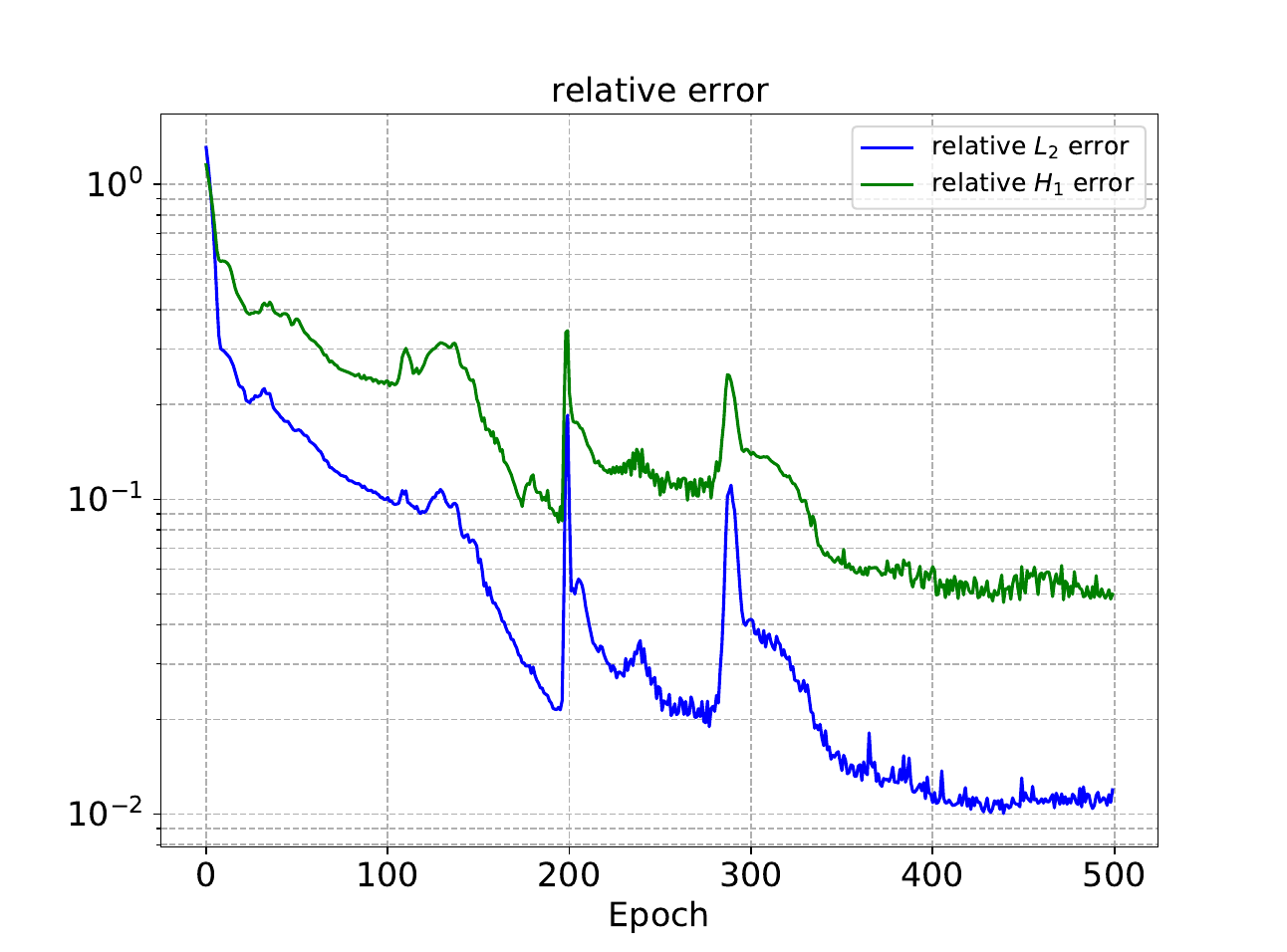}
	\end{minipage}
	\caption{The training loss and the relative error for kinetic Fokker-Planck equation.}
	\label{fig:2d_sta_KFP_loss_error}
\end{figure}

The number of training points is set to $6\times 10^4$. The initial training points are drawn from a uniform distribution over $[0,2]^2\times[-5,5]^2$. Both B-KRnet and KRnet consist of \(8\) CDF or affine coupling layers with $32$ hidden neurons.
 The hyperparameters $\lambda_{pde}$ is set to $1$ and $\lambda_{b}$ is set to $10$. 
 The Adam optimizer is used with a batch size of 3,000. The initial learning rate is 0.0002, with a decay factor of half every 400 epochs.
 The number of epochs between two adjacent adaptivity iterations is set to 1, and the number of adaptivity iterations conducted for this
problem is set to $N_{\text{adaptive}}=500$. A validation dataset with $10^6$ samples is used 
to calculate the relative errors throughout the entire training process.
 
The comparisons between the ground truth and the predicted solution at $\bm{x}=(0.5, 0.5)$ and $\bm{x}=(1.8, 1.8)$ are presented in \cref{fig:2d_sta_KFP_solu}. The training loss and the relative errors are presented in \cref{fig:2d_sta_KFP_loss_error}. We can observe that our method agrees well with the ground truth. 

\section{Conclusion}\label{sec:conclusions}
We have developed a bounded normalizing flow B-KRnet in this work, which can be considered as a bounded version of KRnet. B-KRnet is built on a structure with decreasing active transformation dimensions whose core layer is the CDF coupling layer. It models unknown PDFs via an invertible mapping from the target distribution to a uniform distribution. Taking advantage of invertibility, B-KRnet could provide an explicit PDF model and generate exact random samples as well. We apply B-KRnet to density estimation and approximation. The samples generated by B-KRnet show excellent agreement with the ground truth. Moreover, we develop an adaptive density approximation scheme for solving PDEs whose solutions are PDFs. A series of examples, including a four-dimensional problem, a Keller-Segel equation, and a kinetic Fokker-Planck equation, is presented to demonstrate the efficiency of our method. B-KRnet can also be coupled with adaptive sampling techniques to improve the neural network approximation of other types of PDEs \cite{tang2023pinns,zeng2023adaptive}. There are, however, many important issues that need to be addressed. The smoothness of B-KRnet is limited to the first order. For some equations with variational forms, the regularity of the current B-KRnet is sufficient for solving a second-order PDE when coupled with the deep Ritz method. How to update training points adaptively in the framework of the deep Ritz method is an interesting question. In the framework of PINN, to avoid converting a second-order PDE into a first-order system, we need to improve the regularity of B-KRnet, which can be addressed with high-order polynomials. These issues will be left for future study.

\bibliographystyle{siamplain}
\bibliography{references}
\end{document}